\documentclass[sigconf]{acmart}

\AtBeginDocument{%
  }

\setcopyright{none}
\renewcommand{\footnotetextcopyrightpermission}[1]{}

\makeatletter
\AtBeginDocument{%
  \fancyhead[LE]{\ACM@linecountL\@headfootfont Preprint.}%
  \fancyhead[RO]{\@headfootfont Preprint.\ACM@linecountR}%
}
\makeatother

\usepackage{amsmath,mathtools}
\usepackage{booktabs}
\usepackage{makecell}
\usepackage{multirow}
\usepackage{array}
\usepackage{enumitem}
\usepackage{algorithm}
\usepackage{algpseudocode}
\usepackage{tikz}
\usetikzlibrary{positioning,shapes.geometric}
\usepackage{xspace}
\usepackage{placeins}

\makeatletter
\renewcommand{\fs@ruled}{%
  \def\@fs@cfont{\bfseries}%
  \let\@fs@capt\floatc@ruled
  \def\@fs@pre{{\color{black}\hrule height.8pt depth0pt \kern2pt}}%
  \def\@fs@post{{\kern2pt\color{black}\hrule\relax}}%
  \def\@fs@mid{{\kern2pt\color{black}\hrule\kern2pt}}%
  \let\@fs@iftopcapt\iftrue
}
\makeatother

\makeatletter
\patchcmd{\@mkauthors@iii}
  {\unvbox\mktitle@bx\par\medskip\leavevmode}
  {\unvbox\mktitle@bx\par\medskip\vspace{1.2\baselineskip}\leavevmode}
  {}{}
\makeatother

\newcommand{\method}{\textsc{DARS}\xspace}
\newcommand{\planner}{q_{\theta}}
\newcommand{\renderer}{p_{\phi}}
\newcommand{\sigmoid}{\operatorname{sigmoid}}
\newcommand{\E}{\mathbb{E}}
\newcommand{\Var}{\operatorname{Var}}

\newcommand{\promptpanel}[2]{%
  \begin{tikzpicture}
    \node[
      draw=black,
      rounded corners=10pt,
      line width=0.8pt,
      fill=gray!8,
      inner sep=10pt,
      text width=0.94\textwidth,
      align=left
    ] (box) {%
      \small
      #2
    };
    \node[
      anchor=west,
      rounded corners=3pt,
      fill=black,
      text=white,
      inner xsep=7pt,
      inner ysep=3pt,
      font=\small\bfseries
    ] at ([xshift=12pt,yshift=-2pt]box.north west) {#1};
  \end{tikzpicture}%
}
\makeatletter
\let\DARSmainaddcontentsline\addcontentsline
\newcommand{\DARSsuppaddcontentsline}[3]{%
  \def\DARS@stream{#1}%
  \def\DARS@toc{toc}%
  \def\DARS@lof{lof}%
  \def\DARS@lot{lot}%
  \ifx\DARS@stream\DARS@toc
    \DARSmainaddcontentsline{sto}{#2}{#3}%
  \else\ifx\DARS@stream\DARS@lof
    \DARSmainaddcontentsline{slof}{#2}{#3}%
  \else\ifx\DARS@stream\DARS@lot
    \DARSmainaddcontentsline{slot}{#2}{#3}%
  \else
    \DARSmainaddcontentsline{#1}{#2}{#3}%
  \fi\fi\fi
}
\newcommand{\DARSsupplementarytableofcontents}{%
  \@starttoc{sto}\contentsname
}
\newcommand{\DARSsupplementarylistoffigures}{%
  \@starttoc{slof}\listfigurename
}
\newcommand{\DARSsupplementarylistoftables}{%
  \@starttoc{slot}\listtablename
}
\newcommand{\DARSstartsupplementaryindexes}{%
  \let\addcontentsline\DARSsuppaddcontentsline
  \def\ext@figure{slof}%
  \def\ext@table{slot}%
  \let\tableofcontents\DARSsupplementarytableofcontents
  \let\listoffigures\DARSsupplementarylistoffigures
  \let\listoftables\DARSsupplementarylistoftables
}
\makeatother

\begin{document}

\title[DARS for Instruction-Based Image Editing]{DARS: Dual-Level Credit Assignment RL with Structured Reasoning for Instruction-Based Image Editing}

\author{%
  Haoxiang Cao\textsuperscript{1,2,\S}\quad
  Jiajiong Cao\textsuperscript{2}\quad
  Xuanpu Zhang\textsuperscript{2,\S}\quad
  Changqian Yu\textsuperscript{2,\ensuremath{\ddagger}}\quad
  Chaoqun Wang\textsuperscript{1,\ensuremath{\dagger}}\\[1.4ex]
  {\normalsize\normalfont
  \textsuperscript{1}South China Normal University\qquad
  \textsuperscript{2}KlingAI Research}}
\renewcommand{\shortauthors}{Cao et al.}

% Add the consolidated author-role note through acmart's title-footnote
% mechanism so it is placed at the bottom of the first (left) column.
\makeatletter
\g@addto@macro\@authornotes{%
  \begingroup
  \renewcommand{\thefootnote}{}
  \footnotetext{%
    \textsuperscript{\ensuremath{\ddagger}}Project lead.\quad
    \textsuperscript{\ensuremath{\dagger}}Corresponding author.\par\noindent
    \textsuperscript{\S}Work done during internship in KlingAI Research.}
  \endgroup}
\makeatother

\begin{abstract}
Instruction-based image editing uses a planner-renderer pipeline: a vision-language model (VLM) first converts the instruction into an edit plan, and a diffusion model then executes that plan. Training such systems with only final-image rewards is inefficient because a poor edit does not reveal whether additional optimization should place more emphasis on the planner or the renderer, and even planner-dominant cases remain difficult to localize within a free-form reasoning trace. We present \method, a reinforcement learning framework for dual-level credit assignment in this two-stage setting. Across modules, multi-plan multi-render rollouts estimate between-plan and within-plan reward variability for soft module routing, while mean rewards across rollouts provide hardness estimates for an adaptive curriculum. Within the planner, a four-field structured reasoning output enables a prefix-gated reward and token-level advantage reweighting, turning outcome-level feedback into localized supervision. Experiments on five benchmarks show that \method outperforms a Joint~RL baseline with the same backbone, data, reward model, and rollout budget, with the largest gains on reasoning-intensive edits. 
% A validation study shows that the rollout-based routing signals align with GPT-5 pseudo-labels of dominant failure patterns.

\end{abstract}

\begin{teaserfigure}
  \centering
  \vspace{1\baselineskip}
  \includegraphics[width=\textwidth]{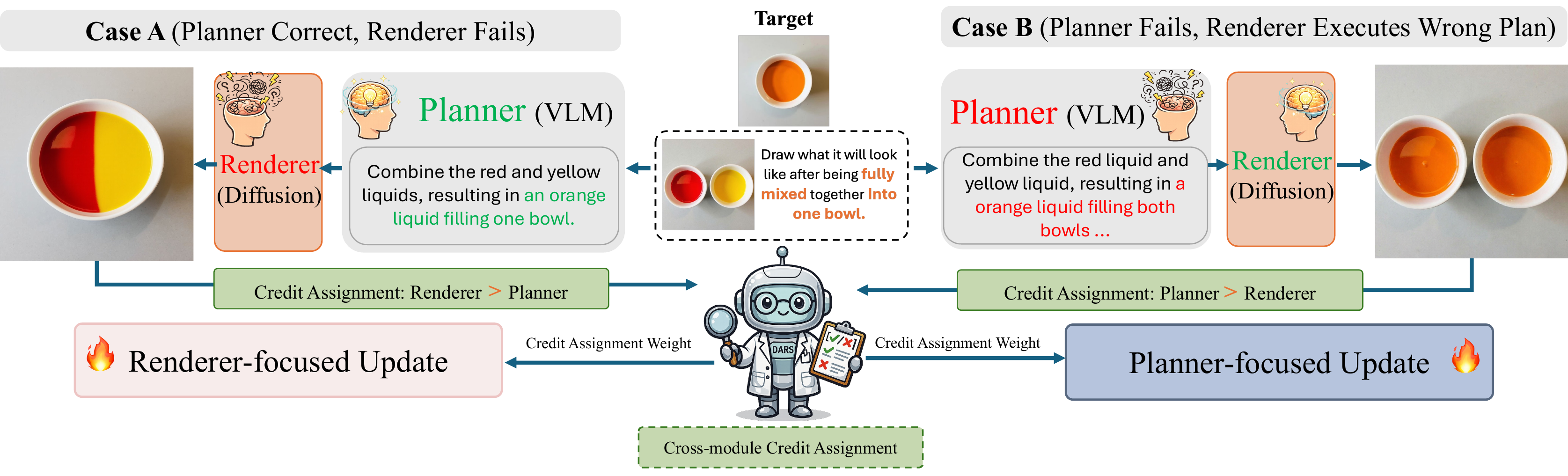}
  \captionsetup{skip=15pt}
  \caption{Two visually poor editing outcomes can require opposite cross-module update emphases. \textbf{Case~A}: the planner captures the requested edit reasonably well, but the renderer fails to realize it, so this sample should place more corrective update weight on the renderer. \textbf{Case~B}: the sampled plan is a less useful intermediate for the instruction, and the renderer faithfully executes it, so this sample should place more corrective update weight on the planner. \method is designed to distinguish render-dominant from plan-dominant update needs through routing signals; when planner-side emphasis is needed, its structured slots support finer within-planner diagnosis.}
  \Description{Teaser showing two low-scoring failure cases in a two-stage image editing pipeline that require opposite cross-module credit-assignment decisions: one should emphasize renderer correction, while the other should emphasize planner correction.}
  \label{fig:teaser}
  \vspace{1.3\baselineskip}
\end{teaserfigure}

\maketitle

\section{Introduction}
\label{sec:intro}

Many image editing systems~\cite{ReasonEdit,EditThinker,PromptRL,ThinkRL-Edit,RePlan} adopt a planner-renderer pipeline: a vision-language model (VLM)~\cite{Qwen3-VL,qwen3.5,Qwen2.5-VL} interprets the instruction and produces an edit plan, and a diffusion model~\cite{flux2,Qwen-Image,LongCat-Image,Step1X-Edit} renders the image conditioned on that plan. This decomposition improves controllability, but also turns joint optimization into a dual-level credit-assignment problem. Across modules, as illustrated in Figure~\ref{fig:teaser}, a poor image does not reveal whether the current sample would benefit more from planner-side or renderer-side corrective emphasis. Within the planner, even when planner-side updates are the more useful direction, the same outcome-level feedback still does not reveal which part of the plan should be corrected.

The difficulty comes from the fact that the two modules fail in different ways. The planner is an autoregressive language model~\cite{GPT}, so its errors are typically semantic or compositional. The renderer is a diffusion-based generator~\cite{DDPM}, so its failures appear as visual artifacts, preservation damage, or incomplete execution. A single post-render score~\cite{EditReward,EditScore} collapses these qualitatively different errors into the same feedback, making credit assignment ambiguous. Existing RL~\cite{ThinkGen,RePlan,PromptEnhancer,EditThinker} strategies, therefore, either optimize one module at a time, leaving the other bottleneck untouched, or update both modules uniformly from the same reward. In either case, a low-scoring edit still does not indicate whether additional optimization should focus more on planning or on rendering.

This makes credit assignment~\cite{VCRL} under-specified at two levels. First, different samples should not contribute equally to planner and renderer updates. Second, even when the planner is the main bottleneck, outcome-level feedback still does not say which part of the plan is wrong: the requested modification, the preservation constraint, the scene-level objective, or the execution hint.

\begin{figure*}[t]
  \centering
  \includegraphics[width=0.95\textwidth]{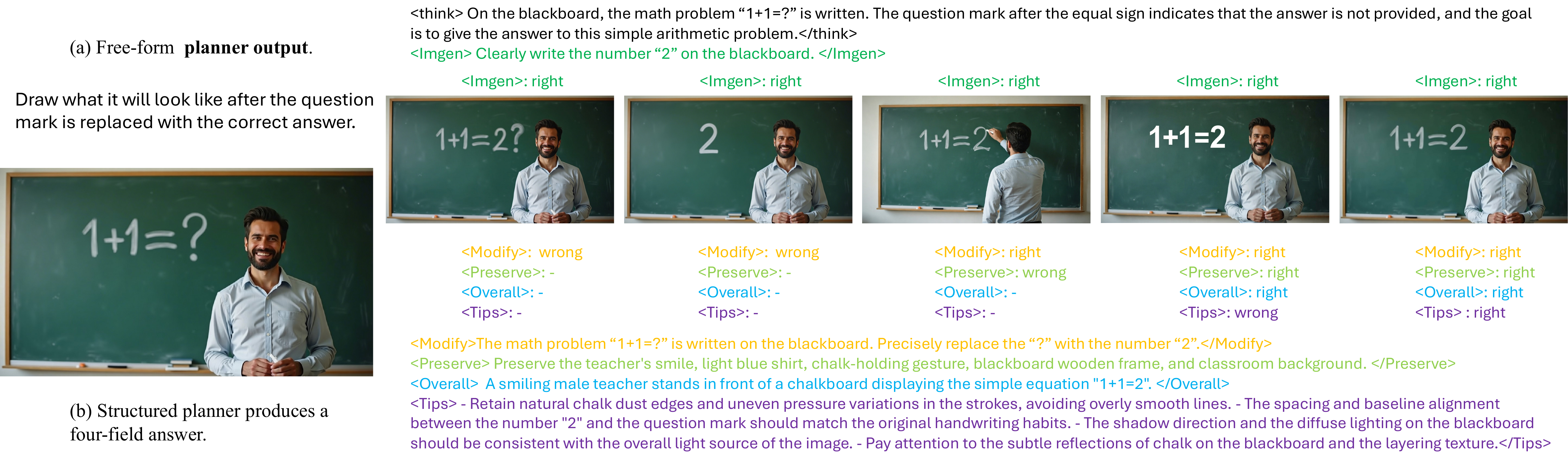}
  \caption{Free-form vs.\ structured planner output. With a free-form planner trace, post-render feedback remains attached to a single opaque text sequence, making within-planner credit assignment difficult. Our four-field structured answer (\emph{Modify}, \emph{Preserve}, \emph{Overall}, \emph{Tips}) exposes semantically distinct slots, so the same rendered rollouts can be diagnosed field by field and converted into slot-wise planner credit signals.}
  \Description{A double-column comparison of free-form planner output and a structured four-field planner answer, showing that structured outputs enable field-wise diagnosis and within-planner credit assignment from the same rendered outcomes.}
  \label{fig:freeform_vs_structured}
\end{figure*}

To address this dual-level credit-assignment problem, we present \method, a joint-RL framework for planner-renderer image editing. We sample multiple plans and multiple renderings per plan, then use the resulting reward statistics in two different ways: between-plan and within-plan variability determine how much extra update weight is routed to the planner or renderer, while mean rewards across rollouts provide a hardness estimate for adaptive curriculum scheduling~\cite{Curriculum_learning}. On the planner side, we replace an opaque free-form trace with a four-field structured reasoning output (\emph{Modify}, \emph{Preserve}, \emph{Overall}, \emph{Tips}), so the same rendered outcomes can support within-planner credit assignment through slot-level rewards and token-level advantage reweighting (Figure~\ref{fig:freeform_vs_structured}).

% \begin{figure}[t]
%   \centering
%   \includegraphics[width=0.92\columnwidth]{fig/planer_demo.png}
%   \caption{Free-form vs.\ structured planner output. Free-form reasoning is opaque to reward decomposition; our four-field format enables slot-wise reward and finer diagnostics.}
%   \Description{Side-by-side comparison of free-form planner output and structured four-slot planner output for the same editing instruction.}
%   \label{fig:freeform_vs_structured}
% \end{figure}

In summary, our contributions are:
\begin{itemize}[leftmargin=1.5em]
  \item We formulate joint RL for planner-renderer image editing as a dual-level credit-assignment problem, spanning cross-module update allocation and within-planner reasoning diagnosis.
  \item We propose a rollout variance decomposition that separates across-plan from within-plan variability for cross-module routing, together with a rollout-reward hardness estimate for adaptive curriculum scheduling.
  \item We introduce a four-field structured reasoning output for the planner together with a prefix-gated reward and token-level advantage reweighting, enabling localized planner optimization from post-render feedback.
\end{itemize}
Experiments on five benchmarks show that \method is especially effective on reasoning-intensive edits~\cite{RISEBench,KRIS-Bench}, and ablations indicate that cross-module routing and structured planner supervision contribute complementary gains.

\section{Related Work}
\label{sec:related_work}

\paragraph{Instruction-based image editing.}
Instruction-based image editing~\cite{Qwen-Image,LongCat-Image,flux2,FLUX1,SDEdit,Prompt-to-Prompt,InstructPix2Pix} aims to modify a source image according to a natural-language instruction while preserving irrelevant content. Recent work increasingly augments editing with explicit reasoning or planning. RePlan~\cite{RePlan} introduces region-aware planning for complex edits, while ReasonEdit~\cite{ReasonEdit} and UniReason~1.0~\cite{UniReason} strengthen editing with richer semantic and world-knowledge reasoning. Step1X-Edit-v1p2~\cite{ReasonEdit}, ThinkRL-Edit~\cite{ThinkRL-Edit}, and EditThinker~\cite{EditThinker} further show that multi-round reflection can improve editing fidelity, and ThinkGen~\cite{ThinkGen} demonstrates the benefit of explicit intermediate reasoning in a coupled reasoning-generation pipeline. Taken together, these methods establish the value of introducing a planner-like stage before generation. Their main emphasis, however, is on designing stronger intermediate reasoning or refinement procedures. Our focus is different: we study how to optimize an already existing planner-renderer pipeline once a structured intermediate stage has been introduced.

\paragraph{RL for two-stage planning-based editing.}
Reinforcement learning has recently been used to improve instruction faithfulness and multimodal alignment in image generation~\cite{Flow-GRPO,DSPO,DiffusionNFT,DanceGRPO} and editing. PromptRL~\cite{PromptRL} studies reward-based optimization for flow-based generation. In two-stage editing pipelines, existing RL methods differ mainly in which module is optimized: ThinkRL-Edit~\cite{ThinkRL-Edit} focuses on the generation side by tuning the diffusion model and text encoder under reflective plans, EditThinker~\cite{EditThinker} concentrates optimization on the VLM reasoning module, and ThinkGen~\cite{ThinkGen} alternates updates between the reasoning model and the generator. Once editing is organized as an explicit prompt-rewriting, planning, or reasoning stage followed by visual execution, the optimization problem becomes inherently two-stage: one module produces an intermediate textual plan, and another realizes it in the image space. In this setting, existing RL strategies largely follow three patterns: (i)~\emph{stage-wise or alternating} optimization, which separates planner and renderer updates; (ii)~\emph{single-module} optimization, which freezes one component and tunes the other; and (iii)~\emph{uniform joint} optimization, which updates both modules from the same outcome-level reward. The first two reduce the opportunity for coordinated co-improvement, while the third leaves the learning signal underspecified because a low final score does not reveal which module should receive more optimization emphasis. Our method is most closely related to the third line of work, but differs in explicitly addressing this module-allocation problem through rollout-derived routing while retaining joint training.

\paragraph{Curriculum learning and structured reasoning.}
Curriculum learning~\cite{Curriculum-DPO,Curriculum-DPO++} improves optimization by ordering or weighting training samples according to difficulty. Standard curriculum designs, however, treat hardness as a one-dimensional property of each sample. In two-stage image editing, this abstraction is incomplete: two samples with similar difficulty may still differ in whether their uncertainty is primarily plan-dominant or render-dominant, suggesting that sample scheduling alone is insufficient without module-aware weighting. On the reasoning side, PromptEnhancer~\cite{PromptEnhancer}, ThinkGen~\cite{ThinkGen}, and EditThinker~\cite{EditThinker} show the value of richer intermediate text, but free-form reasoning traces remain difficult to evaluate locally after rendering. They often depend on engineered cold-start formats or expert-designed prompting recipes~\cite{Decomposed_Prompting}, which makes them harder to scale and optimize consistently under RL. Our method occupies a different point in this design space: instead of relying on long free-form traces, we use a compact four-field schema that remains close to the planner's native generation format while exposing semantically aligned slots for post-render reward decomposition and token-level credit assignment.

\section{Preliminary}
\label{sec:prelim}

\subsection{Problem Setup}
\label{subsec:prelim_problem_setup}

The input is $\mathbf{x} = (I_{\mathrm{src}}, c)$, where $I_{\mathrm{src}}$ is the source image and $c$ is the editing instruction. A VLM \emph{planner} $\planner(\mathbf{e}\mid\mathbf{x})$ with parameters $\theta$ predicts a structured edit plan
\begin{equation}
  \mathbf{e}
  =
  \left(
  e^{\mathrm{mod}},
  e^{\mathrm{pre}},
  e^{\mathrm{ovr}},
  e^{\mathrm{tip}}
  \right),
\end{equation}
specifying modifications, preservation constraints, the scene objective, and execution hints. A conditional image generator (\emph{renderer}) $\renderer(\mathbf{y}\mid I_{\mathrm{src}}, \mathbf{e})$ with parameters $\phi$ produces the image $\mathbf{y}$. The renderer receives the source image and the structured plan; the raw instruction $c$ is not separately passed to it.
For reward computation, Gemini 3 Pro is used once before RL training to generate and cache a slot-aligned checklist $\mathbf{C}(\mathbf{x}) = \left(C^{\mathrm{mod}}, C^{\mathrm{pre}}, C^{\mathrm{ovr}}, C^{\mathrm{tip}}\right)$ from the source image and raw instruction. During RL training, Qwen3-VL-32B performs all online planner-side and renderer-side Yes/No scoring for the sampled rollouts. Each checklist field enumerates verifiable requirements for the corresponding planner slot.

The reward model produces two module-targeted scores from each rendered outcome. The \emph{renderer reward} $R_{\mathrm{rend}}(\mathbf{x}, \mathbf{e}, \mathbf{y}; \mathbf{C})$ measures checklist satisfaction by the edited image. The \emph{planner reward} $R_{\mathrm{plan}}(\mathbf{x}, \mathbf{e}, \mathbf{y}; \mathbf{C})$ evaluates whether each structured slot is semantically appropriate and supported by the realized edit, so plan quality is assessed in the execution context. A shared reward
\begin{equation}
  R_{\mathrm{share}} = \eta_{\mathrm{plan}} R_{\mathrm{plan}} + \eta_{\mathrm{rend}} R_{\mathrm{rend}}
\end{equation}
uses fixed mixing weights $\eta_{\mathrm{plan}}, \eta_{\mathrm{rend}} \ge 0$ for routing and curriculum estimation, while $R_{\mathrm{plan}}$ and $R_{\mathrm{rend}}$ remain the optimization targets. All in-house ablations use the same reward model, so their differences reflect the proposed credit-assignment and planner-structure choices under a fixed reward pipeline. The central challenge is credit assignment: for a given sample, (i)~which update path should receive a stronger learning signal, and (ii)~which parts of the planner output, if any, should be corrected.

\subsection{Policy Optimization Background}
\label{subsec:prelim_grpo}

\textbf{GRPO}~\cite{GRPO} samples $G$ candidate outputs per input and uses group-relative advantages with clipped surrogate and KL regularization. \textbf{Flow-GRPO}~\cite{Flow-GRPO} extends this to diffusion-flow generators~\cite{Rectified_Flow,sd3} by defining the policy ratio at each denoising step. Both are used off-the-shelf; our contribution lies in the rollout-derived signals that weight their updates.

\begin{figure*}[t]
  \centering
  \includegraphics[width=0.95\textwidth]{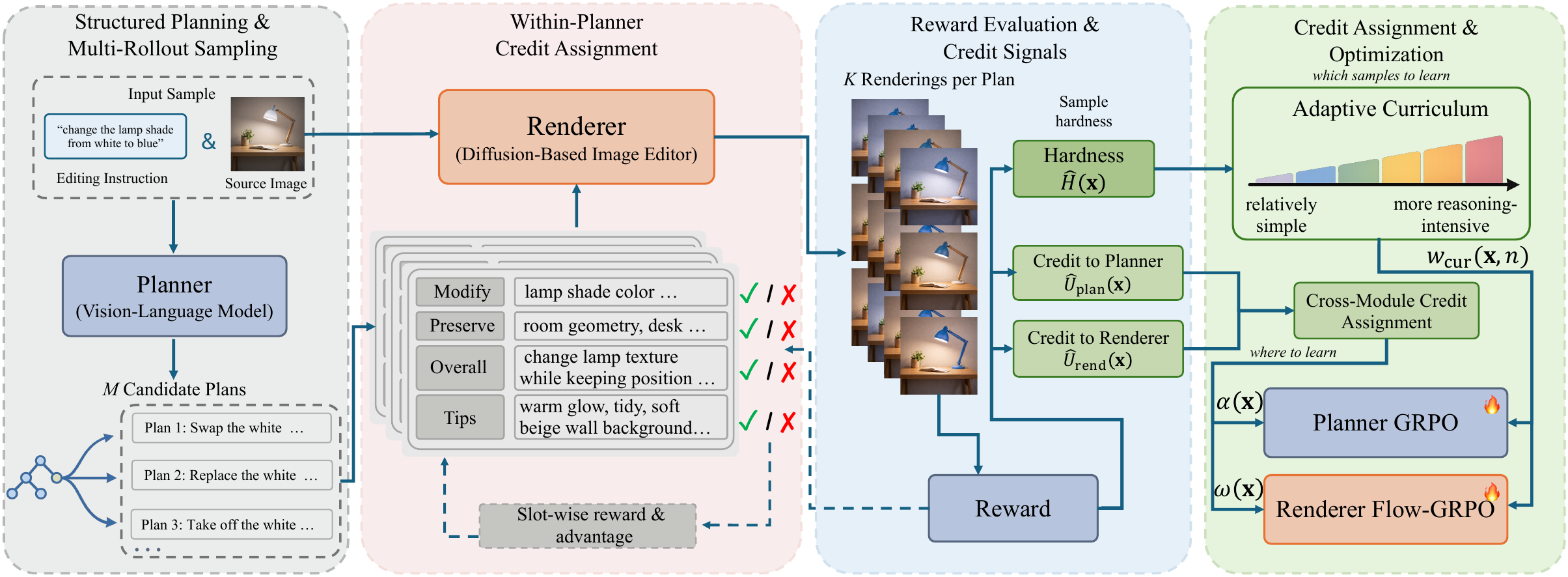}
  \caption{Framework of \method. The structured planner produces a four-field answer; multiple plans and renderings per plan are sampled. Mean rewards across rollouts yield hardness estimates for adaptive curriculum scheduling, while reward variability is decomposed into across-plan and within-plan components to produce module-routing signals for module-specific weighting. Prefix gating further provides slot-wise planner credit signals for finer within-planner diagnosis.}
  \Description{Framework diagram of DARS showing the structured planner, multi-plan multi-render rollouts, rollout variance decomposition, adaptive curriculum, soft module routing, and slot-wise planner optimization.}
  \label{fig:pipeline}
\end{figure*}

% 怀松 和 报哥建议 ：子图 标题放大 ， Planner Update 不要 Update 直接 Planner  GRPO， Renderer-side Difficulty 以及 Planner-side Difficulty 不同颜色 以及 两个Difficulty 和两个 GRPO 上下对应关系要 对应

\section{Method}
\label{sec:method}

\method addresses credit assignment at two levels in two-stage image editing pipelines: across modules and within the planner. The framework contains three coupled components:
\begin{enumerate}[leftmargin=1.5em]
  \item \emph{Structured reasoning and slot-wise planner credit assignment} (Section~\ref{subsec:method_structured_planner}): the planner outputs a four-field answer enabling a prefix-gated reward and finer diagnostics;
  \item \emph{Rollout variance decomposition for cross-module credit assignment} (Section~\ref{subsec:method_difficulty}): multi-plan and multi-render rollouts produce variance-based statistics that distinguish across-plan variability from within-plan rollout variability;
  \item \emph{Adaptive curriculum and soft module routing} (Section~\ref{subsec:method_curriculum_routing}): a rollout-derived hardness score controls sample weighting, while decomposed variability guides how extra update weight is assigned across modules.
\end{enumerate}
The underlying principle is to separate three questions that end-to-end editing RL conflates: \emph{which module should receive extra corrective emphasis} (cross-module credit assignment via module routing), \emph{how planner errors should be diagnosed more finely} (structured reasoning with slot-wise planner credit assignment), and \emph{which samples are currently most informative} (adaptive curriculum). Figure~\ref{fig:pipeline} provides an overview.

\subsection{Structured Reasoning and Slot-Wise Planner Credit Assignment}
\label{subsec:method_structured_planner}

\paragraph{Structured reasoning output.}
Open-ended planner reasoning is difficult to optimize under sparse editing rewards because feedback arrives only after image generation. Following Section~\ref{sec:prelim}, we represent the planner output as a four-field structured answer $\mathbf{e} = \left(e^{\mathrm{mod}}, e^{\mathrm{pre}}, e^{\mathrm{ovr}}, e^{\mathrm{tip}}\right)$. Each field has a task-specific role: $e^{\mathrm{mod}}$ (\emph{Modify}) states what should change; $e^{\mathrm{pre}}$ (\emph{Preserve}) states what must remain unchanged; $e^{\mathrm{ovr}}$ (\emph{Overall}) captures scene-level coherence; $e^{\mathrm{tip}}$ (\emph{Tips}) provides short execution hints for the renderer. Each field remains free-form natural language, preserving the VLM backbone's expressiveness while providing semantic anchoring for reward decomposition. For every training input, a slot-aligned checklist $\mathbf{C}(\mathbf{x})$ serves as the shared evaluation anchor for both module rewards.

\paragraph{Prefix-gated reward.}
With structured planner outputs, feedback can be decomposed at the slot level. For each field $j \in \{\mathrm{mod}, \mathrm{pre}, \mathrm{ovr}, \mathrm{tip}\}$, the reward model inspects the slot text $e^{(j)}$, the source image and instruction, the edited image $\mathbf{y}$, and the corresponding checklist field $C^{(j)}$. Encoding \emph{Yes} as $1$ and \emph{No} as $0$, the slot score $s^{(j)} \in [0,1]$ is the unweighted fraction of Yes judgments over all questions in $C^{(j)}$. It measures semantic adequacy, specificity, and whether the realized edit supports that field as executable guidance. These scores are \emph{planner-side post-render diagnostics}: they assess plan quality through realized renderings, so a textually plausible but unexecutable plan does not receive high reward. Averaging each plan's score over $K$ independent renderings reduces renderer noise.

The four fields admit a natural \emph{causal ordering}: modification intent ($e^{\mathrm{mod}}$) is fundamental; preservation constraints ($e^{\mathrm{pre}}$) are meaningful only if modification is correct; the overall objective ($e^{\mathrm{ovr}}$) synthesizes both; tips ($e^{\mathrm{tip}}$) are useful only when preceding fields are sound. We encode this by prefix gating. For a realized plan--rendering pair $(\mathbf{e}_i,\mathbf{y}_{i,k})$, let $\sigma_{i,k}^{(j)} = \mathbb{I}\bigl[s_{i,k}^{(j)} \ge \delta_j\bigr]$, where $\delta_j$ is the slot threshold and $\lambda_j \ge 0$ weights slot $j$. The per-render planner reward is
\begin{equation}
  R_{\mathrm{plan}}(\mathbf{x}, \mathbf{e}_i, \mathbf{y}_{i,k}; \mathbf{C})
  =
  \sum_{j=1}^{4}
  \lambda_j \cdot s_{i,k}^{(j)} \cdot
  \prod_{l=1}^{j-1} \sigma_{i,k}^{(l)},
  \label{eq:prefix_reward}
\end{equation}
where slots are ordered as $j=1{:}\mathrm{mod},\; 2{:}\mathrm{pre},\; 3{:}\mathrm{ovr},\; 4{:}\mathrm{tip}$. We use the empty-product convention $\prod_{l=1}^{0}(\cdot)=1$, so the Modify term is always ungated. Prefix gating activates later-slot rewards only after earlier slots are correct, encouraging the planner to fix fundamental mistakes first.

The renderer reward focuses on execution quality:
\begin{equation}
  R_{\mathrm{rend}}(\mathbf{x}, \mathbf{e}, \mathbf{y}; \mathbf{C})
  =
  \sum_{j=1}^{4} \gamma_j u^{(j)},
\end{equation}
where $u^{(j)} \in [0,1]$ is analogously the unweighted fraction of renderer-side Yes judgments over all questions in $C^{(j)}$, and $\gamma_j \ge 0$ is the corresponding renderer-side weight. The shared reward is
\begin{equation}
  R_{\mathrm{share}}(\mathbf{x}, \mathbf{e}, \mathbf{y})
  =
  \eta_{\mathrm{plan}} R_{\mathrm{plan}}
  +
  \eta_{\mathrm{rend}} R_{\mathrm{rend}}.
  \label{eq:shared_reward}
\end{equation}

\paragraph{Token-level advantage reweighting.}
For each plan $\mathbf{e}_i$ and rendering $\mathbf{y}_{i,k}$, the reward model returns per-render slot scores $s_{i,k}^{(j)}$, which enter the per-render prefix reward $R_{\mathrm{plan}}$ via Eq.~\ref{eq:prefix_reward}. The plan-level reward $r_i^{\mathrm{plan}}$ averages the resulting per-render rewards across $K$ renderings. For token-level advantage modulation, the slot scores are likewise averaged: $\bar{s}_i^{(j)}=\frac{1}{K}\sum_{k=1}^{K}s_{i,k}^{(j)}$. Tokens in slot $j$ then use
\begin{equation}
  \hat{A}_{i,j}^{\mathrm{plan}}
  =
  \hat{A}_i^{\mathrm{plan}}
  \cdot
  \prod_{l=1}^{j-1} \bar{\sigma}_i^{(l)}
  \cdot
  \begin{cases}
    \bar{s}_i^{(j)}, & \hat{A}_i^{\mathrm{plan}} \ge 0, \\
    1 - \bar{s}_i^{(j)}, & \hat{A}_i^{\mathrm{plan}} < 0,
  \end{cases}
  \label{eq:slot_advantage}
\end{equation}
where $\hat{A}_i^{\mathrm{plan}}$ is the group-relative plan-level advantage and $\bar{\sigma}_i^{(l)} = \mathbb{I}[\bar{s}_i^{(l)} \ge \delta_l]$ is an analogous prefix gate computed from rendering-averaged slot scores. It uses the same threshold form and causal ordering as Eq.~\ref{eq:prefix_reward}, but is distinct from the per-render gates $\sigma_{i,k}^{(l)}$. This sign-aware modulation prevents low-quality slots from inheriting the positive credit of an otherwise good plan, while assigning stronger negative updates to the slots most associated with poor overall plan performance. The prefix gate $\prod_{l<j}\bar{\sigma}_i^{(l)}$ ensures that downstream slots receive gradient only when all upstream slots pass their quality thresholds, maintaining the same causal dependency enforced in the reward (Eq.~\ref{eq:prefix_reward}) through to the token-level update.

\subsection{Rollout Variance Decomposition for Cross-Module Credit Assignment}
\label{subsec:method_difficulty}

For a fixed input $\mathbf{x}$, let $\mathbf{e} \sim \planner(\cdot \mid \mathbf{x})$ and $\mathbf{y} \sim \renderer(\cdot \mid I_{\mathrm{src}}, \mathbf{e})$. We decompose the variance of $R_{\mathrm{share}}$:
\begin{align}
  U_{\mathrm{plan}}(\mathbf{x})
  &=
  \Var_{\mathbf{e}}
  \Bigl(
  \E_{\mathbf{y} \mid I_{\mathrm{src}}, \mathbf{e}}
  [R_{\mathrm{share}}]
  \Bigr), \\
  U_{\mathrm{rend}}(\mathbf{x})
  &=
  \E_{\mathbf{e}}
  \Bigl[
  \Var_{\mathbf{y} \mid I_{\mathrm{src}}, \mathbf{e}}
  (R_{\mathrm{share}})
  \Bigr].
\end{align}
By the law of total variance, $\Var(R_{\mathrm{share}} \mid \mathbf{x}) = U_{\mathrm{plan}}(\mathbf{x}) + U_{\mathrm{rend}}(\mathbf{x})$. The first term measures how much the expected reward changes across plans; the second measures reward fluctuation under a fixed plan. Large across-plan variability suggests that selecting or improving the plan could materially change the outcome; large within-plan variability suggests that outcome quality remains unstable even after the plan is fixed. Since the edited image varies across renderer rollouts and both $R_{\mathrm{plan}}$ and $R_{\mathrm{rend}}$ are derived from that realized image, within-plan variability is induced by renderer stochasticity but can propagate into both module-targeted scores. We use these quantities as rollout-level cross-module credit-assignment signals for \emph{where additional optimization is currently most actionable}; accordingly, $U_{\mathrm{plan}}$ and $U_{\mathrm{rend}}$ are referred to as plan-dominant and render-dominant variability.

For each sample, we draw $M$ plans and $K$ renderings per plan, yielding rewards $r_{i,k}^{\mathrm{share}}$. The per-plan mean is $\bar{r}_i = \frac{1}{K} \sum_{k} r_{i,k}^{\mathrm{share}}$ and the grand mean is $\bar{r} = \frac{1}{M} \sum_{i} \bar{r}_i$. We compute:
\begin{align}
  s_{\mathrm{within}}^2(\mathbf{x})
  &=
  \frac{1}{M}
  \sum_{i=1}^{M}
  \frac{1}{K-1}
  \sum_{k=1}^{K}
  \bigl(r_{i,k}^{\mathrm{share}} - \bar{r}_i\bigr)^2, \\
  s_{\mathrm{between}}^2(\mathbf{x})
  &=
  \frac{1}{M-1}
  \sum_{i=1}^{M}
  \bigl(\bar{r}_i - \bar{r}\bigr)^2.
\end{align}
The within-plan estimate is $\widehat{U}_{\mathrm{rend}} = s_{\mathrm{within}}^2$, and the across-plan estimate corrects for finite-$K$ noise:
\begin{equation}
  \widehat{U}_{\mathrm{plan}}(\mathbf{x})
  =
  \max
  \left(
  0,\;
  s_{\mathrm{between}}^2
  -
  \tfrac{1}{K}
  s_{\mathrm{within}}^2
  \right).
\end{equation}
The $M \times K$ rollouts are shared with policy optimization, so variance estimation requires no extra renderer budget.

\subsection{Adaptive Curriculum and Soft Module Routing}
\label{subsec:method_curriculum_routing}

Hardness determines \emph{when} a sample should contribute strongly; rollout-variance signals determine \emph{where} additional module emphasis should be assigned.

\paragraph{Curriculum scheduling.}
The rollout-based hardness estimate is $\widehat{H}(\mathbf{x}) = -\frac{1}{MK} \sum_{i,k} r_{i,k}^{\mathrm{share}}$ (larger = harder). After normalization, the resulting hardness $\widetilde{H}(\mathbf{x})$ is compared against a step-dependent boundary $\kappa(n)$ that increases with training step $n$ via the empirical quantile of $\widetilde{H}$ in a running buffer. The curriculum weight is
\begin{equation}
  w_{\mathrm{cur}}(\mathbf{x}, n)
  =
  \sigmoid
  \left(
  \frac{\kappa(n) - \widetilde{H}(\mathbf{x})}{\tau_{\mathrm{c}}}
  \right).
\end{equation}
Here $\tau_{\mathrm{c}}$ is a temperature parameter controlling how sharply the curriculum weight transitions around the current boundary $\kappa(n)$. Early in training, simpler samples receive large weights; as the model improves, $\kappa(n)$ rises to progressively expose harder samples.

\paragraph{Cross-module credit assignment.}
We use the relative magnitude of the variance signals:
\begin{equation}
  \alpha(\mathbf{x})
  =
  \frac{\widehat{U}_{\mathrm{plan}} + \varepsilon_U}
  {\widehat{U}_{\mathrm{plan}} + \widehat{U}_{\mathrm{rend}} + 2\varepsilon_U},
  \quad
  \omega(\mathbf{x}) = 1 - \alpha(\mathbf{x}).
\end{equation}
Here $\varepsilon_U > 0$ is the routing smoothing constant. This soft routing acts as a residual reweighting on top of the shared curriculum weight: every sample still updates both modules, while the variance signals control the extra module-specific weight. Curriculum determines \emph{when} a sample receives larger weight; module routing determines \emph{where additional weight is placed}. We validate these routing weights through both agreement with GPT-5 pseudo-labels (Section~\ref{subsec:exp_validation}) and downstream ablation (Section~\ref{subsec:exp_ablation}).

\paragraph{How curriculum and credit assignment interact.}
The two mechanisms play complementary roles over time. Early in training, the curriculum boundary $\kappa(n)$ is low, so easier samples receive most of the weight; for these samples, module routing fine-tunes which module gets emphasis. As training progresses and $\kappa(n)$ rises, harder samples with heterogeneous variance are admitted. Module routing then becomes more important, because these newly admitted samples do not benefit equally from additional updates to both modules. If a sample has high hardness but near-zero total variance, curriculum still controls its admission while module routing contributes little differential emphasis; optimization is then driven mainly by the two task rewards $R_{\mathrm{plan}}$ and $R_{\mathrm{rend}}$.

\begin{table*}[t]
  \caption{Main comparison across five benchmarks. All entries are official \emph{overall} scores (higher is better; compare within columns). PICA-Bench reports ``simple / detailed'' prompt scores. We highlight Qwen-Image-Edit-2511-based methods separately and include a controlled \emph{Joint RL + Adaptive Curriculum} baseline matching \method in backbone, data, reward model, and rollout budget.}
  \label{tab:main_results}
  \centering
  \begin{tabular}{lccccc}
    \toprule
    Method & KRIS-Bench & RISE-Bench & ImgEdit-Bench & GEdit-Bench-EN & PICA-Bench \\
    \midrule
    \multicolumn{6}{l}{\textit{Other baselines}} \\
    \midrule
    ThinkGen~\cite{ThinkGen} & 59.57 & 10.56 & 3.97 & 7.16 & 60.82/65.09 \\
    UniREdit~\cite{UniREdit} & 61.02 & 13.33 & 3.70 & 6.47 & 60.40/66.55 \\
    PromptRL~\cite{PromptRL} & 61.24 & 10.00 & 3.87 & 6.74 & 51.02/60.38 \\
    Step1X-Edit-v1p2~\cite{ReasonEdit} & 62.58 & 11.67 & 4.00 & 7.47 & 56.15/60.55 \\
    UniReason 1.0~\cite{UniReason} & 63.26 & 15.00 & 4.06 & 6.52 & 57.30/61.62 \\
    ThinkRL-Edit~\cite{ThinkRL-Edit} & 66.04 & 14.44 & 4.19 & 6.48 & 55.13/55.84 \\
    PhysicEdit~\cite{PhysicEdit} & 66.78 & 18.89 & 4.05 & 7.37 & 61.48/68.96 \\
    EditThinker~\cite{EditThinker} & 68.80 & 18.33 & 4.31 & 7.59 & 59.64/67.19 \\
    \midrule
    \multicolumn{6}{l}{\textit{Qwen-Image-Edit-2511-based methods}} \\
    \midrule
    Qwen-Image-Edit-2511~\cite{Qwen-Image} & 64.42 & 17.50 & \underline{4.36} & 6.97 & 63.20/\underline{72.27} \\
    RePlan~\cite{RePlan} & 55.72 & 9.72 & 3.44 & 6.43 & 52.00/54.92 \\
    PromptEnhancerV2~\cite{PromptEnhancer} & 71.54 & 18.90 & 4.18 & 7.59 & 61.90/70.77 \\
    Joint RL + Adpt.\ Curriculum (ctrl.) & \underline{72.15} & \underline{25.70} & 4.20 & \underline{7.83} & \underline{63.55}/72.22 \\
    \textbf{\method} & \textbf{80.72} & \textbf{27.50} & \textbf{4.39} & \textbf{7.86} & \textbf{64.19}/\textbf{72.75} \\
    \bottomrule
  \end{tabular}
\end{table*}

\subsection{Joint Optimization}
\label{subsec:method_optimization}

The planner and renderer operate in different output spaces and are optimized with different policy-gradient objectives: text-GRPO~\cite{GRPO} for the planner and flow-GRPO~\cite{Flow-GRPO} for the renderer. The key modification relative to standard GRPO is that each module's objective is weighted by the curriculum factor $w_{\mathrm{cur}}(\mathbf{x}, n)$ plus a residual module-routing emphasis. Every sample contributes a base update to both modules, scaled by its curriculum weight; module routing only adds module-specific weight according to the surrogate decomposition.

For planner optimization, each plan $\mathbf{e}_i$ receives a renderer-averaged reward $r_i^{\mathrm{plan}} = \frac{1}{K} \sum_{k} R_{\mathrm{plan}}(\mathbf{x}, \mathbf{e}_i, \mathbf{y}_{i,k}; \mathbf{C})$, from which $\hat{A}_i^{\mathrm{plan}}$ is computed. Tokens in slot $j$ then use the sign-aware slot-weighted advantage from Eq.~\ref{eq:slot_advantage}: positive plan-level advantages reinforce high-scoring slots, while negative plan-level advantages concentrate corrective updates on low-scoring slots. For renderer optimization, each rendered sample keeps its own reward $r_{i,k}^{\mathrm{rend}} = R_{\mathrm{rend}}(\mathbf{x}, \mathbf{e}_i, \mathbf{y}_{i,k}; \mathbf{C})$. The objectives are:
\begin{align}
  \mathcal{J}_{\mathrm{plan}}(\theta)
  &=
  \E_{\mathbf{x}}
  \Bigl[
  w_{\mathrm{cur}}\,
  \bigl(1 + \rho\,\alpha\bigr)\,
  \mathcal{L}_{\mathrm{GRPO}}^{\mathrm{plan}}(\theta; \mathbf{x})
  \Bigr], \\
  \mathcal{J}_{\mathrm{rend}}(\phi)
  &=
  \E_{\mathbf{x}, \mathbf{e}}
  \Bigl[
  w_{\mathrm{cur}}\,
  \bigl(1 + \rho\,\omega\bigr)\,
  \mathcal{L}_{\mathrm{flow\text{-}GRPO}}^{\mathrm{rend}}(\phi; \mathbf{x}, \mathbf{e})
  \Bigr],
\end{align}
where $\rho \ge 0$ controls the strength of module routing. $\mathcal{L}_{\mathrm{GRPO}}^{\mathrm{plan}}$ is the standard clipped surrogate with KL regularization, applied at the token level using $\hat{A}_{i,j}^{\mathrm{plan}}$ for tokens in slot $j$, and $\mathcal{L}_{\mathrm{flow\text{-}GRPO}}^{\mathrm{rend}}$ uses per-render advantages from $\{r_{i,k}^{\mathrm{rend}}\}$. The final objective is $\mathcal{J} = \mathcal{J}_{\mathrm{plan}} + \mathcal{J}_{\mathrm{rend}}$. The routing weights are estimated from $R_{\mathrm{share}}$, while each module's policy gradient uses its own reward ($R_{\mathrm{plan}}$ or $R_{\mathrm{rend}}$). Module routing, therefore, modulates \emph{how much} each module is updated without changing \emph{what} it learns. Full expanded expressions are in the supplementary.

\section{Experiments}
\label{sec:exp}

\begin{figure*}[t]
  \centering
  \includegraphics[width=0.82\textwidth]{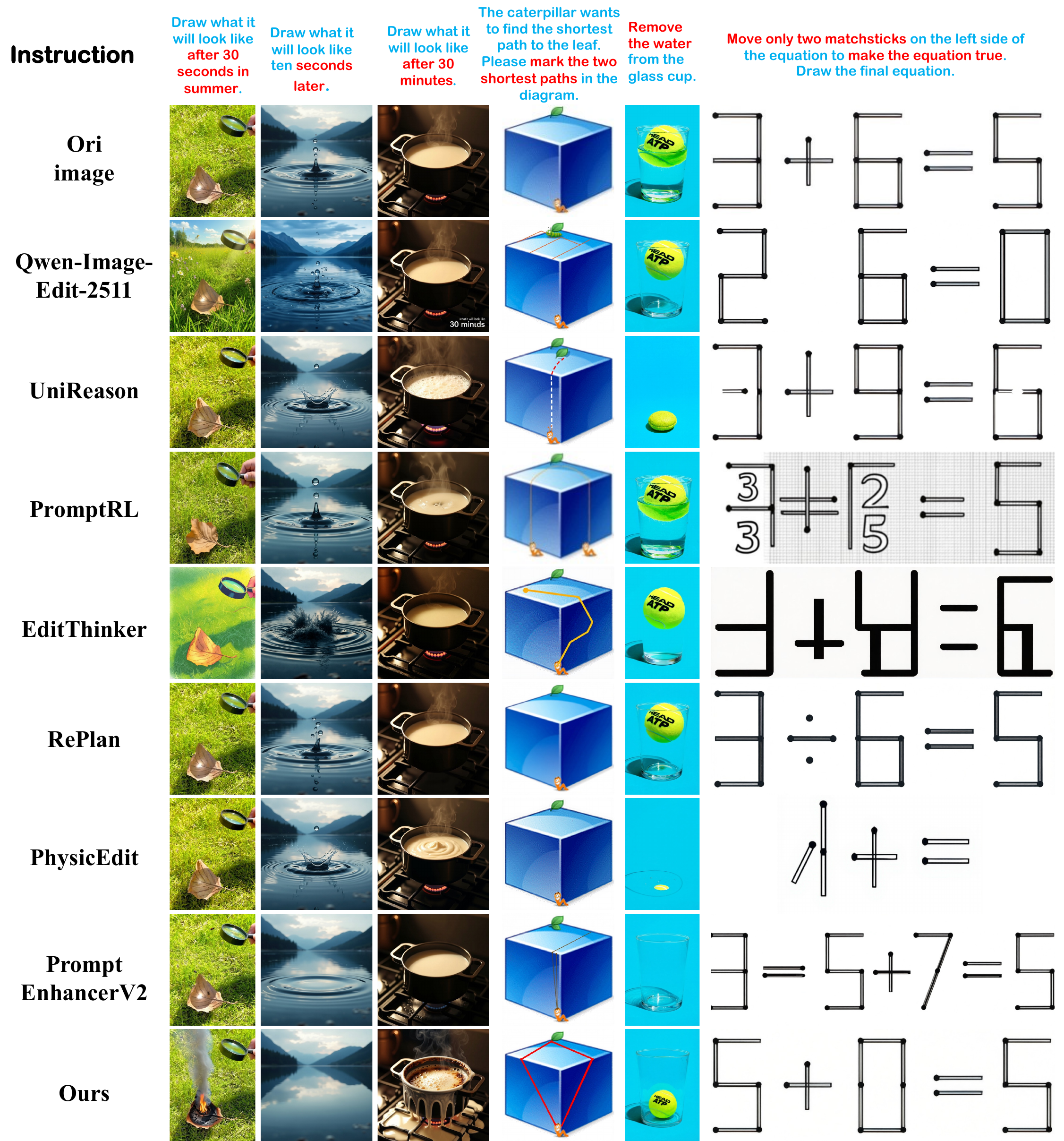}
  \caption{Qualitative comparison on reasoning-intensive editing tasks requiring temporal, spatial, logical, physical, and world-knowledge reasoning, often in combination. Compared with baseline methods, \method achieves the strongest overall performance, producing more faithful edits with better preservation and stronger scene-level consistency.}
  \Description{Qualitative comparison figure for DARS, comparing source images, user instructions, baseline outputs, and DARS outputs across relatively simple, abstract, and physically constrained editing scenarios.}
  \label{fig:qualitative}
\end{figure*}

\subsection{Experimental Setup}
\label{subsec:exp_setup}

\paragraph{Training.}
We train \method on 10K examples (5K from the RL split of THINKEDIT-140K ~\cite{EditThinker} and 5K from UniREdit-Data-100K ~\cite{UniREdit}). The planner is initialized from Qwen3-VL-4B-Instruct, and the renderer from Qwen-Image-Edit-2511. We use $M{=}4$ plans and $K{=}4$ renderings per plan ($M \times K = 16$ rollouts per input). The learning rate is $2\times10^{-6}$, the global batch size is $256$, the KL penalty coefficient~\cite{ppo} is $10^{-3}$, and the maximum image resolution is $1024 \times 1024$. The offline checklist generator (Gemini 3 Pro) and the online reward judge (Qwen3-VL-32B) are fixed across \method and all in-house ablations, so differences reflect the proposed credit-assignment and planner-structure choices under a fixed reward pipeline. The controlled baseline and \method use the same rollout budget on identical hardware; \method adds lightweight rollout statistics and slot-wise scoring on top of that shared budget.
We evaluate on five complementary benchmarks spanning reasoning-intensive editing, general-purpose instruction following, preservation-sensitive localized editing, and physics-aware realism: KRIS-Bench~\cite{KRIS-Bench}, RISE-Bench~\cite{RISEBench}, ImgEdit-Bench~\cite{ImgEdit_bench}, GEdit-Bench-EN~\cite{Step1X-Edit}, and PICA-Bench~\cite{PICABench}. All reported numbers are official overall scores, and the higher the number, the better.
Benchmark results are obtained using each benchmark's official evaluation code and are fully independent of the training reward model.

\paragraph{Baselines.}
We compare against 11 methods spanning multi-round reflection editors (ThinkRL-Edit~\cite{ThinkRL-Edit}, Step1X-Edit-v1p2~\cite{ReasonEdit}, and EditThinker~\cite{EditThinker}), reasoning- or planning-based editors (ThinkGen~\cite{ThinkGen}, RePlan~\cite{RePlan}, UniREdit~\cite{UniREdit}, and UniReason~1.0~\cite{UniReason}, which includes a single reflection step), reward-optimized or prompt-enhancement methods (PromptRL~\cite{PromptRL} and PromptEnhancerV2~\cite{PromptEnhancer}), a physics-specialized baseline (PhysicEdit~\cite{PhysicEdit}), and a strong end-to-end editor (Qwen-Image-Edit-2511~\cite{Qwen-Image}). We further include a controlled \emph{Joint RL + Adaptive Curriculum} baseline that matches \method in backbone, data, reward model, rollout budget, and adaptive curriculum, while replacing \method's structured reasoning planner and dual-level credit assignment with free-form reasoning. Detailed architecture descriptions are in the supplementary.

\subsection{Main Results}
\label{subsec:exp_main_results}

Table~\ref{tab:main_results} shows that the full \method system attains the best score on all five benchmark regimes. Against previously reported methods, the gains are consistent across evaluation criteria. Against the in-house controlled baseline with matched backbone, data, reward model, rollout budget, and adaptive curriculum, \method improves by $+8.57$ on KRIS-Bench, $+1.80$ on RISE-Bench, $+0.19$ on ImgEdit-Bench, $+0.03$ on GEdit-Bench-EN, and $+0.64/+0.53$ on PICA-Bench. The strongest gains appear on KRIS-Bench and RISE-Bench, consistent with \method's focus on planner-renderer credit assignment for reasoning-intensive edits.

\subsection{Qualitative Results}
\label{subsec:exp_qualitative}

Figure~\ref{fig:qualitative} compares \method against reasoning-oriented editing methods. We include Qwen-Image-Edit-2511 as a baseline without explicit reasoning, together with instruction-rewriting or planning baselines: UniReason~1.0 based on the unified multimodal model Bagel, PromptRL as a VLM-diffusion joint-RL method, EditThinker with up to five rounds of reflective prompt revision, RePlan with region-aware reasoning, PhysicEdit specialized by physics-grounded data tuning, and PromptEnhancerV2, a strong 32B prompt-enhancement model combining expert-designed rules with end-to-end RL. Across the Figure~\ref{fig:qualitative} examples, \method delivers the strongest visual quality, achieving the best balance of instruction faithfulness, preservation, scene coherence, and reasoning consistency.

\subsection{Ablation Studies}
\label{subsec:exp_ablation}

Unless otherwise stated, ablations are reported on RISE-Bench and GEdit-Bench-EN. All variants use \method without auxiliary heuristics. 
Free-form joint RL without a curriculum diverged, whereas the structured reasoning planner remained trainable without it, albeit with worse performance. The ablations below, therefore, use the structured reasoning planner unless explicitly labeled otherwise.

\paragraph{Curriculum and routing (Table~\ref{tab:ablation_curriculum_routing}).}
Adaptive curriculum consistently outperforms static and no-curriculum baselines, confirming that dynamic hardness re-estimation is critical as model competence evolves. Notably, No Curriculum still converges under the structured reasoning planner, whereas free-form joint RL without a curriculum diverges , suggesting that the four-field structure provides implicit training stabilization. For routing, soft continuous weighting outperforms both uniform and hard-threshold alternatives, consistent with mixed but asymmetric variance patterns in many editing failures. The configuration comparison shows that the full \method improves over both the prompt-only baseline (23.80, no RL) and the controlled free-form Joint~RL baseline (25.70), confirming gains over structured prompting alone and over a matched free-form joint-RL setup.

\paragraph{Planner structure and reward (Table~\ref{tab:ablation_structure_reward}).}
A monotonic trend runs from free-form reasoning to the full four-field answer, with each field providing gains: removing \emph{Preserve} causes a drop on GEdit-Bench-EN, consistent with its emphasis on localized editing and non-target preservation; removing \emph{Overall} produces a larger drop on the reasoning-heavy RISE-Bench, indicating that scene-level coherence is important for complex edits; removing \emph{Tips} yields the smallest consistent degradation, suggesting that execution guidance is helpful once higher-level semantics are in place. These results support distinct slot roles and slot-wise advantage reweighting unavailable under free-form reasoning. For reward composition, the prefix-gated reward outperforms flat averaging and weighted summation. Flat averaging treats all fields as independent and equally important, failing to reflect the causal structure of editing plans. Weighted summation improves over flat averaging but underperforms because it does not enforce dependency between upstream and downstream fields. The prefix-gated reward yields the best results by ensuring that later fields are only rewarded once earlier fields are reliable, reducing the incentive to compensate for upstream errors by overproducing downstream text.

\begin{table}[t]
  \caption{Ablation on curriculum and routing strategies.}
  \label{tab:ablation_curriculum_routing}
  \centering
  \small
  \begin{tabular}{lcc}
    \toprule
    Variant & RISE & GEdit \\
    \midrule
    \multicolumn{3}{l}{\textit{Curriculum}} \\
    No Curriculum & 18.89 & 6.87 \\
    Static Curriculum & 24.56 & 7.15 \\
    \textbf{Adaptive Curriculum} & \textbf{27.50} & \textbf{7.86} \\
    \midrule
    \multicolumn{3}{l}{\textit{Routing}} \\
    No Routing & 20.05 & 7.12 \\
    Hard Routing & 24.86 & 7.49 \\
    \textbf{Soft Routing (Ours)} & \textbf{27.50} & \textbf{7.86} \\
    \midrule
    \multicolumn{3}{l}{\textit{Configuration}} \\
    Structured Prompt Only (No RL) & 23.80 & 7.10 \\
    Free-form Joint RL + Adpt. & 25.70 & 7.83 \\
    \textbf{Full DARS} & \textbf{27.50} & \textbf{7.86} \\
    \bottomrule
  \end{tabular}
\end{table}

\begin{table}[t]
  \caption{Ablation on planner structure and reward composition.}
  \label{tab:ablation_structure_reward}
  \centering
  \small
  \begin{tabular}{lcc}
    \toprule
    Variant & RISE & GEdit \\
    \midrule
    \multicolumn{3}{l}{\textit{Planner structure}} \\
    Free-form Reasoning & 23.94 & 6.98 \\
    Modify Only & 25.21 & 7.11 \\
    Without Preserve & 26.08 & 7.34 \\
    Without Overall & 26.42 & 7.51 \\
    Without Tips & 26.87 & 7.69 \\
    \textbf{Full Structured Answer} & \textbf{27.50} & \textbf{7.86} \\
    \midrule
    \multicolumn{3}{l}{\textit{Reward composition}} \\
    Flat Average & 24.92 & 7.28 \\
    Weighted Sum & 26.31 & 7.57 \\
    \textbf{Prefix-Gated (Ours)} & \textbf{27.50} & \textbf{7.86} \\
    \bottomrule
  \end{tabular}
\end{table}

\subsection{Validating the Uncertainty Signals}
\label{subsec:exp_validation}

The modeling claim of \method is that rollout statistics provide informative routing signals for cross-module optimization. We validate whether the plan-dominant and render-dominant routing scores align with GPT-5 pseudo-labels of dominant failure patterns.

\paragraph{Protocol.}
For each evaluation case, we construct the same $M \times K$ rollout bank used by \method and compute $\widehat{U}_{\mathrm{plan}}$, $\widehat{U}_{\mathrm{rend}}$, and $\alpha$ from that rollout bank. Separately, we sample five realized triplets from the same case and ask GPT-5 to assign each one a pseudo-label among \emph{planner-dominant}, \emph{renderer-dominant}, or \emph{mixed}; the majority vote becomes the case-level pseudo-label. GPT-5 sees only the source image, structured plan, and edited image for each triplet, not the rollout statistics or reward-model outputs used by \method. Metrics are reported for pairwise one-vs-one discrimination tasks, with AUROC computed in the corresponding binary setting.

\begin{table}[t]
  \caption{Routing-signal discrimination against majority-voted GPT-5 pseudo-labels.}
  \label{tab:attribution}
  \centering
  \begin{tabular}{lccc}
    \toprule
    Setting & Accuracy & Macro-F1 & AUROC \\
    \midrule
    Planner vs.\ Renderer & 86.0 & 85.8 & 0.930 \\
    Planner vs.\ Mixed & 84.7 & 84.1 & 0.918 \\
    Renderer vs.\ Mixed & 85.3 & 84.9 & 0.923 \\
    \bottomrule
  \end{tabular}
\end{table}

Table~\ref{tab:attribution} shows AUROC above 0.91 in all pairwise discriminations, indicating that the rollout statistics track GPT-5 pseudo-labels of dominant failure patterns well. Combined with the ablation result that soft routing outperforms both hard routing and no routing (Table~\ref{tab:ablation_curriculum_routing}), this provides evidence that the routing signal captures failure-mode differences and improves update allocation.

\section{Limitations}
\label{sec:limitations}

\method requires $M \times K$ rollouts per training example, so the total cost remains higher than single-path updates even though these rollouts are shared with policy optimization. The variance-based routing and post-render planner scores depend on the reward model, and planner-renderer interactions can be harder to disentangle when both modules fail simultaneously. Adapting the framework to video editing or multi-turn settings is left for future work.

\section{Conclusion}
\label{sec:conclusion}

We present \method, which formulates joint RL in two-stage planner-renderer image editing pipelines as a dual-level credit-assignment problem. Rollout variance decomposition provides routing signals for cross-module updates; hardness-derived scores drive adaptive curriculum scheduling; and a structured four-field planner with prefix gating enables slot-wise diagnostics. \method attains the best results across five benchmarks, with the largest gains on reasoning-intensive tasks.

\begin{acks}
This work was supported by 
% Guangdong Basic and Applied Basic Research Foundation (Nos. 2024A1515140109 and 2023A1515110695) and 
KlingAI Research.
We thank Shenhui Zhang, Jiahao Guo, and Bao Tang for their valuable discussions.
\end{acks}

\bibliographystyle{ACM-Reference-Format}
\bibliography{main}

% Append the supplementary material in the same document for arXiv.
\clearpage
% Supplementary page numbers restart at 1, so disable duplicate PDF page anchors.
\hypersetup{pageanchor=false}
\setcounter{page}{1}
\setcounter{section}{0}
\setcounter{subsection}{0}
\setcounter{equation}{0}
\setcounter{figure}{0}
\setcounter{table}{0}
\setcounter{algorithm}{0}
\renewcommand{\thesection}{\arabic{section}}
\renewcommand{\thesubsection}{\thesection.\arabic{subsection}}
\renewcommand{\theequation}{S\arabic{equation}}
\renewcommand{\thefigure}{S\arabic{figure}}
\renewcommand{\thetable}{S\arabic{table}}
\renewcommand{\theHsection}{supp.\arabic{section}}
\renewcommand{\theHsubsection}{\theHsection.\arabic{subsection}}
\renewcommand{\theHequation}{supp.\arabic{equation}}
\renewcommand{\theHfigure}{supp.\arabic{figure}}
\renewcommand{\theHtable}{supp.\arabic{table}}
\renewcommand{\theHalgorithm}{supp.\arabic{algorithm}}
\twocolumn[{
  \centering
    {\huge\sffamily\bfseries
    \parbox{0.98\textwidth}{\centering
      Supplementary Material for \method: Dual-Level Credit Assignment\\
      RL with Structured Reasoning for Instruction-Based Image\\
      Editing\par}}
  \par\vspace{2\baselineskip}
}]
\thispagestyle{empty}
\DARSstartsupplementaryindexes
% Render machine-readable examples with literal ASCII quotation marks.
% Using \char34 avoids LaTeX's typographic opening/closing quotes.
\newcommand{\StrictJSONAnswerExample}{%
  {\ttfamily
  \{\\
  \quad\char34 answer\char34: \char34 Yes\char34,\\
  \quad\char34 brief\_reason\char34: \char34 Short factual reason.\char34\\
  \}}%
}
\newcommand{\StrictJSONChecklistExample}{%
  {\ttfamily
  \{\\
  \quad\char34 modify\char34: [\char34 Question 1?\char34, \char34 Question 2?\char34, \char34 Question 3?\char34],\\
  \quad\char34 preserve\char34: [\char34 Question 1?\char34, \char34 Question 2?\char34, \char34 Question 3?\char34],\\
  \quad\char34 overall\char34: [\char34 Question 1?\char34, \char34 Question 2?\char34, \char34 Question 3?\char34],\\
  \quad\char34 tips\char34: [\char34 Question 1?\char34, \char34 Question 2?\char34, \char34 Question 3?\char34]\\
  \}}%
}

\section*{0. Contents and Index}
\addcontentsline{toc}{section}{0. Contents and Index}
\setcounter{tocdepth}{2}
\tableofcontents

\bigskip
\listoffigures

\bigskip
\listoftables

\clearpage

\section{Additional Methodological Details}

\subsection{Expanded Optimization Objectives}
\label{subsec:supp_objectives}

This subsection expands the compact objectives in the main paper into the exact rollout-level forms used during optimization. Definitions of the reward signals and the high-level design motivation remain in the main text; here we focus only on how the offline cached checklist, online planner-generated four-slot plans, old-policy rollouts, and token- or step-level surrogates are instantiated in training. Throughout this subsection, $i$ indexes sampled plans, $k$ renderings, $j$ structured slots, $t$ planner tokens, and $m$ denoising-trajectory positions.

\paragraph{Planner objective.}
For each input $\mathbf{x}$, we sample a group of $G_{\mathrm{p}} = M$ candidate plans from the behavior policy:
\begin{equation}
  \mathbf{e}_i \sim q_{\theta_{\mathrm{old}}}(\cdot \mid \mathbf{x}),
  \qquad i = 1, \ldots, G_{\mathrm{p}}.
\end{equation}
For each sampled plan $\mathbf{e}_i$, we generate $K$ paired renderings $\mathbf{y}_{i,k} \sim p_{\phi_{\mathrm{old}}}(\cdot \mid I_{\mathrm{src}}, \mathbf{e}_i)$. The planner reward for the $i$-th sampled plan is the planner reward averaged over its rendered outcomes:
\begin{equation}
  r_i^{\mathrm{plan}}
  =
  \frac{1}{K}\sum_{k=1}^{K} R_{\mathrm{plan}}(\mathbf{x}, \mathbf{e}_i, \mathbf{y}_{i,k}; \mathbf{C}),
\end{equation}
where the checklist $\mathbf{C} = \mathbf{C}(\mathbf{x})$ is pre-generated once offline before RL training from the source image and instruction, and then shared across all rollouts of the same sample.
The group-relative advantage is
\begin{equation}
  \hat{A}_i^{\mathrm{plan}}
  =
  \frac{
  r_i^{\mathrm{plan}} -
  \operatorname{mean}\left(\{r_{i'}^{\mathrm{plan}}\}_{i'=1}^{G_{\mathrm{p}}}\right)
  }{
  \operatorname{std}\left(\{r_{i'}^{\mathrm{plan}}\}_{i'=1}^{G_{\mathrm{p}}}\right) + \varepsilon_A
  }.
\end{equation}
Here $\varepsilon_A>0$ is the advantage-normalization stabilizer shared by the planner and renderer objectives.
Let $s_{i,k}^{(j)}$ denote the score of slot $j$ under rendering $\mathbf{y}_{i,k}$, and define the rendering-averaged slot score and reweighted token-level advantage as
\begin{equation}
  \bar{s}_i^{(j)} = \tfrac{1}{K} \sum_{k=1}^{K} s_{i,k}^{(j)},
  \qquad
  \hat{A}_{i,j}^{\mathrm{plan}}
  =
  \hat{A}_i^{\mathrm{plan}}
  \cdot
  \prod_{l=1}^{j-1} \bar{\sigma}_i^{(l)}
  \cdot
  \begin{cases}
    \bar{s}_i^{(j)}, & \hat{A}_i^{\mathrm{plan}} \ge 0, \\
    1 - \bar{s}_i^{(j)}, & \hat{A}_i^{\mathrm{plan}} < 0,
  \end{cases}
\end{equation}
where $\bar{\sigma}_i^{(l)} = \mathbb{I}[\bar{s}_i^{(l)} \ge \delta_l]$ is an analogous prefix gate computed from rendering-averaged slot scores. It has the same threshold form and causal ordering as the per-render gate in the main paper, but is distinct from $\sigma_{i,k}^{(l)}$. This sign-aware modulation prevents low-quality slots from inheriting positive credit from an otherwise good plan, while assigning stronger negative updates to the slots most associated with poor overall plan performance.
Since the planner is autoregressive, the GRPO ratio is defined at the token level. For the $t$-th token in $\mathbf{e}_i$:
\begin{equation}
  \varrho_{i,t}^{\mathrm{plan}}(\theta)
  =
  \frac{
  q_{\theta}(e_{i,t} \mid \mathbf{x}, e_{i,<t})
  }{
  q_{\theta_{\mathrm{old}}}(e_{i,t} \mid \mathbf{x}, e_{i,<t})
  }.
\end{equation}
Let $\operatorname{slot}(t)$ denote the structured slot containing token $t$, as determined by the enclosing ordered tag pair; the opening and closing tag tokens are assigned to that same slot. The token-level surrogate uses the sign-aware slot-weighted advantage $\hat{A}_{i,\operatorname{slot}(t)}^{\mathrm{plan}}$ instead of a uniform plan-level advantage. The full planner objective is:
\begin{equation}
  \begin{aligned}
    \mathcal{J}_{\mathrm{plan}}(\theta)
    &=
    \E_{\mathbf{x}}
    \Biggl[
    w_{\mathrm{cur}}(\mathbf{x}, n)\,
    \bigl(1 + \rho\,\alpha(\mathbf{x})\bigr)\,
    \frac{1}{G_{\mathrm{p}}}
    \sum_{i=1}^{G_{\mathrm{p}}}
    \frac{1}{|\mathbf{e}_i|}
    \sum_{t=1}^{|\mathbf{e}_i|}
    \\
    &\qquad
    \Bigl(
    \min
    \bigl(
    \varrho_{i,t}^{\mathrm{plan}}(\theta)\,\hat{A}_{i,\operatorname{slot}(t)}^{\mathrm{plan}},
    \\
    &\qquad\qquad
    \operatorname{clip}\bigl(
    \varrho_{i,t}^{\mathrm{plan}}(\theta), 1-\epsilon_{\mathrm{p}}, 1+\epsilon_{\mathrm{p}}
    \bigr)\,\hat{A}_{i,\operatorname{slot}(t)}^{\mathrm{plan}}
    \bigr)
    \\
    &\qquad
    -
    \beta_{\mathrm{p}}\,D_{\mathrm{KL}}(q_{\theta} \,\|\, q_{\mathrm{ref}})
    \Bigr)
    \Biggr].
  \end{aligned}
\end{equation}

\paragraph{Renderer objective.}
Using the same sampled plans $\{\mathbf{e}_i\}_{i=1}^{M}$ from the planner rollout bank, we sample $K$ renderer rollouts for each fixed pair $(\mathbf{x}, \mathbf{e}_i)$ from the old diffusion-flow policy. Let $\mathbf{y}_{i,k}$ denote the decoded image from the $k$-th rollout under plan $\mathbf{e}_i$. The renderer-side reward is
\begin{equation}
  r_{i,k}^{\mathrm{rend}} = R_{\mathrm{rend}}(\mathbf{x}, \mathbf{e}_i, \mathbf{y}_{i,k}; \mathbf{C}),
\end{equation}
where again $\mathbf{C} = \mathbf{C}(\mathbf{x})$ is fixed for all rollouts of the same input.
The renderer-side group-relative advantage is computed separately within each fixed plan:
\begin{equation}
  \hat{A}_{i,k}^{\mathrm{rend}}
  =
  \frac{
  r_{i,k}^{\mathrm{rend}} -
  \operatorname{mean}\left(\{r_{i,k'}^{\mathrm{rend}}\}_{k'=1}^{K}\right)
  }{
  \operatorname{std}\left(\{r_{i,k'}^{\mathrm{rend}}\}_{k'=1}^{K}\right) + \varepsilon_A
  }.
\end{equation}
For flow-GRPO, let $m$ index the complete ordered denoising trajectory, $\tau_m$ its diffusion time, and $z_{i,k,m}$ the corresponding latent state. Here $\tau_{m-1}$ is the adjacent lower-noise time after $\tau_m$, with latent state $z_{i,k,m-1}$; the optimized indices form a subset $\mathcal{S}$ of this trajectory. The ratio is
\begin{equation}
  \varrho_{i,k,m}^{\mathrm{flow}}(\phi)
  =
  \frac{
  p_{\phi}(z_{i,k,m-1} \mid I_{\mathrm{src}}, \mathbf{e}_i, z_{i,k,m}, \tau_m)
  }{
  p_{\phi_{\mathrm{old}}}(z_{i,k,m-1} \mid I_{\mathrm{src}}, \mathbf{e}_i, z_{i,k,m}, \tau_m)
  }.
\end{equation}
The full renderer objective is:
\begin{equation}
  \begin{aligned}
    \mathcal{J}_{\mathrm{rend}}(\phi)
    &=
    \E_{\mathbf{x}}
    \Biggl[
    w_{\mathrm{cur}}(\mathbf{x}, n)\,
    \bigl(1 + \rho\,\omega(\mathbf{x})\bigr)\,
    \frac{1}{M}
    \sum_{i=1}^{M}
    \frac{1}{K}
    \sum_{k=1}^{K}
    \frac{1}{|\mathcal{S}|}
    \sum_{m \in \mathcal{S}}
    \\
    &\qquad
    \Bigl(
    \min
    \bigl(
    \varrho_{i,k,m}^{\mathrm{flow}}(\phi)\,\hat{A}_{i,k}^{\mathrm{rend}},
    \operatorname{clip}(\varrho_{i,k,m}^{\mathrm{flow}}(\phi), 1-\epsilon_{\mathrm{r}}, 1+\epsilon_{\mathrm{r}})
    \,\hat{A}_{i,k}^{\mathrm{rend}}
    \bigr)
    \\
    &\qquad\qquad
    -
    \beta_{\mathrm{r}}\,D_{\mathrm{KL}}(p_{\phi} \,\|\, p_{\mathrm{ref}})
    \Bigr)
    \Biggr].
  \end{aligned}
\end{equation}

\subsection{Variance-Based Cross-Module Credit Assignment}
\label{subsec:supp_variance}

The main paper introduces the plan-dominant and render-dominant variability terms. Here, we record only the estimator used in training. For a fixed input $\mathbf{x} = (I_{\mathrm{src}}, c)$, let $\mathbf{e} \sim \planner(\cdot \mid \mathbf{x})$ and $\mathbf{y} \sim \renderer(\cdot \mid I_{\mathrm{src}}, \mathbf{e})$. We decompose the shared-reward variance as
\begin{align}
  U_{\mathrm{plan}}(\mathbf{x})
  &=
  \Var_{\mathbf{e}}
  \Bigl(
  \E_{\mathbf{y} \mid I_{\mathrm{src}}, \mathbf{e}}
  [R_{\mathrm{share}}]
  \Bigr), \\
  U_{\mathrm{rend}}(\mathbf{x})
  &=
  \E_{\mathbf{e}}
  \Bigl[
  \Var_{\mathbf{y} \mid I_{\mathrm{src}}, \mathbf{e}}
  (R_{\mathrm{share}})
  \Bigr].
\end{align}
By the law of total variance,
\begin{equation}
  \Var(R_{\mathrm{share}} \mid \mathbf{x})
  =
  U_{\mathrm{plan}}(\mathbf{x}) + U_{\mathrm{rend}}(\mathbf{x}).
\end{equation}
For each training sample, the same $M \times K$ rollout bank already used for reward computation is reused to estimate these two terms. Let $r_{i,k}^{\mathrm{share}}$ be the shared reward of the $k$-th rendering under the $i$-th sampled plan, $\bar{r}_i = \frac{1}{K}\sum_{k=1}^{K} r_{i,k}^{\mathrm{share}}$ the per-plan mean, and $\bar{r} = \frac{1}{M}\sum_{i=1}^{M} \bar{r}_i$ the grand mean. We compute
\begin{align}
  s_{\mathrm{within}}^2(\mathbf{x})
  &=
  \frac{1}{M}
  \sum_{i=1}^{M}
  \frac{1}{K-1}
  \sum_{k=1}^{K}
  \bigl(r_{i,k}^{\mathrm{share}} - \bar{r}_i\bigr)^2, \\
  s_{\mathrm{between}}^2(\mathbf{x})
  &=
  \frac{1}{M-1}
  \sum_{i=1}^{M}
  \bigl(\bar{r}_i - \bar{r}\bigr)^2.
\end{align}
The within-plan Monte Carlo estimate is
\begin{equation}
  \widehat{U}_{\mathrm{rend}}(\mathbf{x}) = s_{\mathrm{within}}^2(\mathbf{x}),
\end{equation}
and the across-plan estimate subtracts the renderer-noise term induced by averaging only $K$ renderings per plan:
\begin{equation}
  \widehat{U}_{\mathrm{plan}}(\mathbf{x})
  =
  \max
  \left(
  0,\;
  s_{\mathrm{between}}^2(\mathbf{x}) - \tfrac{1}{K}s_{\mathrm{within}}^2(\mathbf{x})
  \right).
\end{equation}
The $\max(0,\cdot)$ truncation avoids negative across-plan estimates caused by finite-sample noise. No additional renderer calls are introduced: the same rollout bank feeds reward estimation, curriculum weighting, routing, and policy optimization.

\subsection{Adaptive Curriculum and Soft Module Routing}
\label{subsec:supp_curriculum_routing}

This subsection records the rollout statistics used to instantiate the main-paper curriculum and routing definitions as per-sample training weights.

\paragraph{Curriculum scheduling.}
The rollout-based hardness estimate is
\begin{equation}
  \widehat{H}(\mathbf{x})
  =
  -\frac{1}{MK}\sum_{i=1}^{M}\sum_{k=1}^{K} r_{i,k}^{\mathrm{share}},
\end{equation}
where larger values indicate harder samples. In implementation, $\widehat{H}(\mathbf{x})$ is normalized to $\widetilde{H}(\mathbf{x})$ using the running hardness statistics maintained over recent training batches, and the curriculum boundary $\kappa(n)$ is taken as an empirical quantile of that same buffer at step $n$. The curriculum weight is
\begin{equation}
  w_{\mathrm{cur}}(\mathbf{x}, n)
  =
  \sigmoid
  \left(
  \frac{\kappa(n) - \widetilde{H}(\mathbf{x})}{\tau_{\mathrm{c}}}
  \right).
\end{equation}
Here $\tau_{\mathrm{c}}$ controls how sharply the weight changes around the current boundary. This makes the curriculum update depend only on rollout hardness, not on which module is currently more variable.

\paragraph{Soft module routing.}
We use the relative magnitude of the variance signals to define
\begin{equation}
  \alpha(\mathbf{x})
  =
  \frac{\widehat{U}_{\mathrm{plan}}(\mathbf{x}) + \varepsilon_U}
  {\widehat{U}_{\mathrm{plan}}(\mathbf{x}) + \widehat{U}_{\mathrm{rend}}(\mathbf{x}) + 2\varepsilon_U},
  \qquad
  \omega(\mathbf{x}) = 1 - \alpha(\mathbf{x}),
\end{equation}
where $\varepsilon_U > 0$ is the routing smoothing constant. Routing is applied only as a residual multiplier on top of the shared curriculum factor, so every admitted sample still updates both modules. The variance decomposition only decides where the extra module-specific emphasis should go; it is not used as a hard assignment rule.

\subsection{Overall Training Procedure}
\label{subsec:supp_algorithm}

Algorithm~\ref{alg:supp_training} summarizes one training iteration of \method. A single shared $M \times K$ rollout bank is reused throughout the iteration: it first produces module-specific rewards, then yields planner-side and renderer-side advantages, and finally supplies the variance, routing, and curriculum statistics used to weight the two objectives. This reuse is important operationally because it keeps checklist loading, reward estimation, and module-specific optimization aligned to the same sampled evidence without introducing extra renderer calls beyond the rollout budget already used for policy optimization.

\begin{algorithm*}[t]
\caption{\method Training Algorithm}
\label{alg:supp_training}
\small
\begin{algorithmic}[1]
\Require Planner policy $\planner$, renderer policy $\renderer$, reference policies $q_{\mathrm{ref}}, p_{\mathrm{ref}}$, objective hyperparameters $\epsilon_{\mathrm{p}}, \epsilon_{\mathrm{r}}, \varepsilon_A, \varepsilon_U, \beta_{\mathrm{p}}, \beta_{\mathrm{r}}, \rho$, dataset $\mathcal{D}$, rollout counts $M, K$, training step $n$
\Ensure Updated planner parameters $\theta$ and renderer parameters $\phi$
\State Refresh old policies: $q_{\theta_{\mathrm{old}}} \gets q_{\theta}$, $p_{\phi_{\mathrm{old}}} \gets p_{\phi}$
\State Sample minibatch $\mathcal{B} \sim \mathcal{D}$
\For{each input $\mathbf{x} \in \mathcal{B}$}
  \State Load pre-generated shared checklist $\mathbf{C}(\mathbf{x})$
  \For{$i = 1$ to $M$}
    \State Sample plan $\mathbf{e}_i \sim q_{\theta_{\mathrm{old}}}(\cdot \mid \mathbf{x})$
    \For{$k = 1$ to $K$}
      \State Sample rendering $\mathbf{y}_{i,k} \sim p_{\phi_{\mathrm{old}}}(\cdot \mid I_{\mathrm{src}}, \mathbf{e}_i)$
      \State Evaluate $R_{\mathrm{plan}}(\mathbf{x}, \mathbf{e}_i, \mathbf{y}_{i,k}; \mathbf{C})$, $R_{\mathrm{rend}}(\mathbf{x}, \mathbf{e}_i, \mathbf{y}_{i,k}; \mathbf{C})$, and $R_{\mathrm{share}}(\mathbf{x}, \mathbf{e}_i, \mathbf{y}_{i,k})$
    \EndFor
  \EndFor
  \State Compute planner-side rewards $\{r_i^{\mathrm{plan}}\}_{i=1}^{M}$ and group-relative advantages $\{\hat{A}_i^{\mathrm{plan}}\}_{i=1}^{M}$
  \State Compute rendering-averaged slot scores $\bar{s}_i^{(j)}=\frac{1}{K}\sum_{k=1}^{K}s_{i,k}^{(j)}$ and prefix gates $\bar{\sigma}_i^{(j)}=\mathbb{I}[\bar{s}_i^{(j)}\ge\delta_j]$ for all sampled plans and slots
  \State Compute sign-aware slot-weighted token advantages $\hat{A}_{i,j}^{\mathrm{plan}}$ from $\hat{A}_i^{\mathrm{plan}}$, $\bar{s}_i^{(j)}$, and $\bar{\sigma}_i^{(j)}$
  \For{$i = 1$ to $M$}
    \State Compute renderer-side rewards $\{r_{i,k}^{\mathrm{rend}}\}_{k=1}^{K}$ and group-relative advantages $\{\hat{A}_{i,k}^{\mathrm{rend}}\}_{k=1}^{K}$ over the $K$ renderings under the fixed plan $\mathbf{e}_i$
  \EndFor
  \State Estimate $\widehat{U}_{\mathrm{plan}}(\mathbf{x})$, $\widehat{U}_{\mathrm{rend}}(\mathbf{x})$, and raw hardness $\widehat{H}(\mathbf{x})$ from the same $M \times K$ rollout bank
  \State Update running hardness statistics, normalize $\widehat{H}(\mathbf{x})$ to $\widetilde{H}(\mathbf{x})$, and obtain the curriculum boundary $\kappa(n)$ from the running buffer
  \State Compute routing weights $\alpha(\mathbf{x})$, $\omega(\mathbf{x})$ and curriculum weight $w_{\mathrm{cur}}(\mathbf{x}, n)=\sigmoid((\kappa(n)-\widetilde{H}(\mathbf{x}))/\tau_{\mathrm{c}})$
  \State Accumulate the sample contributions to $\mathcal{J}_{\mathrm{plan}}(\theta)$ and $\mathcal{J}_{\mathrm{rend}}(\phi)$ without resampling
\EndFor
\State Update $\theta$ by text-GRPO using $\mathcal{J}_{\mathrm{plan}}(\theta)$
\State Update $\phi$ by flow-GRPO using $\mathcal{J}_{\mathrm{rend}}(\phi)$
\end{algorithmic}
\end{algorithm*}

\section{Structured Planning and Reward Design}

\subsection{Structured Planner Output Format}
\label{subsec:supp_slot_roles}

The main paper already motivates the four-slot planner decomposition and the associated prefix-gated dependency structure. Here we record the operational slot semantics used throughout the supplementary examples and prompt templates: \emph{Modify} specifies the requested visible change, \emph{Preserve} specifies the important content that must remain unchanged, \emph{Overall} specifies scene-level coherence requirements, and \emph{Tips} provides localized renderer-facing execution details. The online planner serializes these free-form fields in the fixed order
\begin{center}
\texttt{<Modify>...</Modify>}\quad \texttt{<Preserve>...</Preserve>}\\
\texttt{<Overall>...</Overall>}\quad \texttt{<Tips>...</Tips>}.
\end{center}
The matching tag boundaries determine the token-to-slot mapping $\operatorname{slot}(t)$, with each tag pair assigned to its enclosed slot. This is the same slot schema used by the online planner output, the offline checklist cache, the reward prompts, and the qualitative plan table.

\subsection{Shared Checklist Construction and Slot Alignment}
\label{subsec:supp_checklist}

The reward pipeline uses a single slot-aligned checklist shared by planner reward, renderer reward, hardness estimation, and routing. The implementation has three distinct stages. First, during rollout, the current planner rewrites the raw instruction into a four-slot plan $\mathbf{e}$; Fig.~\ref{fig:supp_rewrite_prompt} shows this planner-side prompt. Second, before RL starts, Gemini 3 Pro preprocesses the training set once and directly generates a cached checklist $\mathbf{C}(\mathbf{x})$ from the source image and raw user instruction; Fig.~\ref{fig:supp_checklist_prompt} shows this offline checklist-generation prompt. Third, during RL training, Qwen3-VL-32B performs online Yes/No scoring by pairing the current rollout plan, the cached checklist, and the realized edited image. Gemini 3 Pro is not called at rollout time.

Operationally, Gemini 3 Pro directly produces the shared checklist
\(
\mathbf{C}(\mathbf{x}) =
\left(C^{\mathrm{mod}}, C^{\mathrm{pre}}, C^{\mathrm{ovr}}, C^{\mathrm{tip}}\right)
\)
from the source image and raw instruction, with a question list for each slot role. This preprocessing is run once before RL training, so the same cached $\mathbf{C}(\mathbf{x})$ is reused across all $M \times K$ rollouts of a sample. At rollout time, the current planner separately produces its own four-slot plan in the same slot schema. The checklist, therefore, defines the shared evaluation target, while the rollout-time plan provides the instance-specific slot text being judged.

Each checklist field is not copied from the planner output. Instead, Gemini 3 Pro maps the source image and raw instruction directly into $3$--$5$ binary verification questions per slot. The conversion follows four implementation constraints: slot exclusivity, so each question belongs to exactly one slot; cross-slot complementarity, so the same requirement is not repeated across slots; no cross-slot leakage, so one slot does not silently test another slot's responsibility; and post-render verifiability, so every question can later be answered from the source image, the relevant slot text, and the realized edited image. In practice, $C^{\mathrm{mod}}$ targets the requested visible change, $C^{\mathrm{pre}}$ targets unchanged content, $C^{\mathrm{ovr}}$ targets global coherence and realism, and $C^{\mathrm{tip}}$ targets localized implementation details.

This slot alignment allows the same checklist to support both module-specific rewards without changing the semantic target. During RL training, Gemini 3 Pro is no longer used. Qwen3-VL-32B performs all online checklist-based scoring. On the planner side, it judges whether each rollout-time slot text was a useful plan for the observed edit under the matching checklist field. On the renderer side, it judges whether the final rendered image actually satisfies that same checklist field. The shared checklist, therefore, ties planner-side diagnosis and renderer-side execution scoring to the same pre-generated per-slot questions.

The offline checklist generator also follows conservative default rules when the instruction under-specifies the edit. If the instruction is local, preservation questions default to protecting identity, pose, background, viewpoint, and other non-target content unless the instruction explicitly asks to modify them. If the instruction requires a background change, that requested change is moved into the Modify or Overall field and is not redundantly enforced as a preservation constraint. Text-editing requests add exact rendered-text correctness to the relevant slot, while removal or inpainting requests add plausibility and local consistency checks so that filled regions remain compatible with surrounding texture, lighting, and geometry.

The concrete prompt templates for these three stages are collected later in the template bank: planner rollout rewriting in Fig.~\ref{fig:supp_rewrite_prompt}, offline checklist generation in Fig.~\ref{fig:supp_checklist_prompt}, and online scoring prompts in Figs.~\ref{fig:supp_plan_modify_prompt}--\ref{fig:supp_rend_tips_prompt}.

\subsection{Reward Instantiation in Practice}
\label{subsec:supp_reward}

\paragraph{Question-level aggregation.}
For input $\mathbf{x}=(I_{\mathrm{src}},c)$, plan $\mathbf{e}_i$, rendering $\mathbf{y}_{i,k}$, slot $j$, and question $q \in C^{(j)}$, let $a_{i,k,q}^{\mathrm{plan}} \in \{0,1\}$ denote the planner-side judgment and define $a_{i,k,q}^{\mathrm{rend}}=\mathbb{I}[\operatorname{Judge}_{\mathrm{rend}}(I_{\mathrm{src}},\mathbf{y}_{i,k},c,e_i^{(j)},q)=\mathrm{Yes}]$. We encode \emph{Yes} as $1$ and \emph{No} as $0$. All questions within a slot are weighted equally, giving the per-render slot scores
\begin{equation}
  s_{i,k}^{(j)}
  =
  \frac{1}{|C^{(j)}|}
  \sum_{q \in C^{(j)}} a_{i,k,q}^{\mathrm{plan}},
  \qquad
  u_{i,k}^{(j)}
  =
  \frac{1}{|C^{(j)}|}
  \sum_{q \in C^{(j)}} a_{i,k,q}^{\mathrm{rend}}.
  \label{eq:supp_question_aggregation}
\end{equation}
Thus, $s_{i,k}^{(j)}$ and $u_{i,k}^{(j)}$ are normalized Yes fractions in $[0,1]$, rather than additional continuous outputs from the judge.

\paragraph{Planner-side diagnostic.}
At scoring time, the shared checklist has already been generated offline by Gemini 3 Pro and loaded from cache, while the current planner policy has produced the rollout-time four-slot plan $\mathbf{e}_i$. Qwen3-VL-32B evaluates every question in each slot through single-question Yes/No judgments conditioned on the slot text, the original input, the realized edited image $\mathbf{y}_{i,k}$, and the aligned checklist field $C^{(j)}$. The resulting $s_{i,k}^{(j)}$ is a post-render diagnostic of whether that slot functioned as a useful plan for the realized edit. The corresponding prompt templates are shown in Figs.~\ref{fig:supp_plan_modify_prompt}--\ref{fig:supp_plan_tips_prompt}. Define the per-render gate $\sigma_{i,k}^{(j)} = \mathbb{I}[s_{i,k}^{(j)} \ge \delta_j]$. The per-render planner reward is
\begin{equation}
  R_{\mathrm{plan}}(\mathbf{x}, \mathbf{e}_i, \mathbf{y}_{i,k}; \mathbf{C})
  =
  \sum_{j=1}^{4}
  \lambda_j \cdot s_{i,k}^{(j)} \cdot
  \prod_{l=1}^{j-1} \sigma_{i,k}^{(l)}.
  \label{eq:supp_per_render_plan_reward}
\end{equation}
We use $(\lambda_{\mathrm{mod}}, \lambda_{\mathrm{pre}}, \lambda_{\mathrm{ovr}}, \lambda_{\mathrm{tip}})=(0.4,0.2,0.2,0.2)$ and $\delta_j=0.66$ for every slot. After evaluating all $K$ renderings under plan $\mathbf{e}_i$, we compute
\begin{equation}
  r_i^{\mathrm{plan}}
  =
  \frac{1}{K}\sum_{k=1}^{K}
  R_{\mathrm{plan}}(\mathbf{x}, \mathbf{e}_i, \mathbf{y}_{i,k}; \mathbf{C}),
  \qquad
  \bar{s}_i^{(j)}
  =
  \frac{1}{K}\sum_{k=1}^{K}s_{i,k}^{(j)}.
  \label{eq:supp_plan_render_average}
\end{equation}
The first quantity yields the plan-level group-relative advantage, while the second and its analogous averaged-score gate $\bar{\sigma}_i^{(j)}=\mathbb{I}[\bar{s}_i^{(j)}\ge\delta_j]$ are used for the sign-aware token-level advantage in Section~\ref{subsec:supp_objectives}.

\paragraph{Renderer-side execution scoring.}
The renderer reward changes the target of judgment while retaining the same checklist fields. For each $j$ and $q \in C^{(j)}$, Qwen3-VL-32B is conditioned on $(I_{\mathrm{src}},\mathbf{y}_{i,k},c,e_i^{(j)},q)$. The instruction and slot text specify the intended requirement; the judge scores whether the rendered image visibly satisfies it, not the quality of the slot text. The resulting $u_{i,k}^{(j)}$ is the fraction of satisfied requirements in that slot. The corresponding prompt templates are shown in Figs.~\ref{fig:supp_rend_modify_prompt}--\ref{fig:supp_rend_tips_prompt}. The per-render renderer reward is
\begin{equation}
  r_{i,k}^{\mathrm{rend}}
  =
  R_{\mathrm{rend}}(\mathbf{x}, \mathbf{e}_i, \mathbf{y}_{i,k}; \mathbf{C})
  =
  \sum_{j=1}^{4} \gamma_j u_{i,k}^{(j)}.
  \label{eq:supp_per_render_renderer_reward}
\end{equation}
Here, $\gamma_j$ controls the contribution of each slot to the final execution score. The $K$ renderer rewards under each fixed plan remain separate when computing the renderer-side group-relative advantage; they are not averaged before normalization.

\paragraph{Shared reward reuse.}
The shared reward used for hardness estimation and variance decomposition is
\begin{equation}
  R_{\mathrm{share}}(\mathbf{x}, \mathbf{e}_i, \mathbf{y}_{i,k})
  =
  \eta_{\mathrm{plan}} R_{\mathrm{plan}}(\mathbf{x}, \mathbf{e}_i, \mathbf{y}_{i,k}; \mathbf{C})
  +
  \eta_{\mathrm{rend}} R_{\mathrm{rend}}(\mathbf{x}, \mathbf{e}_i, \mathbf{y}_{i,k}; \mathbf{C}).
\end{equation}
Its role is restricted to hardness estimation and variance decomposition; planner and renderer policy gradients still use their own module-specific rewards.

\begin{figure*}[p]
  \centering
  \setlength{\fboxsep}{5pt}
  \newcommand{\prefixpanelimgsizeA}{4.5cm}
  \newcommand{\prefixpanelimgsizeCE}{4cm}
  \newcommand{\prefixbottompanelheight}{11.5cm}
  \definecolor{prefixfail}{RGB}{196,40,27}
  \definecolor{prefixpass}{RGB}{18,146,37}
  \definecolor{prefixblocked}{RGB}{110,110,110}
  \begin{minipage}[t]{0.48\textwidth}
    \fbox{%
      \parbox[c][7.6cm][t]{0.95\linewidth}{%
        \raggedright
        \textbf{Panel A. Input image $\mathbf{x}$ and raw instruction}\\[0.35em]
        \centering
        \includegraphics[width=\prefixpanelimgsizeA,height=\prefixpanelimgsizeA]{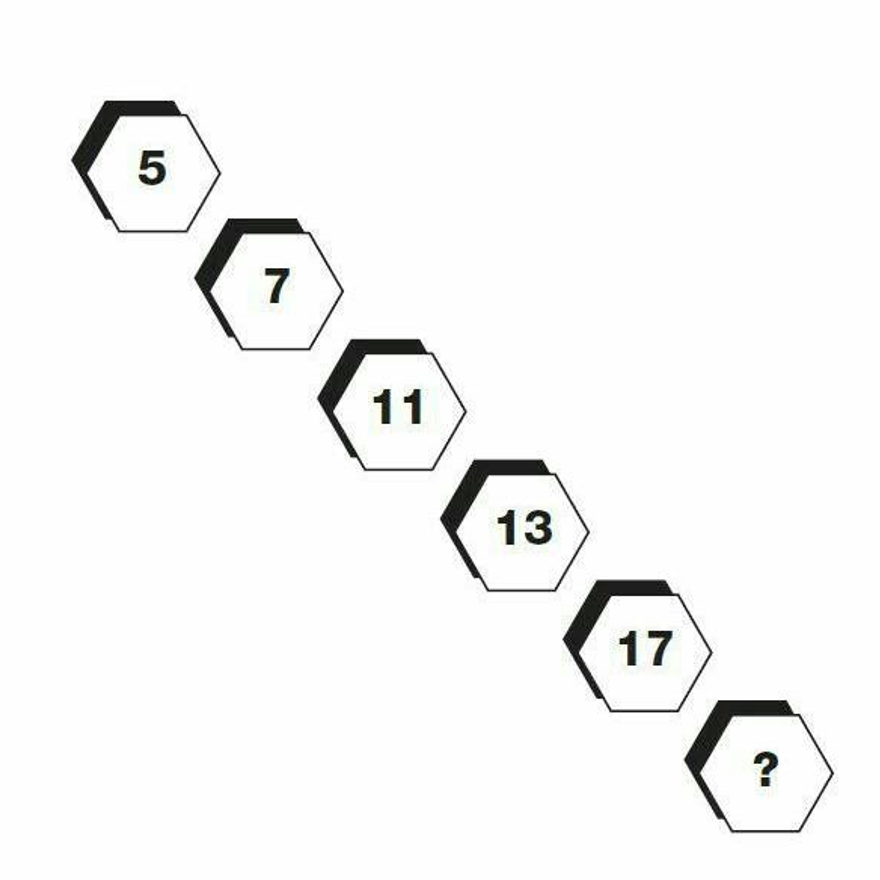}\\[0.45em]
        \raggedright
        \textbf{Raw instruction $c$:} replace the final ``?'' with the correct next number, while keeping the existing numbers, hexagons, and diagonal layout unchanged.\\[0.35em]
        \textbf{Ground-truth target:} the sequence follows prime numbers, so the correct completion is ``19''.
      }%
    }
  \end{minipage}
  \hfill
  \begin{minipage}[t]{0.48\textwidth}
    \fbox{%
      \parbox[c][7.6cm][t]{0.95\linewidth}{%
        \raggedright
        \textbf{Panel B. Checklist $\mathbf{C}(\mathbf{x})$ and plan $\mathbf{e}$}\\[0.25em]
        {\footnotesize
        \begin{minipage}[t]{0.48\linewidth}
          \textbf{Modify $C^{\mathrm{mod}}$}\\
          1. Is the final ``?'' replaced with a numeral?\\
          2. Is the new numeral exactly ``19''?\\
          3. Does the edit correctly continue the prime pattern after 17?\\
          4. Is the target of the change the last hexagon?\\[0.35em]
          \textbf{Preserve $C^{\mathrm{pre}}$}\\
          1. Are 5 and 7 unchanged?\\
          2. Is 11 unchanged?\\
          3. Is 13 unchanged?\\
          4. Are all non-target hexagons otherwise untouched?\\
          5. Does the background remain white?
        \end{minipage}
        \hfill
        \begin{minipage}[t]{0.48\linewidth}
          \textbf{Overall $C^{\mathrm{ovr}}$}\\
          1. Is the six-hexagon diagonal layout preserved?\\
          2. Are size, spacing, and viewpoint consistent?\\
          3. Does the final sequence remain globally coherent?\\
          4. Is the result clean and artifact-free?\\[0.35em]
          \textbf{Tips $C^{\mathrm{tip}}$}\\
          1. Is only the last hexagon edited?\\
          2. Does the new digit match the original font style?\\
          3. Does the stroke/outline style remain consistent?\\
          4. Is the new digit centered with similar spacing?
        \end{minipage}}\\[0.35em]
        \textbf{Structured plan $\mathbf{e}$:} $\mathbf{e} = (e^{\mathrm{mod}}, e^{\mathrm{pre}}, e^{\mathrm{ovr}}, e^{\mathrm{tip}})$, where the four slots are Modify, Preserve, Overall, and Tips.\par\vspace{0.3em}
        \textbf{Scoring rule:} slot score $s^{(j)} = \#\mathrm{Yes}/|C^{(j)}|$ for $j \in \{\mathrm{mod}, \mathrm{pre}, \mathrm{ovr}, \mathrm{tip}\}$.\par\vspace{0.3em}
        \textbf{Gate rule:} slot gate $\sigma_j = \mathbb{I}[s^{(j)} \ge \delta_j]$ with $\delta_{\mathrm{mod}}=\delta_{\mathrm{pre}}=\delta_{\mathrm{ovr}}=\delta_{\mathrm{tip}}=0.66$.
      }%
    }
  \end{minipage}

  \vspace{0.7em}

  \begin{minipage}[t]{0.31\textwidth}
    \fbox{%
      \parbox[c][\prefixbottompanelheight][t]{0.94\linewidth}{%
        \raggedright
        \textbf{Panel C. Failure A: wrong Modify prompt}\\[0.25em]
        \centering
        \includegraphics[width=\prefixpanelimgsizeCE,height=\prefixpanelimgsizeCE]{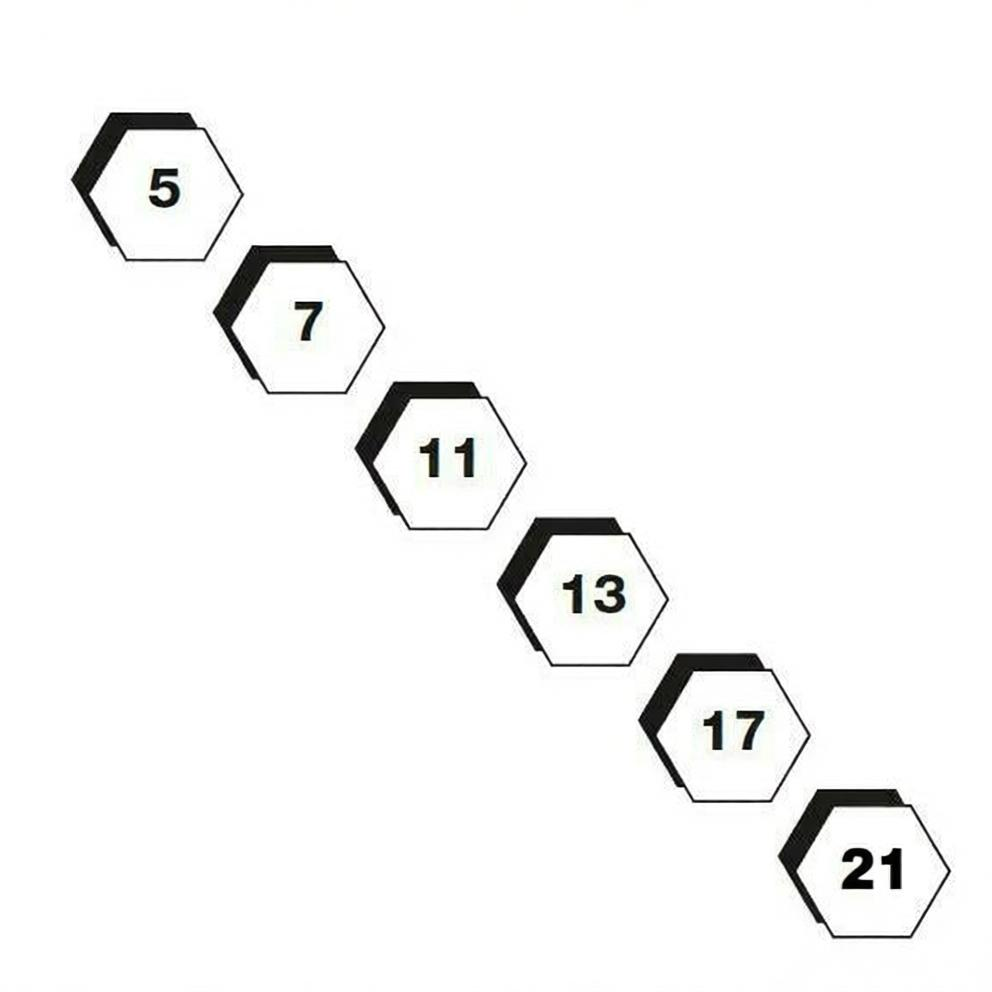}\\[0.3em]
        {\footnotesize\raggedright
        \textbf{Modify $e^{\mathrm{mod}}$:} Put 21 in the last hexagon to continue the sequence.\\
        \textbf{Preserve $e^{\mathrm{pre}}$:} Keep the first five numbered hexagons and the white background unchanged.\\
        \textbf{Overall $e^{\mathrm{ovr}}$:} Preserve the diagonal layout and a smooth increasing pattern.\\
        \textbf{Tips $e^{\mathrm{tip}}$:} Edit only the last hexagon and match the numeral style.\par\vspace{2\baselineskip}
        \textbf{Checklist hits:} Modify $C^{\mathrm{mod}}: 2/4$; Preserve $C^{\mathrm{pre}}: 5/5$.\\
        Overall $C^{\mathrm{ovr}}: 3/4$; Tips $C^{\mathrm{tip}}: 4/4$.\par\vspace{0.3em}
        \textbf{Scores:} Modify $s^{\mathrm{mod}}=0.50$; Preserve $s^{\mathrm{pre}}=1.00$.\\
        Overall $s^{\mathrm{ovr}}=0.75$; Tips $s^{\mathrm{tip}}=1.00$.\par\vspace{0.3em}
        \textbf{Gate:} \textcolor{prefixfail}{Modify gate $\sigma_{\mathrm{mod}}=0$} $\Rightarrow$ \textcolor{prefixblocked}{Preserve, Overall, and Tips blocked}.\par\vspace{0.3em}
        \textbf{Reward:} $R_{\mathrm{plan}}^{(\mathrm{A})}=0.20$.}
      }%
    }
  \end{minipage}
  \hfill
  \begin{minipage}[t]{0.31\textwidth}
    \fbox{%
      \parbox[c][\prefixbottompanelheight][t]{0.94\linewidth}{%
        \raggedright
        \textbf{Panel D. Failure B: weak Preserve prompt}\\[0.25em]
        \centering
        \includegraphics[width=\prefixpanelimgsizeCE,height=\prefixpanelimgsizeCE]{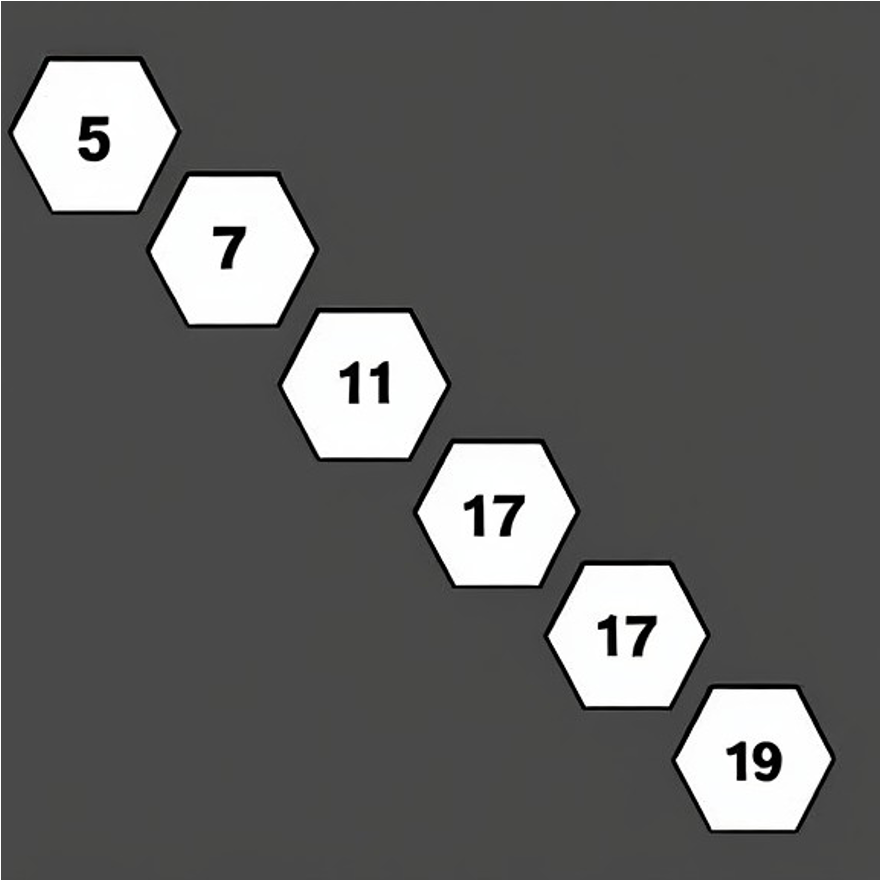}\\[0.3em]
        {\footnotesize\raggedright
        \textbf{Modify $e^{\mathrm{mod}}$:} Put 19 in the last hexagon as the next prime after 17.\\
        \textbf{Preserve $e^{\mathrm{pre}}$:} Keep the sequence readable, but nearby numerals or canvas tones may be adjusted if needed.\\
        \textbf{Overall $e^{\mathrm{ovr}}$:} Preserve the diagonal layout and a plausible completed sequence.\\
        \textbf{Tips $e^{\mathrm{tip}}$:} Harmonize the final region with neighboring hexagons so the ending looks locally consistent.\par\vspace{2\baselineskip}
        \textbf{Checklist hits:} Modify $C^{\mathrm{mod}}: 4/4$; Preserve $C^{\mathrm{pre}}: 2/5$.\\
        Overall $C^{\mathrm{ovr}}: 3/4$; Tips $C^{\mathrm{tip}}: 3/4$.\par\vspace{0.3em}
        \textbf{Scores:} Modify $s^{\mathrm{mod}}=1.00$; Preserve $s^{\mathrm{pre}}=0.40$.\\
        Overall $s^{\mathrm{ovr}}=0.75$; Tips $s^{\mathrm{tip}}=0.75$.\par\vspace{0.3em}
        \textbf{Gate:} \textcolor{prefixpass}{Modify gate $\sigma_{\mathrm{mod}}=1$}, \textcolor{prefixfail}{Preserve gate $\sigma_{\mathrm{pre}}=0$} $\Rightarrow$ \textcolor{prefixblocked}{Overall and Tips blocked}.\par\vspace{0.3em}
        \textbf{Reward:} $R_{\mathrm{plan}}^{(\mathrm{B})}=0.48$.}
      }%
    }
  \end{minipage}
  \hfill
  \begin{minipage}[t]{0.31\textwidth}
    \fbox{%
      \parbox[c][\prefixbottompanelheight][t]{0.94\linewidth}{%
        \raggedright
        \textbf{Panel E. Success: precise four-slot prompt}\\[0.25em]
        \centering
        \includegraphics[width=\prefixpanelimgsizeCE,height=\prefixpanelimgsizeCE]{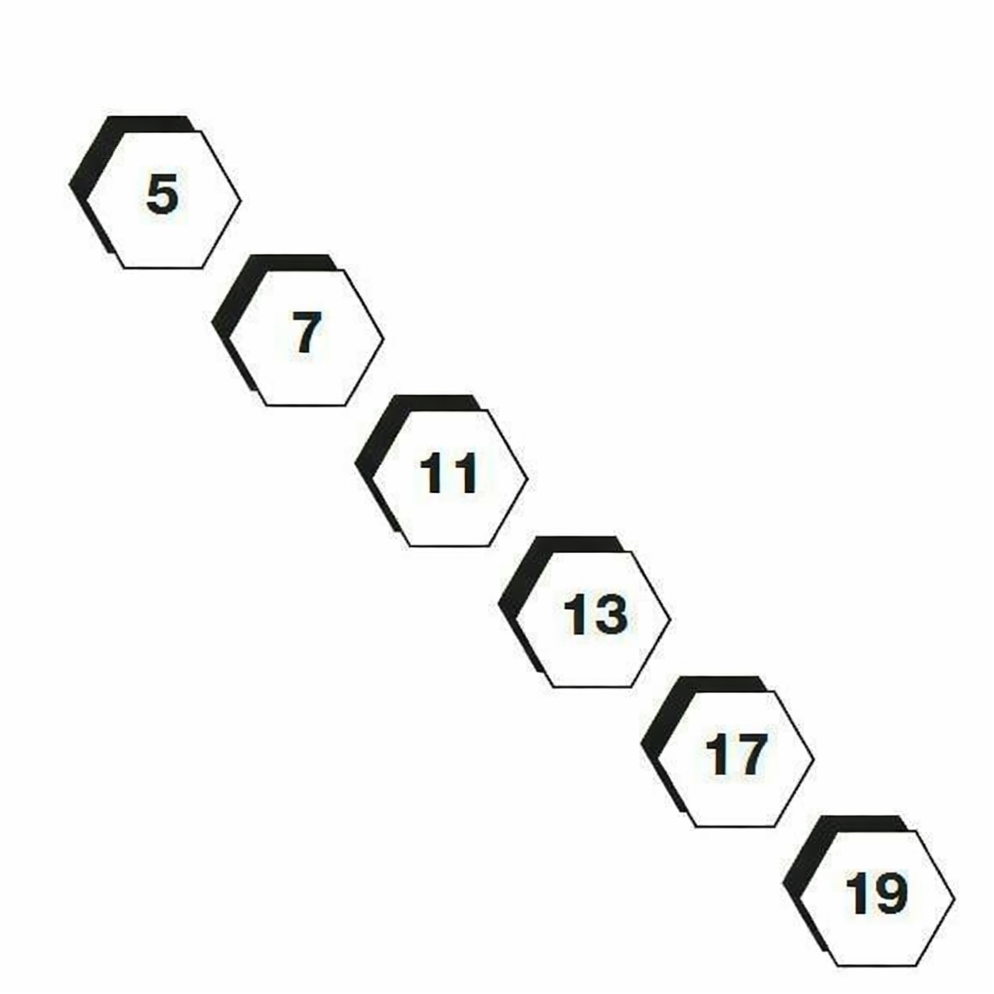}\\[0.3em]
        {\footnotesize\raggedright
        \textbf{Modify $e^{\mathrm{mod}}$:} Put 19 in the last hexagon as the next prime after 17.\\
        \textbf{Preserve $e^{\mathrm{pre}}$:} Keep 5, 7, 11, 13, 17, and the white background unchanged.\\
        \textbf{Overall $e^{\mathrm{ovr}}$:} Preserve the prime progression, clean spacing, and original diagonal layout.\\
        \textbf{Tips $e^{\mathrm{tip}}$:} Edit only the last hexagon and match the font, stroke, and centering.\par\vspace{2\baselineskip}
        \textbf{Checklist hits:} Modify $C^{\mathrm{mod}}: 4/4$; Preserve $C^{\mathrm{pre}}: 5/5$.\\
        Overall $C^{\mathrm{ovr}}: 4/4$; Tips $C^{\mathrm{tip}}: 4/4$.\par\vspace{0.3em}
        \textbf{Scores:} Modify $s^{\mathrm{mod}}=1.00$; Preserve $s^{\mathrm{pre}}=1.00$.\\
        Overall $s^{\mathrm{ovr}}=1.00$; Tips $s^{\mathrm{tip}}=1.00$.\par\vspace{0.3em}
        \textbf{Gate:} \textcolor{prefixpass}{Modify, Preserve, and Overall gates all pass} $(\sigma_{\mathrm{mod}}=\sigma_{\mathrm{pre}}=\sigma_{\mathrm{ovr}}=1)$.\par\vspace{0.3em}
        \textbf{Reward:} $R_{\mathrm{plan}}^{(\mathrm{success})}=1.00$.}
      }%
    }
  \end{minipage}
  \caption[Five-panel schematic of checklist-based prefix-gated planner reward on a sequence-completion edit.]{\textbf{Five-panel schematic of checklist-based prefix-gated planner reward on a sequence-completion edit.} \normalfont Panels~A--B define the shared input, checklist, and structured plan; Panels~C--E illustrate Failure~A, Failure~B, and Success under the same scoring and gating rules. Detailed interpretation is provided in Sec.~\ref{subsec:supp_prefix_example}.}
  \Description{A five-panel schematic for prefix-gated planner reward on a sequence-completion editing example. The first panel shows the source sequence and raw instruction. The second panel lists the modify, preserve, overall, and tips checklists. The remaining three panels show simulated Failure A, Failure B, and Success prompts, their edited sequences, checklist hit counts, scores, gate outcomes, and final rewards.}
  \label{fig:supp_prefix_gating_case}
\end{figure*}

\subsection{Illustrative Prefix-Gated Reward Example}
\label{subsec:supp_prefix_example}

We use a simple sequence-completion edit to illustrate how prefix gating changes planner-side credit assignment in practice. The source image $I_{\mathrm{src}}$ contains a diagonal chain of hexagons labeled ``5, 7, 11, 13, 17, ?'', and the raw instruction $c$ is: ``Replace the final question mark with the correct next number, while keeping the existing numbers, hexagons, and diagonal layout unchanged.'' Figure~\ref{fig:supp_prefix_gating_case} instantiates the same notation used in the main paper on this concrete case: Panels~A--B define the input $\mathbf{x}=(I_{\mathrm{src}},c)$, the structured planner output $\mathbf{e} = (e^{\mathrm{mod}}, e^{\mathrm{pre}}, e^{\mathrm{ovr}}, e^{\mathrm{tip}})$, and the shared checklist $\mathbf{C}(\mathbf{x}) = \left(C^{\mathrm{mod}}, C^{\mathrm{pre}}, C^{\mathrm{ovr}}, C^{\mathrm{tip}}\right)$; Panels~C--E then show three realized planner outcomes, namely Failure~A, Failure~B, and Success.

Consistent with the training definition in Eq.~\ref{eq:supp_question_aggregation}, each slot score is computed from its own checklist as
\(
s^{(j)} = \#\mathrm{Yes}/|C^{(j)}|,
\)
with normalized slot weights
\(
(\lambda_{\mathrm{mod}}, \lambda_{\mathrm{pre}}, \lambda_{\mathrm{ovr}}, \lambda_{\mathrm{tip}})
= (0.4, 0.2, 0.2, 0.2)
\)
and a uniform activation threshold
\(
\delta_{\mathrm{mod}} = \delta_{\mathrm{pre}} = \delta_{\mathrm{ovr}} = \delta_{\mathrm{tip}} = 0.66.
\)
The planner-side reward for a realized outcome is then
\begin{equation}
  R_{\mathrm{plan}}
  =
  \lambda_{\mathrm{mod}} s^{\mathrm{mod}}
  +
  \lambda_{\mathrm{pre}} s^{\mathrm{pre}} \sigma_{\mathrm{mod}}
  +
  \lambda_{\mathrm{ovr}} s^{\mathrm{ovr}} \sigma_{\mathrm{mod}} \sigma_{\mathrm{pre}}
  +
  \lambda_{\mathrm{tip}} s^{\mathrm{tip}} \sigma_{\mathrm{mod}} \sigma_{\mathrm{pre}} \sigma_{\mathrm{ovr}},
\end{equation}

where $\sigma_j = \mathbb{I}[s^{(j)} \ge \delta_j]$. The question-level aggregation, slot weights, activation threshold, and prefix-gating rule are identical to those used in training.

Failure~A in Panel~C shows the clearest advantage of prefix gating. The enhanced plan is locally neat in Preserve, Overall, and Tips, but the core Modify slot is wrong because $e^{\mathrm{mod}}$ proposes 21 instead of 19. The corresponding Modify score is therefore
\(
s^{\mathrm{mod}} = 2/4 = 0.50 < 0.66
\),
so
\(
\sigma_{\mathrm{mod}} = 0
\).
As a result, the prefix gate closes immediately after Modify, and the planner reward becomes
\begin{equation}
  \begin{aligned}
    R_{\mathrm{plan}}^{(\mathrm{A})}
    &=
    0.4 \cdot 0.50
    + 0.2 \cdot 1.00 \cdot 0
    + 0.2 \cdot 0.75 \cdot 0
    + 0.2 \cdot 1.00 \cdot 0 \\
    &= 0.20.
  \end{aligned}
\end{equation}
Even though the rendered result still satisfies all five Preserve checks and most downstream checks, those later slots receive no additional credit once the causally prior Modify requirement fails. By contrast, an ungated weighted sum would still produce 0.75 because the strong downstream slots would partially compensate for the incorrect core reasoning.

Failure~B in Panel~D illustrates a different error pattern. Here the Modify slot is correct, so
\(
s^{\mathrm{mod}} = 4/4 = 1.00
\)
and
\(
\sigma_{\mathrm{mod}} = 1
\).
However, the weak Preserve slot $e^{\mathrm{pre}}$ allows collateral changes near the target, and the rendered result changes the original ``13'' into ``17'' while also darkening the background. This drops the Preserve score to
\(
s^{\mathrm{pre}} = 2/5 = 0.40 < 0.66
\),
so
\(
\sigma_{\mathrm{pre}} = 0
\).
Thus, the reward still keeps the correct Modify credit, but blocks Overall and Tips from contributing:
\begin{equation}
  \begin{aligned}
    R_{\mathrm{plan}}^{(\mathrm{B})}
    &=
    0.4 \cdot 1.00
    + 0.2 \cdot 0.40
    + 0.2 \cdot 0.75 \cdot 0
    + 0.2 \cdot 0.75 \cdot 0 \\
    &= 0.48.
  \end{aligned}
\end{equation}
Again, the ungated weighted sum would be much higher (0.78) because the later slots would still receive credit despite the failed preservation constraint.

Finally, Panel~E shows the successful case. Here, all four slot texts align with the intended edit, and the realized image satisfies every checklist item across Modify, Preserve, Overall, and Tips. Therefore, all gates remain active, and the final planner reward is maximal:
\begin{equation}
  \begin{aligned}
    R_{\mathrm{plan}}^{(\mathrm{success})}
    &=
    0.4 \cdot 1.00
    + 0.2 \cdot 1.00
    + 0.2 \cdot 1.00
    + 0.2 \cdot 1.00 \\
    &= 1.00.
  \end{aligned}
\end{equation}
Taken together, the three cases in Fig.~\ref{fig:supp_prefix_gating_case} make the intended credit-assignment behavior explicit. Prefix gating does not simply average slot quality; instead, it follows the causal dependency structure of the four-slot plan. Modify must first specify the correct edit target, Preserve must then protect non-target content, and only after those upstream requirements are reliable, do Overall and Tips receive effective reward. This is precisely the behavior that the prefix-gated reward is designed to enforce during planner optimization.

% \FloatBarrier
\section{Experimental Setup Details}

\subsection{Training Setup and Hyperparameters}
\label{subsec:supp_training_details}
\label{subsec:supp_hyperparams}

The main paper reports the shared training setup. Here we list only the policy-specific optimization settings needed to reproduce \method; all shared quantities remain exactly as in the main text.

\begin{table}[t]
  \caption[Training recipe summary.]{\textbf{Training recipe summary.}}
  \label{tab:supp_training_recipe}
  \centering
  \footnotesize
  \setlength{\tabcolsep}{3pt}
  \begin{tabular}{@{}>{\raggedright\arraybackslash}m{1.35cm}>{\raggedright\arraybackslash}m{3.5cm}>{\raggedright\arraybackslash}m{3cm}@{}}
    \toprule
    Category & Hyperparameter & Value \\
    \midrule
    \multirow{8}{*}{\begin{tabular}{c}VLM \\ Policy\end{tabular}}
      & Base model & Qwen3-VL-4B-Instruct \\
      & Optimizer & AdamW \\
      & Learning rate & $2 \times 10^{-6}$ \\
      & \makecell[l]{Plans per input ($M$)} & 4 \\
      & \makecell[l]{Total rollouts per input ($M \times K$)} & 16 \\
      & \makecell[l]{Global update batch size} & 256 \\
      & \makecell[l]{KL penalty ($\beta$)} & $1 \times 10^{-3}$ \\
      & \makecell[l]{Max context / response length} & 20,480 tokens / 4,096 tokens \\
    \midrule
    \multirow{13}{*}{\begin{tabular}{c}Diffusion \\ Policy\end{tabular}}
      & Base model & Qwen-Image-Edit-2511 \\
      & Fine-tuning method & LoRA \\
      & \makecell[l]{LoRA rank / scaling factor ($\alpha_{\mathrm{LoRA}}$)} & 64 / 128 \\
      & Optimizer & AdamW \\
      & Learning rate & $2 \times 10^{-6}$ \\
      & Weight decay & $1 \times 10^{-4}$ \\
      & \makecell[l]{Flow-GRPO clip range ($\epsilon_{\mathrm{r}}$)} & $1 \times 10^{-4}$ \\
      & \makecell[l]{Renderings per plan ($K$)} & 4 \\
      & \makecell[l]{SDE training timesteps} & $\{1, 2, 3\}$ \\
      & \makecell[l]{EMA decay / update interval} & 0.9 / 4 steps \\
      & \makecell[l]{Max image resolution} & $1024 \times 1024$ \\
      & \makecell[l]{Inference steps (train / eval)} & 10 / 40 \\
      & Guidance scale & 4.0 \\
    \bottomrule
  \end{tabular}
\end{table}

The planner uses a standard AdamW text-policy update to rewrite the raw instruction online into a four-slot plan, while the renderer uses LoRA-based flow-GRPO with DiT-specific sampling controls. Relative to the controlled Joint~RL baseline, \method adds shared-rollout bookkeeping and slot-wise scoring without additional renderer calls.

\subsection{Controlled Baseline and Evaluation Protocol}
\label{subsec:supp_controlled_eval}

The controlled \emph{Joint RL + Adaptive Curriculum} baseline matches \method in backbones, training data, offline checklist cache, online reward judge, rollout budget, hardware, and adaptive curriculum. It differs only in replacing the structured four-slot planner and dual-level credit assignment with a free-form planner trace.

All benchmark results are obtained from the official evaluation code and are independent of the training-time reward pipeline. Gemini 3 Pro and Qwen3-VL-32B are used only for checklist construction and training-time scoring, not for the final reported benchmark metrics. Across all in-house variants, we report the official overall benchmark scores, with PICA-Bench following its standard \emph{simple} / \emph{detailed} split.

\subsection{KRIS-Bench Breakdown}
\label{subsec:supp_kris_breakdown}

Table~\ref{tab:supp_kris_breakdown} expands the main-paper KRIS-Bench result into its official factual-, conceptual-, and procedural-knowledge groups, so performance differences can be inspected at the subgroup level rather than only through the overall score. We also include the newly obtained fine-grained breakdown for Qwen-Image-Edit-2511 and update the best/second-best markings accordingly.

\begin{table*}[t]
  \caption[Fine-grained KRIS-Bench results.]{\textbf{Fine-grained KRIS-Bench results.} \normalfont We use the official GPT-4.1-based evaluation protocol. KRIS-Bench evaluates edited results with four dimensions: visual consistency, visual quality, instruction following, and knowledge plausibility, where the last dimension is used only for knowledge-intensive subsets such as Social Science, Natural Science, and Logical Reasoning. The raw judge outputs are defined on a 1--5 scale, while the benchmark reports normalized percentage scores with maximum 100. Higher is better for every metric. All values are reported with two decimal places.}
  \label{tab:supp_kris_breakdown}
  \centering
  \resizebox{\textwidth}{!}{%
  \begin{tabular}{lccccccccccc}
    \toprule
    \multirow{2}{*}{Model} & \multicolumn{4}{c}{Factual Knowledge} & \multicolumn{3}{c}{Conceptual Knowledge} & \multicolumn{3}{c}{Procedural Knowledge} & \multirow{2}{*}{Overall $\uparrow$} \\
    \cmidrule(lr){2-5} \cmidrule(lr){6-8} \cmidrule(lr){9-11}
    & \makecell[c]{Attribute\\Perception $\uparrow$} & \makecell[c]{Spatial\\Perception $\uparrow$} & \makecell[c]{Temporal\\Perception $\uparrow$} & \makecell[c]{Average\\Score $\uparrow$} & \makecell[c]{Social\\Science $\uparrow$} & \makecell[c]{Natural\\Science $\uparrow$} & \makecell[c]{Average\\Score $\uparrow$} & \makecell[c]{Logical\\Reasoning $\uparrow$} & \makecell[c]{Instruction\\Decomposition $\uparrow$} & \makecell[c]{Average\\Score $\uparrow$} & \\
    \midrule
    \multicolumn{12}{l}{\textit{Other baselines}} \\
    \midrule
    ThinkGen & 64.03 & 55.17 & 56.31 & 59.57 & 66.00 & 63.19 & 63.86 & 38.04 & 64.83 & 49.52 & 59.57 \\
    UniREdit & 61.10 & 66.75 & 58.45 & 62.42 & 66.20 & 63.04 & 63.80 & 53.08 & 54.61 & 53.74 & 61.02 \\
    PromptRL & 67.85 & 64.83 & 79.17 & 69.04 & 58.65 & 57.39 & 57.69 & 51.38 & 67.29 & 58.24 & 61.24 \\
    Step1X-Edit-v1p2 & 69.67 & 64.33 & 80.18 & 70.21 & 63.50 & 60.62 & 61.32 & 49.75 & 62.67 & 55.29 & 62.58 \\
    UniReason 1.0 & 64.03 & 68.08 & 71.28 & 66.13 & 66.75 & 67.26 & 67.13 & 50.50 & 53.89 & 51.95 & 63.26 \\
    ThinkRL-Edit & 68.18 & 67.75 & \textbf{87.50} & 71.27 & 68.85 & 66.49 & 67.06 & 51.63 & 64.94 & 57.33 & 66.04 \\
    PhysicEdit & 67.52 & 74.42 & 84.91 & 71.92 & 67.25 & 68.10 & 67.89 & 48.00 & 71.28 & 57.98 & 66.78 \\
    EditThinker & \underline{73.67} & 71.75 & 81.53 & \underline{74.54} & \underline{71.85} & 67.95 & 68.89 & 56.04 & 68.22 & 61.26 & 68.80 \\
    \midrule
    \multicolumn{12}{l}{\textit{Qwen-Image-Edit-2511-based methods}} \\
    \midrule
    Qwen-Image-Edit-2511 & 68.82 & 74.75 & 77.70 & 71.60 & 61.75 & 58.48 & 59.27 & 56.21 & 81.83 & 67.19 & 64.42 \\
    RePlan & 56.88 & 50.00 & 74.77 & 58.26 & 53.45 & 56.22 & 55.55 & 49.58 & 57.10 & 52.79 & 55.72 \\
    PromptEnhancerV2 & 71.67 & \underline{75.67} & 82.43 & 74.33 & 71.40 & \underline{70.59} & \underline{70.79} & \underline{56.63} & \underline{86.50} & \underline{69.43} & \underline{71.54} \\
    \textbf{DARS} & \textbf{78.21} & \textbf{84.33} & \underline{85.36} & \textbf{80.75} & \textbf{82.95} & \textbf{81.20} & \textbf{81.62} & \textbf{71.00} & \textbf{89.39} & \textbf{78.89} & \textbf{80.72} \\
    \bottomrule
  \end{tabular}%
  }
\end{table*}

\subsection{RISE-Bench Breakdown}
\label{subsec:supp_rise_breakdown}

Table~\ref{tab:supp_rise_breakdown} expands the main-paper RISE-Bench result into the four official reasoning categories of RISEBench-360: \emph{Temporal}, \emph{Causal}, \emph{Spatial}, and \emph{Logical}. This breakdown shows where gains come from across different reasoning types rather than only in the aggregate.

\begin{table*}[t]
  \caption[Fine-grained RISE-Bench results.]{\textbf{Fine-grained RISE-Bench results.} \normalfont We evaluate on RISEBench-360, the 360-sample version of the benchmark, using the official GPT-4.1-based judging pipeline. The benchmark scores each sample along instruction reasoning, appearance consistency, and visual plausibility, and counts a case as solved only when all applicable dimensions receive full marks. The table therefore reports Accuracy (\%), i.e., the task success rate, for each reasoning category and for the overall set. Higher is better for every metric.}
  \label{tab:supp_rise_breakdown}
  \centering
  \large
  \begin{tabular*}{\textwidth}{@{\extracolsep{\fill}}lccccc@{}}
    \toprule
    Method & Temporal $\uparrow$ & Causal $\uparrow$ & Spatial $\uparrow$ & Logical $\uparrow$ & Overall $\uparrow$ \\
    \midrule
    \multicolumn{6}{l}{\textit{Other baselines}} \\
    \midrule
    ThinkGen & 15.29 & 16.67 & 7.00 & 3.52 & 10.56 \\
    UniREdit & 15.29 & 20.00 & 14.00 & 3.53 & 13.33 \\
    PromptRL & 3.53 & 8.89 & 23.00 & 2.34 & 10.00 \\
    Step1X-Edit-v1p2 & 14.12 & 13.30 & 13.00 & 5.88 & 11.67 \\
    UniReason 1.0 & 16.47 & 27.78 & 13.00 & 2.35 & 15.00 \\
    ThinkRL-Edit & 14.11 & 20.00 & 17.00 & 5.88 & 14.44 \\
    PhysicEdit & 15.29 & 27.78 & 22.00 & \textbf{9.41} & 18.89 \\
    EditThinker & 18.82 & 18.89 & 27.00 & 7.06 & 18.33 \\
    \midrule
    \multicolumn{6}{l}{\textit{Qwen-Image-Edit-2511-based methods}} \\
    \midrule
    Qwen-Image-Edit-2511 & \underline{20.80} & 18.25 & \underline{30.67} & 4.65 & 17.50 \\
    RePlan & 10.59 & 12.22 & 11.00 & 4.70 & 9.72 \\
    PromptEnhancerV2 & 20.00 & \underline{30.00} & 18.00 & 7.06 & \underline{18.90} \\
    \textbf{DARS} & \textbf{30.59} & \textbf{38.89} & \textbf{31.00} & \underline{8.24} & \textbf{27.50} \\
    \bottomrule
  \end{tabular*}
\end{table*}

\subsection{ImgEdit-Bench Breakdown}
\label{subsec:supp_imgedit_breakdown}

Table~\ref{tab:supp_imgedit_breakdown} expands the main-paper ImgEdit-Bench result into its nine official edit-operation groups, making it easier to see where different methods are strong or weak across common editing regimes rather than only in the aggregate. All entries follow the benchmark's official GPT-4.1-based evaluation protocol.

\begin{table*}[t]
  \caption[Fine-grained ImgEdit-Bench results.]{\textbf{Fine-grained ImgEdit-Bench results.} \normalfont The header enumerates the benchmark's nine common editing tasks: \emph{Add}, \emph{Remove}, \emph{Adjust}, \emph{Replace}, \emph{Style}, \emph{Background}, \emph{Action}, \emph{Hybrid}, and \emph{Extract}. ImgEdit-Bench evaluates each result with GPT-4.1 on a 1--5 scale from three perspectives: instruction following, editing quality, and detail preservation. Instruction following measures understanding of the prompt and target concept; editing quality measures how accurately the target region is manipulated; and detail preservation measures fidelity on content that should remain unchanged. Because instruction following is foundational and cannot be fully separated from the other two aspects, the editing-quality and detail-preservation scores are upper-bounded by the instruction-following score. The table reports the average score for each edit category, and higher is better for every metric.}
  \label{tab:supp_imgedit_breakdown}
  \centering
  \begin{tabular*}{\textwidth}{@{\extracolsep{\fill}}lcccccccccc@{}}
    \toprule
    Method & Add $\uparrow$ & Adjust $\uparrow$ & Extract $\uparrow$ & Replace $\uparrow$ & Remove $\uparrow$ & Background $\uparrow$ & Style $\uparrow$ & Hybrid $\uparrow$ & Action $\uparrow$ & Overall $\uparrow$ \\
    \midrule
    \multicolumn{11}{l}{\textit{Other baselines}} \\
    \midrule
    ThinkGen & 4.20 & 3.96 & 3.45 & 4.35 & 3.35 & 4.22 & \textbf{4.91} & 3.45 & 4.16 & 3.97 \\
    UniREdit & 4.09 & 3.53 & 2.27 & 4.27 & 3.81 & 3.83 & 4.65 & 2.46 & 4.37 & 3.70 \\
    PromptRL & 4.07 & 4.19 & 2.95 & 4.23 & 3.67 & 4.09 & 4.46 & 3.04 & 4.13 & 3.87 \\
    Step1X-Edit-v1p2 & 4.28 & 4.40 & 2.29 & 4.36 & 4.17 & 3.94 & 4.74 & \underline{3.61} & 4.23 & 4.00 \\
    UniReason 1.0 & 4.18 & 3.78 & 2.65 & 4.50 & \textbf{4.39} & 4.01 & 4.73 & 2.78 & 4.36 & 4.06 \\
    ThinkRL-Edit & 4.23 & 4.18 & 3.51 & \textbf{4.71} & 4.08 & 4.09 & 4.89 & 3.59 & 4.42 & 4.19 \\
    PhysicEdit & 2.67 & 3.92 & 2.67 & 4.57 & \underline{4.35} & 3.97 & 4.79 & 3.33 & 4.56 & 4.05 \\
    EditThinker & 4.33 & 4.17 & 3.91 & \underline{4.66} & 4.18 & \textbf{4.29} & \underline{4.90} & \textbf{3.65} & \textbf{4.68} & 4.31 \\
    \midrule
    \multicolumn{11}{l}{\textit{Qwen-Image-Edit-2511-based methods}} \\
    \midrule
    Qwen-Image-Edit-2511 & \textbf{4.55} & \textbf{4.48} & \textbf{4.21} & 4.62 & 4.29 & 4.23 & \underline{4.90} & 3.37 & \underline{4.63} & \underline{4.36} \\
    RePlan & 4.11 & 3.78 & 2.07 & 3.41 & 3.02 & 3.85 & 4.48 & 3.07 & 3.16 & 3.44 \\
    PromptEnhancerV2 & 4.28 & 4.40 & 3.43 & 4.43 & 4.18 & \underline{4.27} & 4.77 & \underline{3.61} & 4.31 & 4.18 \\
    \textbf{DARS} & \underline{4.54} & \underline{4.47} & \underline{4.18} & 4.63 & 4.32 & 4.24 & \underline{4.90} & \underline{3.61} & 4.61 & \textbf{4.39} \\
    \bottomrule
  \end{tabular*}
\end{table*}

\subsection{GEdit-Bench-EN Breakdown}
\label{subsec:supp_gedit_breakdown}

Table~\ref{tab:supp_gedit_breakdown} expands the main-paper GEdit-Bench-EN result into its official split-wise metrics, so the relative behavior of different methods can be inspected separately on the \emph{Intersection subset} and the \emph{Full set}. All numbers come from the official evaluation code.

\begin{table*}[t]
  \caption[Fine-grained GEdit-Bench-EN results.]{\textbf{Fine-grained GEdit-Bench-EN results.} \normalfont The header uses the official benchmark notation: \emph{G\_SC} denotes GPT-4.1-based semantic consistency, \emph{G\_PQ} denotes GPT-4.1-based perceptual quality, and \emph{G\_O} denotes the official overall score. Semantic consistency evaluates how well the edited result follows the instruction, while perceptual quality evaluates image naturalness and artifact level; both use a 0--10 scale. ``Intersection subset'' refers to the subset of test images for which all compared models return valid outputs, whereas ``Full set'' refers to the complete GEdit-Bench-EN test set. All values are reported with two decimal places.}
  \label{tab:supp_gedit_breakdown}
  \centering
  \large
  \begin{tabular*}{\textwidth}{@{\extracolsep{\fill}}lcccccc@{}}
    \toprule
    \multirow{2}{*}{Method} & \multicolumn{3}{c}{Intersection subset} & \multicolumn{3}{c}{Full set} \\
    \cmidrule(lr){2-4} \cmidrule(lr){5-7}
    & G\_SC $\uparrow$ & G\_PQ $\uparrow$ & G\_O $\uparrow$ & G\_SC $\uparrow$ & G\_PQ $\uparrow$ & G\_O $\uparrow$ \\
    \midrule
    \multicolumn{7}{l}{\textit{Other baselines}} \\
    \midrule
    ThinkGen & 7.49 & 7.80 & 7.22 & 7.43 & 7.76 & 7.16 \\
    UniREdit & 7.31 & 6.96 & 6.67 & 7.10 & 6.89 & 6.47 \\
    PromptRL & 7.32 & 7.60 & 6.90 & 7.15 & 7.58 & 6.74 \\
    Step1X-Edit-v1p2 & 8.32 & 7.89 & 7.73 & 8.05 & 7.85 & 7.47 \\
    UniReason 1.0 & 7.18 & 7.21 & 6.62 & 7.06 & 7.20 & 6.52 \\
    ThinkRL-Edit & 6.84 & \textbf{8.21} & 6.63 & 6.70 & \textbf{8.16} & 6.48 \\
    PhysicEdit & 8.08 & 7.76 & 7.54 & 7.93 & 7.66 & 7.37 \\
    EditThinker & 8.20 & 7.84 & 7.64 & \underline{8.12} & 7.88 & \underline{7.59} \\
    \midrule
    \multicolumn{7}{l}{\textit{Qwen-Image-Edit-2511-based methods}} \\
    \midrule
    Qwen-Image-Edit-2511 & 7.67 & 7.68 & 7.24 & 7.65 & 7.38 & 6.97 \\
    RePlan & 7.03 & 7.66 & 6.55 & 6.87 & 7.62 & 6.43 \\
    PromptEnhancerV2 & \underline{8.33} & 7.97 & \underline{7.79} & 8.08 & 7.96 & \underline{7.59} \\
    \textbf{DARS} & \textbf{8.57} & \underline{8.01} & \textbf{7.98} & \textbf{8.49} & \underline{8.00} & \textbf{7.86} \\
    \bottomrule
  \end{tabular*}
\end{table*}

\subsection{PICA-Bench Simple-Prompt Breakdown}
\label{subsec:supp_pica_simple_breakdown}

Table~\ref{tab:supp_pica_simple_breakdown} reports the fine-grained PICA-Bench results under the benchmark's \emph{simple prompt} setting, which we also refer to as the \emph{superficial prompt} setting for clarity. We list all official physical-consistency categories: \emph{LP}, \emph{LSE}, \emph{Reflection}, \emph{Refraction}, \emph{Deformation}, \emph{Causality}, \emph{GST}, and \emph{LST}, together with the overall score. As with the other supplementary benchmark tables, all values come from the official evaluation pipeline and are reported with two decimal places.

\begin{table*}[t]
  \caption[Fine-grained PICA-Bench results under the simple-prompt setting.]{\textbf{Fine-grained PICA-Bench results under the simple-prompt setting.} \normalfont This setting corresponds to the superficial-prompt protocol in PICA-Bench and is evaluated by GPT-4.1 for instruction-based image editing models. All reported numbers are Accuracy (\%). In the header, LP, LSE, GST, and LST denote Light Propagation, Light Source Effects, Global State Transition, and Local State Transition, respectively.}
  \label{tab:supp_pica_simple_breakdown}
  \centering
  \begin{tabular*}{\textwidth}{@{\extracolsep{\fill}}lccccccccc@{}}
    \toprule
    Method & LP $\uparrow$ & LSE $\uparrow$ & Reflection $\uparrow$ & Refraction $\uparrow$ & Deformation $\uparrow$ & Causality $\uparrow$ & GST $\uparrow$ & LST $\uparrow$ & Overall $\uparrow$ \\
    \midrule
    \multicolumn{10}{l}{\textit{Other baselines}} \\
    \midrule
    ThinkGen & \underline{64.39} & 66.34 & 64.51 & \textbf{65.46} & 50.06 & 51.26 & \textbf{68.38} & \textbf{66.34} & 60.82 \\
    UniREdit & 61.99 & 66.17 & 67.50 & 54.38 & 53.11 & 52.90 & 65.68 & 56.83 & 60.40 \\
    PromptRL & 56.04 & 52.24 & 59.27 & 41.35 & 50.42 & 45.60 & 52.91 & 45.67 & 51.02 \\
    Step1X-Edit-v1p2 & 63.52 & 58.12 & 66.61 & 62.44 & 53.41 & 48.76 & 52.60 & 51.89 & 56.15 \\
    UniReason 1.0 & 62.15 & 63.20 & 62.73 & 51.35 & 52.55 & 48.84 & 60.21 & 54.76 & 57.30 \\
    ThinkRL-Edit & 60.96 & 56.68 & 61.18 & 41.16 & 56.92 & 46.07 & 59.53 & 52.66 & 55.13 \\
    PhysicEdit & 63.10 & \underline{69.26} & 67.34 & 51.96 & \underline{58.31} & 51.47 & 65.79 & 61.51 & 61.48 \\
    EditThinker & 63.05 & 59.98 & 65.75 & 53.35 & 55.34 & 50.76 & 64.50 & 60.98 & 59.64 \\
    \midrule
    \multicolumn{10}{l}{\textit{Qwen-Image-Edit-2511-based methods}} \\
    \midrule
    Qwen-Image-Edit-2511 & 63.89 & \textbf{69.91} & \underline{71.86} & 62.33 & 57.93 & 50.70 & \underline{67.89} & 61.34 & \underline{63.20} \\
    RePlan & 52.99 & 51.29 & 49.29 & 48.88 & 52.33 & 43.82 & 61.26 & 52.15 & 52.00 \\
    PromptEnhancerV2 & \textbf{66.86} & 66.19 & 70.86 & 59.55 & 53.71 & \underline{53.06} & 62.90 & 63.38 & 61.90 \\
    \textbf{DARS} & 62.52 & 67.85 & \textbf{72.04} & \underline{64.62} & \textbf{64.37} & \textbf{53.59} & 65.50 & \underline{66.16} & \textbf{64.19} \\
    \bottomrule
  \end{tabular*}
\end{table*}

\subsection{PICA-Bench Explicit-Prompt Breakdown}
\label{subsec:supp_pica_explicit_breakdown}

Table~\ref{tab:supp_pica_explicit_breakdown} reports the fine-grained PICA-Bench results under the benchmark's \emph{detailed prompt} setting, which corresponds to the \emph{explicit prompt} protocol used in our evaluation pipeline. We keep the same category layout as in Table~\ref{tab:supp_pica_simple_breakdown}: \emph{LP}, \emph{LSE}, \emph{Reflection}, \emph{Refraction}, \emph{Deformation}, \emph{Causality}, \emph{GST}, \emph{LST}, and the overall score. All values come from the official evaluation pipeline and are reported with two decimal places.

\begin{table*}[t]
  \caption[Fine-grained PICA-Bench results under the explicit-prompt setting.]{\textbf{Fine-grained PICA-Bench results under the explicit-prompt setting.} \normalfont All reported numbers are Accuracy (\%). All values are reported with two decimal places.}
  \label{tab:supp_pica_explicit_breakdown}
  \centering
  \begin{tabular*}{\textwidth}{@{\extracolsep{\fill}}lccccccccc@{}}
    \toprule
    Method & LP $\uparrow$ & LSE $\uparrow$ & Reflection $\uparrow$ & Refraction $\uparrow$ & Deformation $\uparrow$ & Causality $\uparrow$ & GST $\uparrow$ & LST $\uparrow$ & \makecell[c]{Overall\\Score $\uparrow$} \\
    \midrule
    \multicolumn{10}{l}{\textit{Other baselines}} \\
    \midrule
    ThinkGen & 66.17 & 68.29 & 64.82 & \textbf{62.63} & 59.43 & 58.57 & 74.48 & 61.12 & 65.09 \\
    UniREdit & 64.34 & 66.80 & 67.50 & 59.77 & 63.72 & 62.17 & 72.81 & 68.41 & 66.55 \\
    PromptRL & 61.20 & 59.68 & 63.66 & 43.02 & 55.11 & 62.88 & 66.96 & 55.50 & 60.38 \\
    Step1X-Edit-v1p2 & 58.92 & 61.60 & 66.02 & 49.12 & 56.68 & 59.30 & 61.36 & 63.91 & 60.55 \\
    UniReason 1.0 & 59.32 & 64.85 & 60.03 & 49.77 & 56.47 & 60.83 & 69.23 & 61.09 & 61.62 \\
    ThinkRL-Edit & 57.20 & 55.39 & 57.77 & 37.63 & 48.89 & 54.39 & 62.06 & 58.94 & 55.84 \\
    PhysicEdit & 64.39 & 70.14 & 68.36 & 54.10 & \underline{65.90} & 70.29 & 74.61 & 70.79 & 68.96 \\
    EditThinker & 66.79 & 64.20 & 68.21 & 48.67 & 60.49 & 67.03 & 74.42 & 71.19 & 67.19 \\
    \midrule
    \multicolumn{10}{l}{\textit{Qwen-Image-Edit-2511-based methods}} \\
    \midrule
    Qwen-Image-Edit-2511 & 67.36 & \textbf{75.82} & 73.87 & 59.67 & \textbf{66.02} & \textbf{71.07} & \textbf{78.58} & \textbf{73.99} & \underline{72.27} \\
    RePlan & 46.84 & 56.27 & 55.25 & 51.57 & 49.35 & 49.87 & 62.66 & 60.77 & 54.92 \\
    PromptEnhancerV2 & \underline{67.46} & 74.01 & \underline{74.00} & 54.25 & 63.32 & 70.37 & 76.29 & 72.62 & 70.77 \\
    \textbf{DARS} & \textbf{67.79} & \underline{74.97} & \textbf{75.79} & \underline{60.46} & 64.92 & \underline{70.79} & \underline{78.21} & \underline{73.32} & \textbf{72.75} \\
    \bottomrule
  \end{tabular*}
\end{table*}

\subsection{Baseline Architecture Overview}
\label{subsec:supp_baselines}

Table~\ref{tab:supp_baselines} summarizes the main backbone composition and dominant control mechanism of each baseline based on the released model descriptions and our implementation notes.

\begin{table*}[t]
  \caption[Detailed baseline architecture summary.]{\textbf{Detailed baseline architecture summary.}}
  \label{tab:supp_baselines}
  \centering
  \small
  \begin{tabular}{@{}>{\raggedright\arraybackslash}m{2.7cm}>{\raggedright\arraybackslash}m{3.5cm}>{\raggedright\arraybackslash}m{9.8cm}@{}}
    \toprule
    Method & Main Backbone & Key Structural Characteristics \\
    \midrule
    ThinkRL-Edit & Qwen3-VL-30B-A3B + Qwen-Image-Edit & Checklist-style reward-driven RL with explicit CoT planning and reflection before image-to-image diffusion editing. \\
    ThinkGen & Qwen3-VL-8B-Think + OmniGen2-4B DiT & Alternating GRPO between a reasoning VLM and a DiT generator. \\
    Step1X-Edit-v1p2 & Qwen2.5-VL-7B + Step1X-Edit & Multi-round reflection pipeline (up to five rounds) with coupled VLM and editor tuning. \\
    RePlan & Qwen2.5-VL-7B + Qwen-Image-Edit-2511 & Region-aware decomposition: an RL-trained VLM predicts boxes and localized prompts for editing. \\
    EditThinker & Qwen3-VL-8B + Qwen-Image-Edit & Up to five rounds of reflection, with reasoning refinement concentrated on the VLM side. \\
    UniReason 1.0 & Bagel-14B & World-knowledge-aware prompt enhancement with reflection and consistency control. \\
    PromptRL & Qwen2.5-VL-3B + FLUX .1-Kontext-12B & Joint RL over the prompt-side VLM and the image editor. \\
    UniREdit & Bagel-14B & Uses large-scale text reasoning data together with mixed real-world and game-world supervision for reasoning-oriented editing. \\
    PhysicEdit & PhysicEdit data + Qwen-Image-Edit-2509 & Physically grounded reasoning with meta-query-based control for physics-aware image editing. \\
    PromptEnhancerV2 - Img2Img Edit & Qwen2.5-VL-32B + Qwen-Image-Edit-2511 & Prompt-enhancement front-end paired with a strong external image editor; the public description mainly exposes the front-end/control structure rather than the full joint training recipe. \\
    Qwen-Image-Edit-2511 & Qwen-Image-Edit-2511 & Strong native end-to-end image editing model used as a general-purpose reference baseline. \\
    \bottomrule
  \end{tabular}
\end{table*}

% \FloatBarrier
\section{Failure Attribution Validation Details}
\label{subsec:supp_attribution}

We sample 500 evaluation cases for the attribution study, with an even split between relatively simple and difficult edits (250 each), so the validation set covers both low-difficulty and high-difficulty failure modes. For each case, GPT-5 produces five independent failure-attribution labels, and the final case-level pseudo-label is the majority vote. The system prompt used for this annotation step is shown in Fig.~\ref{fig:supp_attribution_prompt}.

The GPT-5 annotator returns one of three raw labels: \texttt{planner\_side}, \texttt{renderer\_side}, or \texttt{both}. In the paper, these are reported respectively as \emph{planner-dominant}, \emph{renderer-dominant}, and \emph{mixed}. After obtaining these case-level three-way pseudo-labels, we form three pairwise binary tasks by filtering the same 500-case pool into \emph{Planner vs.\ Renderer}, \emph{Planner vs.\ Mixed}, and \emph{Renderer vs.\ Mixed}; we do not resample separately for the binary tasks, so the retained sample count in each setting is determined by the realized three-way label frequencies in this shared evaluation pool.

Accuracy and Macro-F1 are computed within each resulting binary task, and AUROC is reported in the same one-vs.-one setting rather than under a three-way formulation. GPT-5 sees only the source image, the structured plan, and the edited image for each sampled triplet; it does not see rollout statistics, checklist-based rewards, or slot scores.

\section{Additional Qualitative Results}

This section adds three complementary qualitative views: additional baseline comparisons, more reasoning-intensive edits, and more general editing cases.

\begin{figure*}[p]
  \centering
  \includegraphics[width=0.916\textwidth,height=0.72\textheight,keepaspectratio]{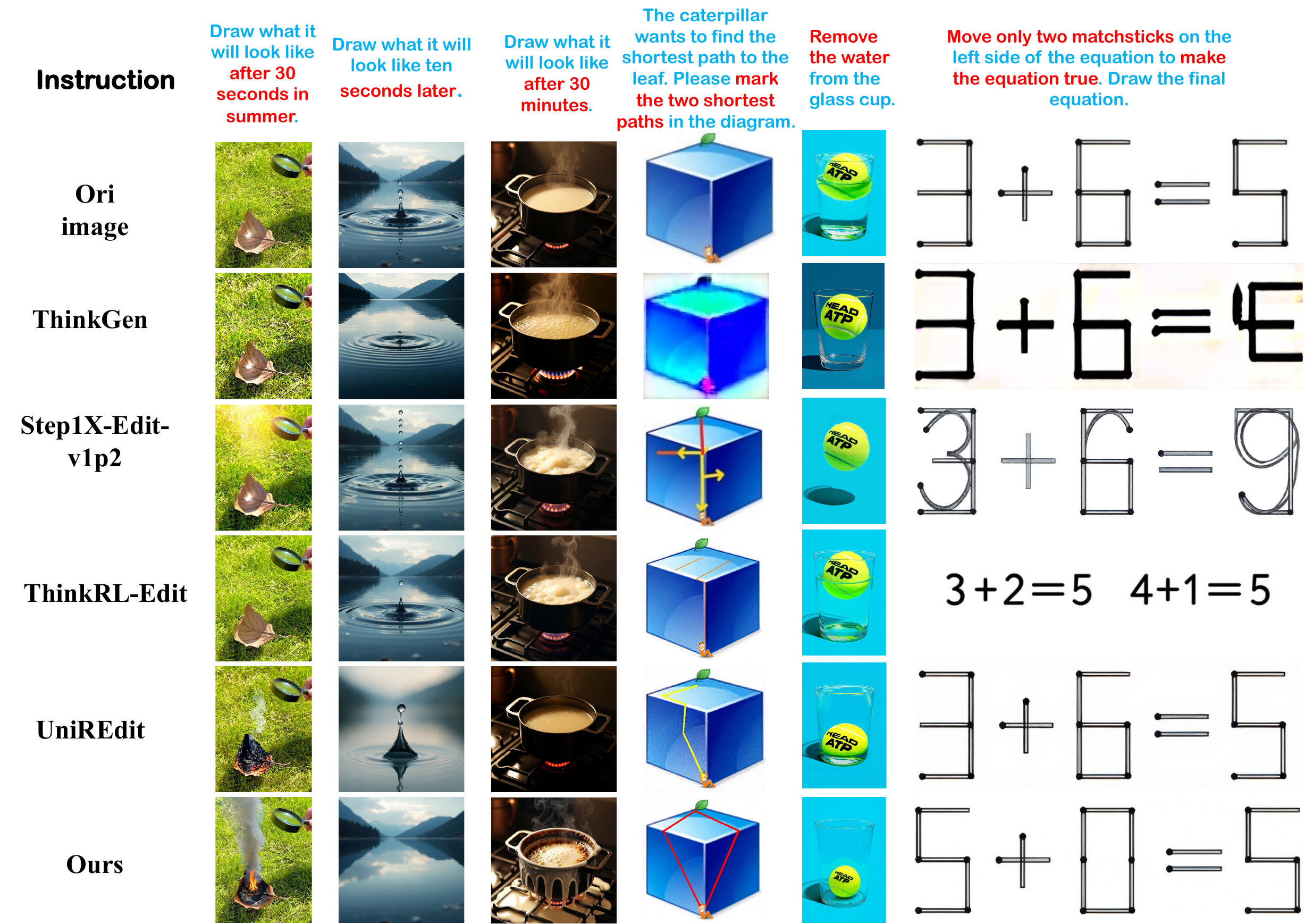}
  \caption[Additional baseline comparisons on challenging edits.]{\textbf{Additional baseline comparisons on challenging edits.} \normalfont The figure adds side-by-side results for several baselines under the same evaluation setting.}
  \Description{Supplementary qualitative comparison figure showing additional baseline outputs on challenging image editing tasks, alongside DARS results for the same cases.}
  \label{fig:supp_extra_baselines}
\end{figure*}

\paragraph{Additional baseline comparisons.}
Figure~\ref{fig:supp_extra_baselines} adds side-by-side comparisons on challenging edits involving temporal prediction, logical diagram reasoning, path planning, object removal, and symbolic transformation. These cases make it easier to inspect where \method preserves the target semantics more reliably than the omitted baselines.

\begin{figure*}[p]
  \centering
  \includegraphics[width=0.8\textwidth,height=0.82\textheight,keepaspectratio]{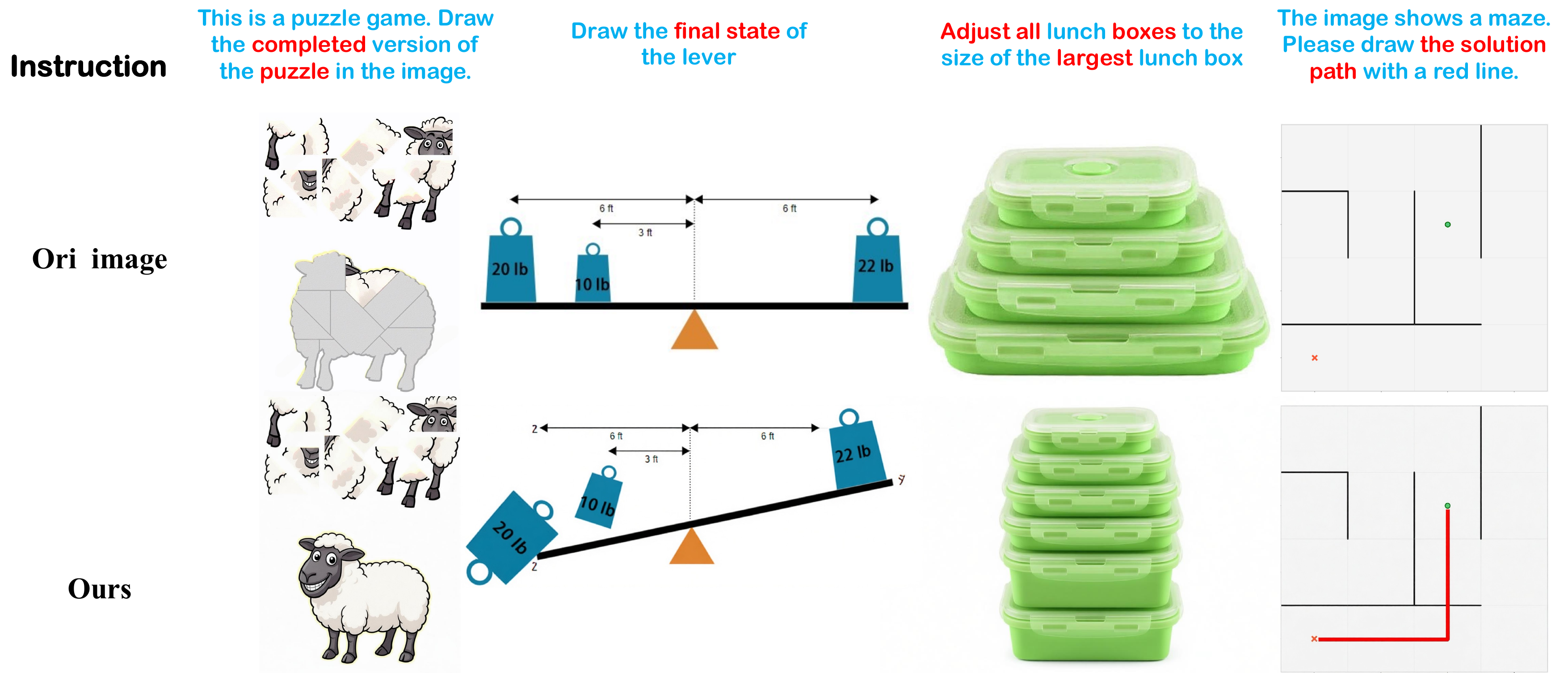}
  \caption[Representative failure cases illustrating current limitations.]{\textbf{Representative failure cases illustrating current limitations.} \normalfont From left to right, the examples highlight failure modes on puzzle completion, dense final-state reasoning in a mechanics-style diagram, precise relative-size control across multiple objects, and maze-path drawing. Together they show that the method remains weaker when the desired target is difficult to specify textually, when the edit depends on tightly coupled multi-step reasoning, or when exact geometric control is required.}
  \Description{Failure-case figure showing four representative limitations: incomplete puzzle completion, an incorrect lever final state, inaccurate size normalization across stacked lunch boxes, and an unstable maze solution path.}
  \label{fig:supp_limitations}
\end{figure*}

\begin{figure*}[t]
  \centering
  \includegraphics[width=1\textwidth,height=0.945\textheight,keepaspectratio]{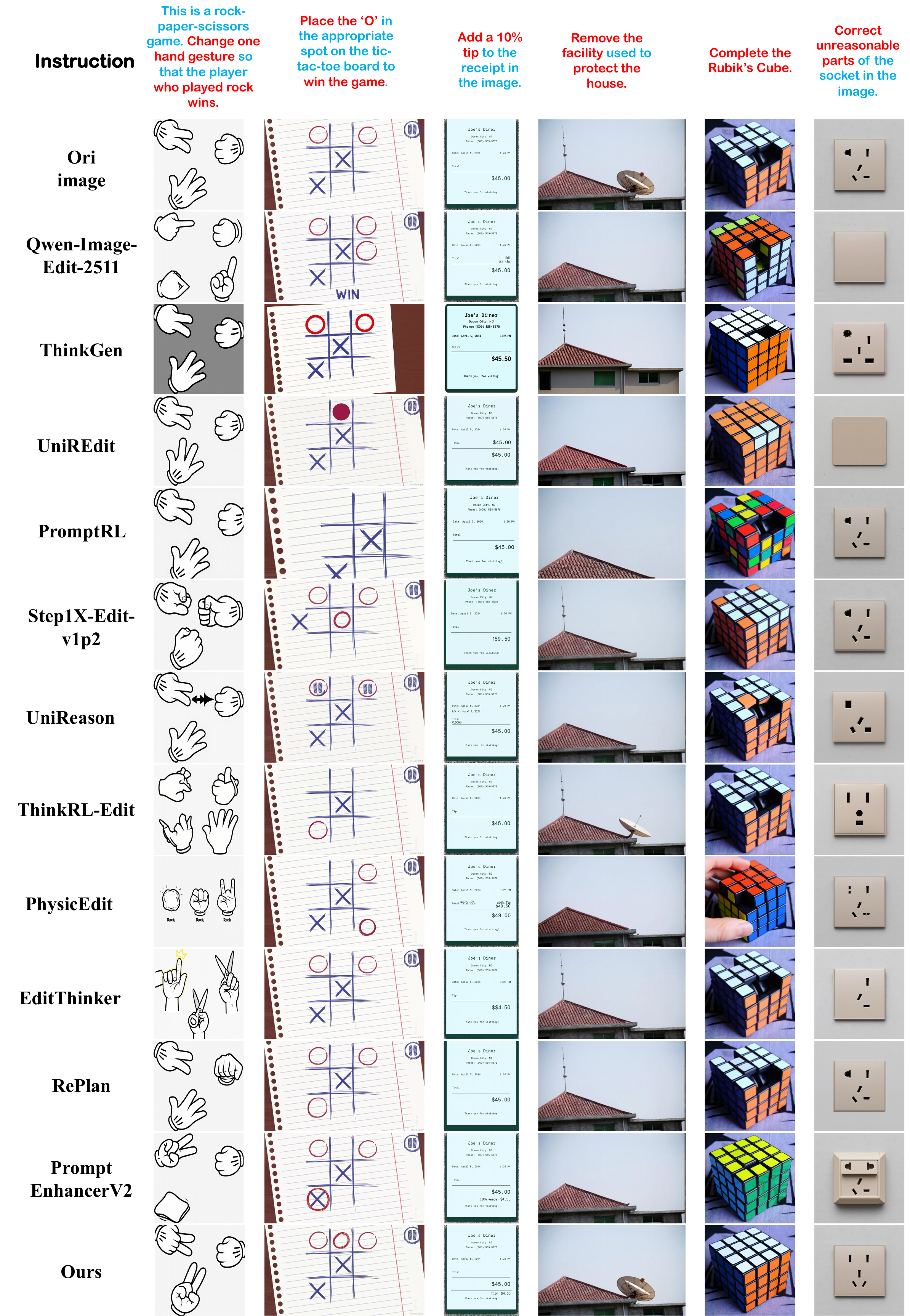}
  \caption[Additional reasoning-intensive editing examples.]{\textbf{Additional reasoning-intensive editing examples.} \normalfont The figure covers temporal, spatial, logical, physical, and knowledge-intensive editing cases.}
  \Description{Supplementary qualitative figure with additional reasoning-intensive image editing examples, showing source images, instructions, model outputs, and DARS results on more complex reasoning-heavy tasks.}
  \label{fig:supp_more_reasoning}
\end{figure*}

\begin{table*}[p]
\caption[Actual four-slot plan decompositions for representative qualitative cases.]{\textbf{Actual four-slot plan decompositions for representative qualitative cases.} \normalfont Each row lists the raw user instruction together with its \emph{Modify}, \emph{Preserve}, \emph{Overall}, and \emph{Tips} fields in the same slot schema used throughout the paper.}
\label{tab:supp_qual_plans}
\centering
\definecolor{planmodify}{RGB}{170,48,48}
\definecolor{planpreserve}{RGB}{32,118,72}
\definecolor{planoverall}{RGB}{41,88,167}
\definecolor{plantips}{RGB}{148,96,28}
\footnotesize
\setlength{\tabcolsep}{1pt}
\renewcommand{\arraystretch}{0.82}
\begin{tabular}{@{}>{\raggedright\arraybackslash}m{0.025\textwidth}>{\raggedright\arraybackslash}m{0.105\textwidth}>{\color{planmodify}\raggedright\arraybackslash}m{0.22\textwidth}>{\color{planpreserve}\raggedright\arraybackslash}m{0.20\textwidth}>{\color{planoverall}\raggedright\arraybackslash}m{0.20\textwidth}>{\color{plantips}\raggedright\arraybackslash}m{0.215\textwidth}@{}}
\toprule
\textbf{ID} & \textbf{Raw} & \textcolor{planmodify}{\textbf{Modify}} & \textcolor{planpreserve}{\textbf{Preserve}} & \textcolor{planoverall}{\textbf{Overall}} & \textcolor{plantips}{\textbf{Tips}} \\
\midrule
1 & Draw what it will look like after 30 seconds in summer. & Show the dry leaf catching fire under the concentrated sunlight from the magnifying glass, with a small flame and a plume of grey smoke rising from the bright focal point on the leaf. & Preserve the surrounding green grass, the hand holding the magnifying glass, the position of the lens, and the overall sunny daylight illumination. & The scene should depict the early stage of combustion in a visually coherent and physically plausible way under strong summer sunlight. & Add a charred, blackened texture around the focal point and ensure the smoke and flame originate naturally from the concentrated light spot on the brown leaf. \\
\addlinespace[0.48em]
2 & Draw what it will look like ten seconds later. & Edit the image to depict the scene ten seconds later by removing the suspended water droplets, the central splash pillar, and the concentric ripple rings. & Preserve the background mountains, the tree-covered hills, the cloudy sky, the soft natural lighting, and the original lake setting. & The final image should look calm, smooth, and undisturbed, as if the splash event has fully settled into a serene static scene. & Maintain the original low camera angle, depth of field, and color grading, and ensure the reflections on the still water align naturally with the surrounding landscape. \\
\addlinespace[0.48em]
3 & Draw what it will look like after 30 minutes. & Change the scene to show the pot after the milk has been boiling for 30 minutes, with messy frothy milk spills down the black pot and onto the gas stove grate, along with a reduced, thickened, overheated surface inside the pot. & Preserve the shape and material of the pot, the visible gas flame, the stove grate structure, the blurred background, and the original warm lighting and perspective. & The result should convey a realistic long-boiling, overheating situation with coherent interactions between milk, pot, stove, steam, and smoke. & Add bubbly and browned residue along the inner rim, mix darker smoke with the steam, and make the spilled milk follow gravity and pool plausibly on the stove surfaces. \\
\addlinespace[0.48em]
4 & The caterpillar wants to find the shortest path to the leaf. Please mark the two shortest paths in the diagram. & Mark the two shortest surface paths from the orange caterpillar at the bottom front corner of the cube to the green leaf on the top face using two distinct overlaid lines. & Preserve the original blue cube, the green leaf, the orange caterpillar, and the white background exactly as they are. & The final image should look like a clean diagrammatic solution that remains geometrically consistent with the cube's perspective. & Draw one path across the front-left face and top face, and the other across the front-right face and top face, rendering both lines clearly and aligned to the cube surface geometry. \\
\addlinespace[0.48em]
5 & Move only two matchsticks on the left side of the equation to make the equation true. Draw the final equation. & Transform the left side of the matchstick equation so the final result reads 3 - 3 = 0 by changing the two left digits accordingly. & Preserve the minus sign, the equals sign, the digit 0 on the right, the white background, and the overall matchstick style. & The final image must show a coherent, mathematically correct, and well-aligned matchstick equation. & Turn the 8 into a 3 by removing the two left vertical matches, turn the 7 into a 3 by adding a middle and bottom horizontal match, and match the existing dark match heads and rectangular shafts. \\
\addlinespace[0.48em]
6 & Remove the water from the glass cup. & Remove the water from the glass cup so the glass interior appears empty and dry, and the tennis ball appears without liquid-induced distortion. & Preserve the ``HEAD ATP'' text, the fuzzy tennis-ball texture, the shape and material of the clear glass cup, and the bright cyan background. & The final result should look like a realistic dry glass container holding the ball naturally at the bottom under the same viewing conditions. & Eliminate optical refraction, magnification, and distortion from the ball, while maintaining the original lighting, shadows, and camera viewpoint. \\
\addlinespace[0.48em]
7 & This is a rock-paper-scissors game. Change one hand gesture so that the player who played rock wins. & Change the bottom hand gesture from paper to scissors so that the rock player wins the game. & Preserve the top-left hand showing scissors and the right hand showing rock exactly as they are, along with the black-and-white cartoon drawing style. & The final image should remain a consistent rock-paper-scissors illustration with correct game logic and unchanged visual composition. & Redraw the bottom hand as a fist with the index and middle fingers extended in a V-shape, matching the original line thickness and glove-like cartoon design. \\
\addlinespace[0.48em]
8 & Place the ``O'' in the appropriate spot on the tic-tac-toe board to win the game. & Add a red O mark to the top center square of the tic-tac-toe grid to complete the winning row. & Preserve the blue grid lines, the blue X marks, the existing red O marks, the lined notebook paper background, and the pause icon in the top-right corner. & The final image should look like a consistent hand-drawn tic-tac-toe board where the O player wins naturally. & Position the new red O in the top-center cell, match the existing red ink color and rough sketch style, and keep the stroke thickness consistent with the other O marks. \\
\addlinespace[0.48em]
9 & Add a 10\% tip to the receipt in the image. & Add a new line below the total amount of \$45.00 labeled ``Tip:'' followed by ``\$4.50''. & Preserve all existing receipt text, layout, spacing, and visual style. & The receipt should still look like a coherent and authentic printed receipt with the new tip line integrated naturally. & Format the new line using the same font, alignment, spacing, and typographic style as the rest of the receipt. \\
\addlinespace[0.48em]
10 & Remove the facility used to protect the house. & Remove the lightning rod structure from the peak of the red tiled roof, including the vertical pole and its spherical components. & Preserve the red roof tiles, the concrete building structure, the windows, the large satellite dish on the right, and the original daylighting and perspective. & The final image should look like a realistic roof scene with no visible trace of the removed protection facility. & Seamlessly reconstruct the roof ridge and fill the removed area with surrounding sky so the roofline and sky gradient remain continuous and undisturbed. \\
\addlinespace[0.48em]
11 & Complete the Rubik's Cube. & Fill in the missing cubies to restore the puzzle into a complete 4x4x4 cube with solid cubic geometry. & Preserve all existing blue, orange, and white stickers, the black plastic body, and the purple patterned background. & The completed puzzle should look like a coherent, fully assembled cube with consistent structure, perspective, and material appearance. & Add the missing top- and right-face blocks with white stickers on top-facing surfaces and orange stickers on right-facing surfaces, aligning them perfectly with the existing grid lines and shading. \\
\addlinespace[0.48em]
12 & Correct unreasonable parts of the socket in the image. & Correct the unrealistic and scattered arrangement of holes on the wall socket by replacing them with a standard, symmetrical, and functional outlet pattern centered on the plate. & Preserve the beige square faceplate, the outer frame, the light gray speckled wall background, and the overall lighting and texture. & The final socket should look like a realistic manufactured electrical outlet with a clean, balanced, and logically organized layout. & Use a recognized outlet configuration such as a standard 3-pin or universal socket design, organize the holes into logical pairs or groups, and render their depth, shading, and geometry with precise plastic-like realism. \\
\bottomrule
\end{tabular}
\end{table*}

\begin{figure*}[p]
  \centering
  \includegraphics[width=1\textwidth,height=0.94\textheight,keepaspectratio]{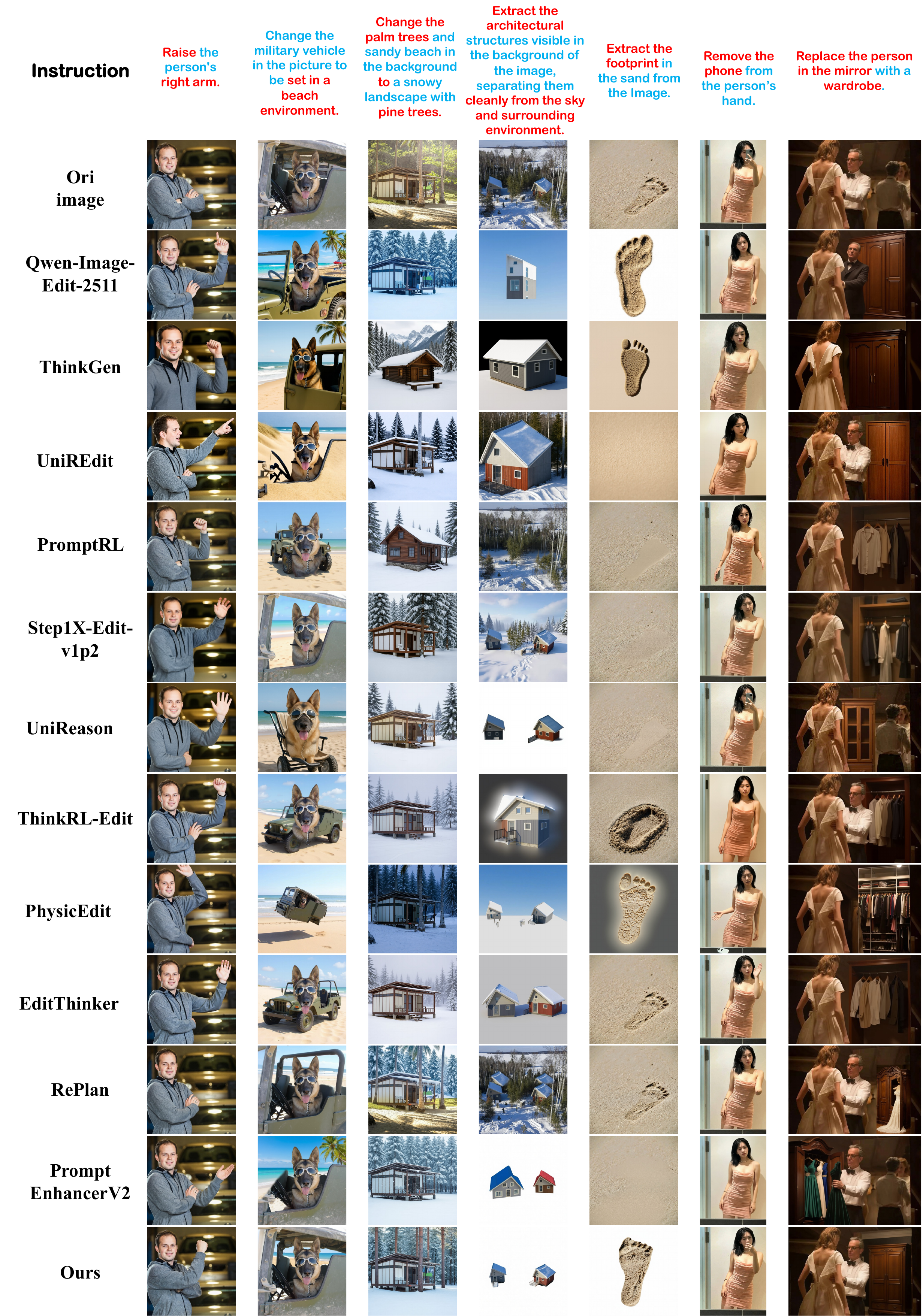}
  \caption[Additional general editing examples.]{\textbf{Additional general editing examples.} \normalfont The figure covers localized modifications, preservation-sensitive edits, semantic scene changes, extraction, and removal cases.}
  \Description{Supplementary qualitative figure with additional general image editing examples, complementing the reasoning-intensive cases with more standard editing scenarios and DARS outputs.}
  \label{fig:supp_more_general}
\end{figure*}

\paragraph{Additional reasoning-intensive edits.}
Figure~\ref{fig:supp_more_reasoning} shows that the advantage of \method remains visible across several reasoning-heavy regimes, including rule-based game solving (e.g., rock-paper-scissors and tic-tac-toe), commonsense inference edits such as tip calculation and identifying the lightning rod as the protective facility to remove, and object completion or correction tasks involving missing or unreasonable structures.

\paragraph{Additional general editing cases.}
Figure~\ref{fig:supp_more_general} shows that the benefit of \method is not limited to explicitly puzzle-like instructions. Across more diverse general editing cases with richer object composition and more complex inter-object relationships, \method consistently demonstrates strong fine-grained control over local edits while preserving non-target content. These examples further suggest that the method maintains robust scene understanding and object-level recognition even in visually complex images, enabling precise edits without sacrificing overall consistency.

\subsection{Structured Plans for Qualitative Cases}

Table~\ref{tab:supp_qual_plans} lists the four-slot decompositions for representative cases in Figs.~\ref{fig:supp_extra_baselines}--\ref{fig:supp_more_general}, so readers can directly inspect the semantic allocation across \emph{Modify}, \emph{Preserve}, \emph{Overall}, and \emph{Tips}.

\section{Prompt Templates}
\label{subsec:supp_prompt_bank}

We collect here the prompt templates used in three places in the pipeline. Figure~\ref{fig:supp_attribution_prompt} is used only in the GPT-5 failure-attribution validation study. Figure~\ref{fig:supp_rewrite_prompt} is the planner-side four-slot rewriting prompt used during rollout to produce $\mathbf{e}$. Figure~\ref{fig:supp_checklist_prompt} is the offline Gemini 3 Pro prompt used to generate the cached shared checklist $\mathbf{C}(\mathbf{x})$. During RL itself, Qwen3-VL-32B is used only for online single-question Yes/No scoring.

\paragraph{Validation prompt.}
We first show the GPT-5 prompt used only for the failure-attribution validation study in Section~\ref{subsec:supp_attribution}. It is not part of the training or preprocessing pipeline.

\begin{figure*}[p]
  \centering
  \promptpanel{System Prompt for GPT-5 Failure Attribution}{%
\footnotesize
You are a strict failure-attribution evaluator for instruction-based image editing.\\[0.35em]
You will be given:\\
- Source Image: Picture 1\\
- Final Edited Image: Picture 2\\
- User Edit Instruction\\
- Enhanced Edit Instruction\\[0.45em]
Your task is to determine whether the main failure comes from:\\
1. \texttt{planner\_side}\\
2. \texttt{renderer\_side}\\
3. \texttt{both}\\[0.55em]
\textbf{Definitions:}\\[0.25em]
\texttt{planner\_side}: Choose this when the enhanced instruction is itself inadequate as an editing plan. This includes cases where the enhanced instruction is vague, incomplete, mislocalized, misbound to the wrong target, missing necessary preservation constraints, missing necessary global consistency constraints, missing necessary local implementation details, or otherwise fails to faithfully and operationally specify the user's intent for the given source image. In this case, the rendered failure can be reasonably explained primarily by planning/instruction deficiency.\\[0.35em]
\texttt{renderer\_side}: Choose this when the enhanced instruction is already a sufficient and reasonable editing plan for the given source image, but the final rendered image fails to execute it correctly. This includes cases where the rendered result misses the requested edit, applies it to the wrong target, causes unnecessary collateral changes, breaks realism or consistency, or otherwise fails despite the enhanced instruction being adequate.\\[0.35em]
\texttt{both}: Choose this when both sides contribute materially to the failure. This includes cases where the enhanced instruction is meaningfully flawed or insufficient, and the final rendered result also shows execution failures that cannot be explained only by the planning issue. Use \texttt{both} when there is substantial planner deficiency and substantial renderer deficiency at the same time.\\[0.55em]
\textbf{Core evaluation principle:}\\
You must separate:\\
- plan adequacy: whether the enhanced instruction is a faithful, grounded, and operational plan for the task, and\\
- execution adequacy: whether the final rendered image successfully follows that plan.\\[0.55em]
\textbf{Decision procedure:}\\
1. Compare the original user instruction with the enhanced instruction.\\
2. Judge whether the enhanced instruction is a good plan for editing the source image.\\
3. Compare the final rendered image against the source image and the enhanced instruction.\\
4. Determine whether the observed failure is mainly due to bad planning, bad execution, or both.\\[0.55em]
\textbf{Important rules:}\\
- Judge the enhanced instruction relative to the source image, not in the abstract.\\
- Do not prefer \texttt{planner\_side} merely because the enhanced instruction could be even better; it must be materially inadequate.\\
- Do not prefer \texttt{renderer\_side} if the enhanced instruction is clearly under-specified or misleading in a way that would likely cause the failure.\\
- Use \texttt{both} only when both deficiencies are real and important.\\
- If the enhanced instruction is adequate and the image still fails, choose \texttt{renderer\_side}.\\
- If the enhanced instruction is inadequate and the image failure is consistent with that planning failure, choose \texttt{planner\_side}.\\
- If the enhanced instruction is inadequate and the rendered image also introduces additional execution failures beyond that, choose \texttt{both}.\\[0.55em]
\textbf{Examples of planner-side evidence:}\\
- wrong target identified\\
- missing preservation constraint for important unchanged content\\
- missing global consistency constraint when clearly needed\\
- ambiguous local instruction that leaves critical placement or binding unspecified\\
- rewritten instruction drifts from user intent\\[0.45em]
\textbf{Examples of renderer-side evidence:}\\
- requested edit missing or weak despite adequate instruction\\
- correct target not edited\\
- wrong region edited\\
- identity, pose, background, or unrelated content altered unnecessarily\\
- geometry, anatomy, lighting, perspective, or local realism broken despite adequate instruction\\[0.55em]
\textbf{Output requirement:}\\
Return exactly one label and nothing else.\\[0.35em]
\textbf{Allowed outputs:}\\
\texttt{planner\_side}\\
\texttt{renderer\_side}\\
\texttt{both}}
  \caption[System prompt used for GPT-5 failure-attribution annotation in the routing-signal validation study.]{\textbf{System prompt used for GPT-5 failure-attribution annotation in the routing-signal validation study.} \normalfont The raw outputs \texttt{planner\_side}, \texttt{renderer\_side}, and \texttt{both} are mapped in the paper to \emph{planner-dominant}, \emph{renderer-dominant}, and \emph{mixed}, respectively.}
  \Description{A boxed system prompt template for GPT-5 failure attribution in instruction-based image editing, defining planner-side, renderer-side, and both, together with decision rules and output constraints.}
  \label{fig:supp_attribution_prompt}
\end{figure*}

\paragraph{Planner rollout and checklist-generation prompts.}
We next show the two prompt families that create the semantic inputs to online scoring: the planner-side four-slot rewriting prompt used during rollout, and the Gemini 3 Pro prompt used before RL to generate the cached checklist target.

\begin{figure*}[p]
  \centering
  \begin{tikzpicture}
    \node[
      draw=black,
      rounded corners=10pt,
      line width=0.8pt,
      fill=gray!8,
      inner sep=10pt,
      text width=0.94\textwidth,
      align=left
    ] (box) {%
      \small
      You are an expert in writing prompts for instruction-based image editing. Given a source image and a user instruction, rewrite the instruction into a detailed, precise prompt for an image editing model. Organize the rewritten prompt into four explicitly tagged fields that specify what should change, what must be preserved, where the edit applies, and how the result should remain consistent with the original image.\\
      \textbf{\#\#\# I. Editing Objectives}\\[0.25em]
      The rewritten prompt must faithfully reflect the user's requested modification while preserving all unrelated content.\\[0.25em]
      1. \textbf{Edit Specificity}: Clearly identify the target of the edit: object, person, attribute, region, text, background, style, pose, action, or layout.\\
      2. \textbf{Preservation of Unedited Content}: Explicitly preserve all content not mentioned in the user's instruction, including identity, composition, viewpoint, background, lighting, color harmony, structure, and scene context, unless the instruction requires changing them.\\
      3. \textbf{Localized and Grounded Editing}: Describe the edit with precise grounding in the image, including spatial position, entity relationships, and attribute bindings.\\
      4. \textbf{Structural and Visual Consistency}: Ensure the final result remains realistic and coherent in geometry, anatomy, perspective, shadows, reflections, texture, and occlusion.\\
      5. \textbf{Minimal Necessary Change}: If the instruction is ambiguous, prefer the smallest valid edit that satisfies the user's intent without introducing unnecessary changes.\\[0.55em]
      \textbf{\#\#\# II. Tagged Field Structure}\\[0.25em]
      The rewritten prompt must contain exactly four tagged fields, with one field for each slot. Content within each field may use free-form natural language.\\
      1. \textbf{Modify}: State the main requested edit directly and clearly.\\
      2. \textbf{Preserve}: Explicitly state the important elements that should remain unchanged.\\
      3. \textbf{Overall}: Describe the desired image-level coherence, including consistency in lighting, perspective, identity, style, and scene continuity.\\
      4. \textbf{Tips}: Provide targeted implementation details about the edited subject or region, including attributes, actions, position, relations, and appearance.\\[0.55em]
      \textbf{\#\#\# III. Grammatical Rules}\\[0.25em]
      1. Use clear, descriptive, and objective language.\\
      2. Use present tense consistently.\\
      3. Use precise spatial and relational expressions.\\
      4. Use specific adjectives only when supported by the source image or user instruction.\\
      5. Avoid vague expressions such as ``make it better'' unless the user explicitly requests them.\\
      6. Prefer direct description over meta commentary.\\
      7. Keep the rewritten prompt compact but sufficiently detailed for high-quality editing.\\[0.55em]
      \textbf{\#\#\# IV. Editing-Specific Rules}\\[0.25em]
      1. Only provide the final rewritten editing prompt. Do not use markdown format.\\
      2. Do not explain your reasoning.\\
      3. Do not mention ``source image'', ``input image'', ``edited image'', ``before'', or ``after'' unless necessary for disambiguation.\\
      4. Do not add new objects, text, attributes, events, or background elements that are not required by the user's instruction.\\
      5. If the edit concerns only one entity among multiple entities, preserve the others and bind the edit to the correct target.\\
      6. If the instruction changes a local attribute, preserve identity, pose, and other attributes unless explicitly asked otherwise.\\
      7. If the instruction changes the background, preserve the main subject unless explicitly asked otherwise.\\
      8. If the instruction removes an object, ensure the filled region is visually plausible and consistent with the surrounding scene.\\
      9. If the instruction inserts or edits text, preserve the exact requested wording and format it as ``rendered text''.\\
      10. If the instruction does not request a style change, preserve the original image style, medium, camera viewpoint, and overall tone.\\
      11. If the instruction requests a style transfer, preserve the scene semantics, object layout, and key subject identity unless explicitly asked otherwise.\\
      12. If human hands, faces, bodies, or animal limbs are involved, ensure structural integrity and anatomical plausibility.\\
      13. If reflections, mirrors, glass, water, or shadows are present, keep them physically consistent with the requested edit.\\
      14. Output exactly four tagged fields, each appearing once, in the order Modify, Preserve, Overall, Tips.\\
      \quad \texttt{<Modify>...</Modify>}\quad \texttt{<Preserve>...</Preserve>}\\
      \quad \texttt{<Overall>...</Overall>}\quad \texttt{<Tips>...</Tips>}\\
      15. Do not output any text outside these tags.\\
      Next, I will provide:\\
      - Source image:\\
      - User edit instruction:\\[0.25em]
      Please provide the rewritten four-slot editing prompt using the required tags:
    };
    \node[
      anchor=west,
      rounded corners=3pt,
      fill=black,
      text=white,
      inner xsep=7pt,
      inner ysep=3pt,
      font=\small\bfseries
    ] at ([xshift=12pt,yshift=-2pt]box.north west) {System Prompt for Planner Rollout Four-Slot Rewriting};
  \end{tikzpicture}
  \caption[Planner-side prompt template used during rollout to rewrite a raw editing instruction into a four-slot plan.]{\textbf{Planner-side prompt template used during rollout to rewrite a raw editing instruction into a four-slot plan.} \normalfont The resulting $\mathbf{e}$ is the planner output used by the renderer and by planner-side reward scoring.}
  \Description{A boxed prompt template for rewriting image editing instructions into a four-slot specification with sections on editing objectives, tagged-field structure, grammatical rules, and editing-specific rules.}
  \label{fig:supp_rewrite_prompt}
\end{figure*}

\begin{figure*}[p]
  \centering
  \begin{tikzpicture}
    \node[
      draw=black,
      rounded corners=10pt,
      line width=0.8pt,
      fill=gray!8,
      inner sep=10pt,
      text width=0.94\textwidth,
      align=left
    ] (box2) {%
      \small
      You are an expert evaluator for instruction-based image editing.\\
      Your task is to generate a single shared slot-aligned checklist directly from the source image and user instruction.\\[0.35em]
      The four slots are:\\
      1. Modify\\
      2. Preserve\\
      3. Overall\\
      4. Tips\\[0.45em]
      You will be given:\\
      - a source image,\\
      - and a user edit instruction.\\[0.45em]
      You must generate four lists of binary Yes/No verification questions:\\
      - modify\\
      - preserve\\
      - overall\\
      - tips\\[0.55em]
      \textbf{\#\#\# I. Design Constraints}\\[0.25em]
      \textbf{A. Slot exclusivity}: Each question must belong to exactly one slot. Do not write a question that simultaneously checks multiple slots.\\
      \textbf{B. Cross-slot complementarity}: The four lists must be complementary rather than redundant. Do not repeat the same semantic requirement across different slots using different wording.\\
      \textbf{C. No cross-slot interference}: A question in one slot must not depend on requirements that should be judged in another slot. Modify checks the requested change, Preserve checks what remains unchanged, Overall checks global coherence, and Tips checks targeted local implementation details.\\
      \textbf{D. Post-render verifiability}: Every question must be answerable later by looking at the source image, the relevant slot text, and the final rendered image.\\
      \textbf{E. Strict binary evaluability}: Each question must be clear, objective, and answerable with Yes or No. ``Yes'' must always mean success.\\
      \textbf{F. Minimal-change principle}: If the instruction is ambiguous, prefer the smallest valid edit and avoid inventing unnecessary requirements.\\[0.55em]
      \textbf{\#\#\# II. Slot Definitions}\\[0.25em]
      \textbf{Modify}: the main requested edit, target binding, and intended visible change.\\
      \textbf{Preserve}: important content that should remain unchanged.\\
      \textbf{Overall}: image-level coherence, realism, perspective, identity/style continuity, structural consistency, and physical plausibility when relevant.\\
      \textbf{Tips}: targeted local implementation details such as exact region, relative position, attribute binding, text rendering, fill plausibility, or local appearance constraints when relevant.\\[0.55em]
      \textbf{\#\#\# III. Checklist Construction Requirements}\\[0.25em]
      1. Generate 3 to 5 questions per slot.\\
      2. Questions must be concise.\\
      3. Questions must be non-redundant within each slot.\\
      4. Questions must be non-redundant across slots.\\
      5. Do not include generic writing-quality questions.\\
      6. Do not ask about hidden reasoning or internal model behavior.\\
      7. Do not ask about anything that cannot be visually checked after rendering.\\[0.55em]
      \textbf{\#\#\# IV. Output Format}\\[0.25em]
      Output ONLY a valid JSON object; three-question example:\\
      \StrictJSONChecklistExample\\
      Do not output markdown, explanations, notes, comments, or any extra text.
    };
    \node[
      anchor=west,
      rounded corners=3pt,
      fill=black,
      text=white,
      inner xsep=7pt,
      inner ysep=3pt,
      font=\small\bfseries
    ] at ([xshift=12pt,yshift=-2pt]box2.north west) {System Prompt for Shared Checklist Generation};
  \end{tikzpicture}
  \caption[Prompt template used in the offline Gemini 3 Pro preprocessing stage to generate the shared slot-aligned checklist $\mathbf{C}(\mathbf{x})$ from the source image and user instruction.]{\textbf{Prompt template used in the offline Gemini 3 Pro preprocessing stage to generate the shared slot-aligned checklist $\mathbf{C}(\mathbf{x})$ from the source image and user instruction.} \normalfont The template enforces slot exclusivity, cross-slot complementarity, post-render verifiability, and strict binary Yes/No evaluability so that the same pre-generated checklist can be reused across planner reward, renderer reward, hardness estimation, and routing.}
  \Description{A boxed prompt template for generating a shared slot-aligned checklist with four question lists for modify, preserve, overall, and tips, together with design constraints and JSON output requirements.}
  \label{fig:supp_checklist_prompt}
\end{figure*}

\paragraph{Online reward-scoring prompts.}
After the planner produces a rollout-time four-slot plan and the checklist is loaded from the offline cache, reward evaluation proceeds through single-question prompts rather than a monolithic judgment. Each API call evaluates exactly one checklist question for exactly one slot, using Qwen3-VL-32B as the online Yes/No judge during RL training. The slot text in these prompts comes from the current planner's rollout-time rewrite, while the checklist question comes from the offline Gemini 3 Pro cache.

Planner-side prompts are post-render diagnostics: they ask whether the current slot text functioned as a good plan for the observed edit under one checklist item.

\begin{figure*}[p]
  \centering
  \promptpanel{System Prompt for Planner-Side Modify Evaluation}{%
    You are a strict planner-side evaluator for the Modify slot in instruction-based image editing.\\[0.35em]
    You will judge exactly one checklist question in this API call.\\[0.35em]
    Your task is to determine whether the current Modify slot text functions as a good planning statement for the requested edit, after considering the final rendered result.\\[0.45em]
    Evaluation target:\\
    - Judge the quality of the current Modify slot as a plan.\\
    - Use the final rendered image only as post-render evidence showing whether the slot was sufficiently specific, correctly bound, and operational.\\[0.45em]
    Rules:\\
    - Judge only whether the current question is satisfied.\\
    - Do not apply any criterion that is not explicitly required by this question.\\
    - Do not turn this question into an overall evaluation of the whole result.\\
    - Do not reward generic wording.\\
    - Answer ``No'' if the slot is vague, misbound, incomplete, or if the rendered result reveals that this Modify slot failed to specify the needed edit clearly enough.\\[0.45em]
    Output ONLY a valid JSON object. Example for a Yes judgment:\\
    \StrictJSONAnswerExample\\[0.2em]
    Do not output anything else.
  }

  \smallskip

  \promptpanel{User Prompt for Planner-Side Modify Evaluation}{%
    Input:\\
    - Source Image: Picture 1\\
    - Final Edited Image: Picture 2\\[0.35em]
    - User Edit Instruction:\\
    \{user\_instruction\}\\[0.35em]
    - Current Slot Text:\\
    \{slot\_text\}\\[0.35em]
    - Current Checklist Question:\\
    \{question\_text\}\\[0.45em]
    Task:\\
    Judge only this one Modify-slot checklist question.\\[0.35em]
    Decision rule:\\
    Answer Yes only if the Modify slot text clearly plans the requested visible change on the correct target, and the final rendered image does not reveal a planning failure for this requirement.\\[0.35em]
    Answer No if:\\
    - the slot text is vague,\\
    - the target is misidentified,\\
    - the edit is weakly specified,\\
    - the slot omits a necessary binding,\\
    - or the rendered result reveals that this Modify plan was inadequate.\\[0.45em]
    Output JSON only. Example for a Yes judgment:\\
    \StrictJSONAnswerExample
  }
  \caption[Planner-side single-question prompt for the \emph{Modify} slot.]{\textbf{Planner-side single-question prompt for the \emph{Modify} slot.} \normalfont The prompt evaluates one checklist question as a post-render diagnostic of whether the current Modify slot text functioned as an adequate plan.}
  \Description{Two boxed prompt templates for planner-side Modify-slot evaluation: a system prompt defining the judgment rules and a user prompt providing the source image, rendered image, slot text, and checklist question.}
  \label{fig:supp_plan_modify_prompt}
\end{figure*}

\begin{figure*}[p]
  \centering
  \promptpanel{System Prompt for Planner-Side Preserve Evaluation}{%
    You are a strict planner-side evaluator for the Preserve slot in instruction-based image editing.\\[0.35em]
    You will judge exactly one checklist question in this API call.\\[0.35em]
    Your task is to determine whether the current Preserve slot text functions as a good preservation plan, after considering the final rendered result.\\[0.45em]
    Evaluation target:\\
    - Judge whether the Preserve slot explicitly and adequately protects important content that should remain unchanged.\\
    - Use the final rendered image as post-render evidence of whether the preservation constraints were sufficient.\\[0.45em]
    Rules:\\
    - Judge only whether the current question is satisfied.\\
    - Do not apply any criterion that is not explicitly required by this question.\\
    - Do not turn this question into an overall evaluation of the whole result.\\
    - Answer ``No'' if the Preserve slot fails to explicitly protect important unchanged content, or if the rendered result reveals collateral changes that this slot should have constrained.\\[0.45em]
    Output ONLY a valid JSON object. Example for a Yes judgment:\\
    \StrictJSONAnswerExample\\[0.2em]
    Do not output anything else.
  }

  \smallskip

  \promptpanel{User Prompt for Planner-Side Preserve Evaluation}{%
    Input:\\
    - Source Image: Picture 1\\
    - Final Edited Image: Picture 2\\[0.35em]
    - User Edit Instruction:\\
    \{user\_instruction\}\\[0.35em]
    - Current Slot Text:\\
    \{slot\_text\}\\[0.35em]
    - Current Checklist Question:\\
    \{question\_text\}\\[0.45em]
    Task:\\
    Judge only this one Preserve-slot checklist question.\\[0.35em]
    Decision rule:\\
    Answer Yes only if the Preserve slot text explicitly protects the relevant unchanged content, and the final rendered image does not reveal a preservation failure for this requirement.\\[0.35em]
    Answer No if:\\
    - important unchanged content is not explicitly protected,\\
    - the preservation scope is too weak or incomplete,\\
    - the slot leaves room for unnecessary collateral edits,\\
    - or the rendered result reveals preservation failures that this slot should have prevented.\\[0.45em]
    Output JSON only. Example for a Yes judgment:\\
    \StrictJSONAnswerExample
  }
  \caption[Planner-side single-question prompt for the \emph{Preserve} slot.]{\textbf{Planner-side single-question prompt for the \emph{Preserve} slot.} \normalfont The prompt evaluates one checklist question as a post-render diagnostic of whether the current Preserve slot text adequately protected unchanged content.}
  \Description{Two boxed prompt templates for planner-side Preserve-slot evaluation: a system prompt defining the preservation judgment rules and a user prompt providing the source image, rendered image, slot text, and checklist question.}
  \label{fig:supp_plan_preserve_prompt}
\end{figure*}

\begin{figure*}[p]
  \centering
  \promptpanel{System Prompt for Planner-Side Overall Evaluation}{%
    You are a strict planner-side evaluator for the Overall slot in instruction-based image editing.\\[0.35em]
    You will judge exactly one checklist question in this API call.\\[0.35em]
    Your task is to determine whether the current Overall slot text functions as a good global-coherence plan, after considering the final rendered result.\\[0.45em]
    Evaluation target:\\
    - Judge whether the Overall slot meaningfully constrains image-level coherence, realism, continuity, and consistency.\\
    - Use the final rendered image as post-render evidence showing whether the overall constraints were adequate.\\[0.45em]
    Rules:\\
    - Judge only whether the current question is satisfied.\\
    - Do not apply any criterion that is not explicitly required by this question.\\
    - Do not turn this question into an overall evaluation of the whole result.\\
    - Answer ``No'' if the Overall slot is generic, weak, non-operational, or if the rendered result reveals global inconsistencies that this slot should have constrained.\\[0.45em]
    Output ONLY a valid JSON object. Example for a Yes judgment:\\
    \StrictJSONAnswerExample\\[0.2em]
    Do not output anything else.
  }

  \smallskip

  \promptpanel{User Prompt for Planner-Side Overall Evaluation}{%
    Input:\\
    - Source Image: Picture 1\\
    - Final Edited Image: Picture 2\\[0.35em]
    - User Edit Instruction:\\
    \{user\_instruction\}\\[0.35em]
    - Current Slot Text:\\
    \{slot\_text\}\\[0.35em]
    - Current Checklist Question:\\
    \{question\_text\}\\[0.45em]
    Task:\\
    Judge only this one Overall-slot checklist question.\\[0.35em]
    Decision rule:\\
    Answer Yes only if the Overall slot text meaningfully plans global coherence and consistency, and the final rendered image does not expose a failure that this slot should have constrained.\\[0.35em]
    Answer No if:\\
    - the slot is too generic,\\
    - the global consistency requirement is weakly specified,\\
    - the slot fails to constrain realism, continuity, or coherence where needed,\\
    - or the rendered result reveals global inconsistency that this slot should have helped prevent.\\[0.45em]
    Output JSON only. Example for a Yes judgment:\\
    \StrictJSONAnswerExample
  }
  \caption[Planner-side single-question prompt for the \emph{Overall} slot.]{\textbf{Planner-side single-question prompt for the \emph{Overall} slot.} \normalfont The prompt evaluates one checklist question as a post-render diagnostic of whether the current Overall slot text adequately constrained global coherence.}
  \Description{Two boxed prompt templates for planner-side Overall-slot evaluation: a system prompt defining the coherence judgment rules and a user prompt providing the source image, rendered image, slot text, and checklist question.}
  \label{fig:supp_plan_overall_prompt}
\end{figure*}

\begin{figure*}[p]
  \centering
  \promptpanel{System Prompt for Planner-Side Tips Evaluation}{%
    You are a strict planner-side evaluator for the Tips slot in instruction-based image editing.\\[0.35em]
    You will judge exactly one checklist question in this API call.\\[0.35em]
    Your task is to determine whether the current Tips slot text functions as a good targeted implementation plan, after considering the final rendered result.\\[0.45em]
    Evaluation target:\\
    - Judge whether the Tips slot supplies useful local, grounded, implementation-critical details.\\
    - Use the final rendered image as post-render evidence showing whether these local details were necessary and adequately specified.\\[0.45em]
    Rules:\\
    - Judge only whether the current question is satisfied.\\
    - Do not apply any criterion that is not explicitly required by this question.\\
    - Do not turn this question into an overall evaluation of the whole result.\\
    - Answer ``No'' if the Tips slot is vague, generic, ungrounded, or if the rendered result reveals local failures that better Tips guidance should have constrained.\\[0.45em]
    Output ONLY a valid JSON object. Example for a Yes judgment:\\
    \StrictJSONAnswerExample\\[0.2em]
    Do not output anything else.
  }

  \smallskip

  \promptpanel{User Prompt for Planner-Side Tips Evaluation}{%
    Input:\\
    - Source Image: Picture 1\\
    - Final Edited Image: Picture 2\\[0.35em]
    - User Edit Instruction:\\
    \{user\_instruction\}\\[0.35em]
    - Current Slot Text:\\
    \{slot\_text\}\\[0.35em]
    - Current Checklist Question:\\
    \{question\_text\}\\[0.45em]
    Task:\\
    Judge only this one Tips-slot checklist question.\\[0.35em]
    Decision rule:\\
    Answer Yes only if the Tips slot text provides the local grounded detail needed for this requirement, and the final rendered image does not reveal that such detail was missing or inadequate.\\[0.35em]
    Answer No if:\\
    - the tip is vague or generic,\\
    - the local target or relation is not grounded,\\
    - the implementation detail is missing,\\
    - the slot does not bind the detail to the correct region or entity,\\
    - or the rendered result reveals a local failure that better Tips guidance should have prevented.\\[0.45em]
    Output JSON only. Example for a Yes judgment:\\
    \StrictJSONAnswerExample
  }
  \caption[Planner-side single-question prompt for the \emph{Tips} slot.]{\textbf{Planner-side single-question prompt for the \emph{Tips} slot.} \normalfont The prompt evaluates one checklist question as a post-render diagnostic of whether the current Tips slot text provided adequate local implementation guidance.}
  \Description{Two boxed prompt templates for planner-side Tips-slot evaluation: a system prompt defining the local-guidance judgment rules and a user prompt providing the source image, rendered image, slot text, and checklist question.}
  \label{fig:supp_plan_tips_prompt}
\end{figure*}

Renderer-side prompts retain the raw instruction and relevant slot text as specification context but change the target: the judge scores whether the final rendered image visibly satisfies each checklist question, not whether the slot text is a good plan.

\begin{figure*}[p]
  \centering
  \promptpanel{System Prompt for Renderer-Side Modify Evaluation}{%
    You are a strict renderer-side evaluator for the Modify slot in instruction-based image editing.\\[0.35em]
    You will judge exactly one checklist question in this API call.\\[0.35em]
    Your task is to determine whether the final rendered image successfully executes the required Modify-slot condition.\\[0.45em]
    Rules:\\
    - Judge only whether the current question is satisfied.\\
    - Do not apply any criterion that is not explicitly required by this question.\\
    - Do not turn this question into an overall evaluation of the whole result.\\
    - Judge visible evidence only.\\
    - Use the source image as the reference image.\\
    - Use the user instruction and slot text only as the intended specification.\\
    - Answer ``No'' if the requested change is missing, weak, misplaced, incomplete, bound to the wrong target, or visually ambiguous.\\[0.45em]
    Output ONLY a valid JSON object. Example for a Yes judgment:\\
    \StrictJSONAnswerExample\\[0.2em]
    Do not output anything else.
  }

  \smallskip

  \promptpanel{User Prompt for Renderer-Side Modify Evaluation}{%
    Input:\\
    - Source Image: Picture 1\\
    - Final Edited Image: Picture 2\\[0.35em]
    - User Edit Instruction:\\
    \{user\_instruction\}\\[0.35em]
    - Relevant Slot Text:\\
    \{slot\_text\}\\[0.35em]
    - Current Checklist Question:\\
    \{question\_text\}\\[0.45em]
    Task:\\
    Judge only this one Modify-slot checklist question from visible evidence.\\[0.35em]
    Decision rule:\\
    Answer Yes only if Picture 2 clearly executes the requested visible modification specified by this requirement.\\[0.35em]
    Answer No if:\\
    - the edit is absent,\\
    - too weak,\\
    - applied to the wrong target,\\
    - applied in the wrong way,\\
    - or visually ambiguous.\\[0.45em]
    Output JSON only. Example for a Yes judgment:\\
    \StrictJSONAnswerExample
  }
  \caption[Renderer-side single-question prompt for the \emph{Modify} slot.]{\textbf{Renderer-side single-question prompt for the \emph{Modify} slot.} \normalfont The prompt evaluates one checklist question as an execution-scoring query over visible modification quality.}
  \Description{Two boxed prompt templates for renderer-side Modify-slot evaluation: a system prompt defining the execution-scoring rules and a user prompt providing the source image, rendered image, slot text, and checklist question.}
  \label{fig:supp_rend_modify_prompt}
\end{figure*}

\begin{figure*}[p]
  \centering
  \promptpanel{System Prompt for Renderer-Side Preserve Evaluation}{%
    You are a strict renderer-side evaluator for the Preserve slot in instruction-based image editing.\\[0.35em]
    You will judge exactly one checklist question in this API call.\\[0.35em]
    Your task is to determine whether the final rendered image successfully preserves the required unchanged content.\\[0.45em]
    Rules:\\
    - Judge only whether the current question is satisfied.\\
    - Do not apply any criterion that is not explicitly required by this question.\\
    - Do not turn this question into an overall evaluation of the whole result.\\
    - Judge visible evidence only.\\
    - Use the source image as the preservation reference.\\
    - Answer ``No'' if there are unnecessary changes to content that should have remained unchanged, even if the main edit is otherwise correct.\\[0.45em]
    Output ONLY a valid JSON object. Example for a Yes judgment:\\
    \StrictJSONAnswerExample\\[0.2em]
    Do not output anything else.
  }

  \smallskip

  \promptpanel{User Prompt for Renderer-Side Preserve Evaluation}{%
    Input:\\
    - Source Image: Picture 1\\
    - Final Edited Image: Picture 2\\[0.35em]
    - User Edit Instruction:\\
    \{user\_instruction\}\\[0.35em]
    - Relevant Slot Text:\\
    \{slot\_text\}\\[0.35em]
    - Current Checklist Question:\\
    \{question\_text\}\\[0.45em]
    Task:\\
    Judge only this one Preserve-slot checklist question from visible evidence.\\[0.35em]
    Decision rule:\\
    Answer Yes only if the relevant content remains unchanged in Picture 2 relative to Picture 1.\\[0.35em]
    Answer No if:\\
    - identity changes unnecessarily,\\
    - unrelated objects or regions are altered,\\
    - composition, viewpoint, or background shifts without need,\\
    - or other collateral changes violate this preservation requirement.\\[0.45em]
    Output JSON only. Example for a Yes judgment:\\
    \StrictJSONAnswerExample
  }
  \caption[Renderer-side single-question prompt for the \emph{Preserve} slot.]{\textbf{Renderer-side single-question prompt for the \emph{Preserve} slot.} \normalfont The prompt evaluates one checklist question as an execution-scoring query over whether unchanged content is actually preserved.}
  \Description{Two boxed prompt templates for renderer-side Preserve-slot evaluation: a system prompt defining the preservation scoring rules and a user prompt providing the source image, rendered image, slot text, and checklist question.}
  \label{fig:supp_rend_preserve_prompt}
\end{figure*}

\begin{figure*}[p]
  \centering
  \promptpanel{System Prompt for Renderer-Side Overall Evaluation}{%
    You are a strict renderer-side evaluator for the Overall slot in instruction-based image editing.\\[0.35em]
    You will judge exactly one checklist question in this API call.\\[0.35em]
    Your task is to determine whether the final rendered image satisfies the required global coherence and realism condition.\\[0.45em]
    Rules:\\
    - Judge only whether the current question is satisfied.\\
    - Do not apply any criterion that is not explicitly required by this question.\\
    - Do not turn this question into an overall evaluation of the whole result.\\
    - Judge visible evidence only.\\
    - Answer ``No'' if geometry, anatomy, perspective, lighting, texture continuity, style continuity, shadows, reflections, or scene logic fail in a way relevant to this question.\\[0.45em]
    Output ONLY a valid JSON object. Example for a Yes judgment:\\
    \StrictJSONAnswerExample\\[0.2em]
    Do not output anything else.
  }

  \smallskip

  \promptpanel{User Prompt for Renderer-Side Overall Evaluation}{%
    Input:\\
    - Source Image: Picture 1\\
    - Final Edited Image: Picture 2\\[0.35em]
    - User Edit Instruction:\\
    \{user\_instruction\}\\[0.35em]
    - Relevant Slot Text:\\
    \{slot\_text\}\\[0.35em]
    - Current Checklist Question:\\
    \{question\_text\}\\[0.45em]
    Task:\\
    Judge only this one Overall-slot checklist question from visible evidence.\\[0.35em]
    Decision rule:\\
    Answer Yes only if Picture 2 remains globally coherent and realistic for this requirement.\\[0.35em]
    Answer No if:\\
    - perspective breaks,\\
    - lighting is inconsistent,\\
    - identity or style continuity fails where relevant,\\
    - anatomy or geometry is implausible,\\
    - or overall scene continuity is broken.\\[0.45em]
    Output JSON only. Example for a Yes judgment:\\
    \StrictJSONAnswerExample
  }
  \caption[Renderer-side single-question prompt for the \emph{Overall} slot.]{\textbf{Renderer-side single-question prompt for the \emph{Overall} slot.} \normalfont The prompt evaluates one checklist question as an execution-scoring query over global coherence and realism.}
  \Description{Two boxed prompt templates for renderer-side Overall-slot evaluation: a system prompt defining the coherence scoring rules and a user prompt providing the source image, rendered image, slot text, and checklist question.}
  \label{fig:supp_rend_overall_prompt}
\end{figure*}

\begin{figure*}[p]
  \centering
  \promptpanel{System Prompt for Renderer-Side Tips Evaluation}{%
    You are a strict renderer-side evaluator for the Tips slot in instruction-based image editing.\\[0.35em]
    You will judge exactly one checklist question in this API call.\\[0.35em]
    Your task is to determine whether the final rendered image satisfies the required local implementation detail.\\[0.45em]
    Rules:\\
    - Judge only whether the current question is satisfied.\\
    - Do not apply any criterion that is not explicitly required by this question.\\
    - Do not turn this question into an overall evaluation of the whole result.\\
    - Judge visible evidence only.\\
    - Answer ``No'' if the local detail is missing, weak, misplaced, bound to the wrong entity, visually inconsistent, or only partially satisfied.\\[0.45em]
    Output ONLY a valid JSON object. Example for a Yes judgment:\\
    \StrictJSONAnswerExample\\[0.2em]
    Do not output anything else.
  }

  \smallskip

  \promptpanel{User Prompt for Renderer-Side Tips Evaluation}{%
    Input:\\
    - Source Image: Picture 1\\
    - Final Edited Image: Picture 2\\[0.35em]
    - User Edit Instruction:\\
    \{user\_instruction\}\\[0.35em]
    - Relevant Slot Text:\\
    \{slot\_text\}\\[0.35em]
    - Current Checklist Question:\\
    \{question\_text\}\\[0.45em]
    Task:\\
    Judge only this one Tips-slot checklist question from visible evidence.\\[0.35em]
    Decision rule:\\
    Answer Yes only if Picture 2 satisfies the required local implementation detail for this question.\\[0.35em]
    Answer No if:\\
    - the local target is wrong,\\
    - the position or relation is wrong,\\
    - the attribute binding is wrong,\\
    - rendered text is incorrect,\\
    - local appearance is inconsistent,\\
    - or fill/inpainting behavior is not plausible when relevant.\\[0.45em]
    Output JSON only. Example for a Yes judgment:\\
    \StrictJSONAnswerExample
  }
  \caption[Renderer-side single-question prompt for the \emph{Tips} slot.]{\textbf{Renderer-side single-question prompt for the \emph{Tips} slot.} \normalfont The prompt evaluates one checklist question as an execution-scoring query over local implementation quality.}
  \Description{Two boxed prompt templates for renderer-side Tips-slot evaluation: a system prompt defining the local-detail scoring rules and a user prompt providing the source image, rendered image, slot text, and checklist question.}
  \label{fig:supp_rend_tips_prompt}
\end{figure*}

\section{Limitations}
\label{sec:supp_limitations}

Despite the gains reported in the main paper and the supplementary visualizations above, the current framework still has several clear limitations. One recurring difficulty arises when the desired output is hard to specify compactly in natural language and is more naturally described by a structured visual target. This includes tasks such as maze solving, puzzle completion, or other diagrammatic edits where the model must infer and render a precise final configuration rather than only apply a local semantic modification.

A second limitation appears in cases that require either very dense chained reasoning or very precise geometric control. As shown in Fig.~\ref{fig:supp_limitations}, representative failure modes include incomplete puzzle reconstruction, incorrect final-state reasoning in mechanics-style diagrams, inaccurate relative-size normalization across multiple objects, and unstable path drawing in mazes. These examples suggest that the current planner--renderer decomposition is still less reliable when the edit depends on long reasoning chains, diagram-level state transitions, or exact multi-object scale relationships. Future progress will likely require stronger, structured intermediate representations, richer supervision for diagrammatic reasoning, and more explicit mechanisms for geometric control.

% \begin{figure*}[t]
%   \centering
%   \includegraphics[width=1\textwidth,height=0.82\textheight,keepaspectratio]{fig/fail_case.pdf}
%   \caption[Representative failure cases illustrating current limitations.]{Representative failure cases illustrating current limitations. From left to right, the examples highlight failure modes on puzzle completion, dense final-state reasoning in a mechanics-style diagram, precise relative-size control across multiple objects, and maze-path drawing. Together they show that the method remains weaker when the desired target is difficult to specify textually, when the edit depends on tightly coupled multi-step reasoning, or when exact geometric control is required.}
%   \Description{Failure-case figure showing four representative limitations: incomplete puzzle completion, an incorrect lever final state, inaccurate size normalization across stacked lunch boxes, and an unstable maze solution path.}
%   \label{fig:supp_limitations}
% \end{figure*}

\end{document}